\pdfoutput=1
\RequirePackage{fix-cm}
\documentclass[11pt]{article}
\usepackage[preprint]{acl}
\usepackage{times}
\usepackage{latexsym}
\usepackage[T1]{fontenc}
\usepackage[utf8]{inputenc}
\usepackage{microtype}
\usepackage{inconsolata}
\usepackage{graphicx}
\usepackage{booktabs}
\usepackage{amsmath,amssymb}
\usepackage{multirow}
\usepackage{xcolor}
\usepackage{colortbl}
\usepackage{enumitem}
\usepackage{placeins}
\usepackage{adjustbox}
\usepackage{array}
\usepackage[most]{tcolorbox}

\definecolor{acllink}{HTML}{000099}
\hypersetup{colorlinks=true, citecolor=acllink, linkcolor=acllink, urlcolor=acllink,
  pdftitle={The Best Second Opinion Is Right Where the Primary Is Wrong: Partner Selection for Selective Span Tagging},
  pdfauthor={Varun Kotte}, pdfsubject={}, pdfkeywords={}, pdfcreator={}}

\definecolor{tkfill}{HTML}{F3F7FF}
\definecolor{tkframe}{HTML}{A9C6FF}
\newtcolorbox{takeaway}{enhanced, colback=tkfill, colframe=tkframe, boxrule=0.6pt, arc=2.5pt,
  left=4pt, right=4pt, top=2.5pt, bottom=2.5pt, before skip=5pt plus 2pt, after skip=4pt plus 2pt}

\DeclareMathSizes{10.95}{10.95}{8}{6.6}
\DeclareMathSizes{10}{10}{7}{6.6}
\DeclareMathSizes{9}{9}{6.6}{6.6}
\DeclareMathSizes{8}{8}{6.6}{6.6}
\DeclareMathSizes{7}{7}{6.6}{6.6}
\newcommand{\hd}[1]{\begin{tabular}[b]{@{}c@{}}#1\end{tabular}}
\newcommand{\eaurc}{eAURC}

\newcommand{\savg}{\ensuremath{s_{\mathrm{avg}}}}
\newcommand{\slab}{\ensuremath{s_{\mathrm{lab}}}}
\definecolor{checkrow}{HTML}{ECE8F5}
\definecolor{okverm}{HTML}{D55E00}
\definecolor{okblue}{HTML}{0072B2}

\renewcommand{\topfraction}{0.92}\renewcommand{\dbltopfraction}{0.92}
\renewcommand{\textfraction}{0.08}\renewcommand{\floatpagefraction}{0.85}\renewcommand{\dblfloatpagefraction}{0.85}

\title{The Best Second Opinion Is Right Where the Primary Is Wrong:\\ Partner Selection for Selective Span Tagging}

\newcommand{\affil}{Adobe Inc.}
\author{Varun Kotte \\ \affil \\ \texttt{kottevarun@gmail.com}}

\begin{document}
\maketitle
\begin{abstract}
A span tagger that ranks its outputs by risk can defer the riskiest ones to a person, and a second model can sharpen this ranking.
Which second model should it be?
Partner selection is conditional, not global: the best second opinion is the model that is right where the primary is wrong.
Across eight tagger types, three seeds and three English benchmarks (576 evaluated primary--partner pairings), choosing the partner by held-out accuracy on the primary's errors recovers 82\% of the best partner's gain, against 69--70\% for span F1 or a random pick.
Frozen before a fourth dataset was chosen, the rule replicated there (Spearman 0.40 vs.\ 0.04 for span F1).
Partner value is also directional.
The predictive distribution is the second opinion: hard labels discard half of a partner's value, and scoring the served label improves 94\% of pairings.
Fixed before the fourth dataset was chosen, this score also replicated there.
For LoRA-tuned decoders, independently trained partners beat every readout of the primary network.
Applied zero-shot to 51 MASSIVE locales, every primary has a helpful trained partner in every locale, and the preferred partner follows competence.
\end{abstract}

\section{Introduction}
\label{sec:intro}

\begin{figure}[t]
  \centering
  \includegraphics[width=\columnwidth]{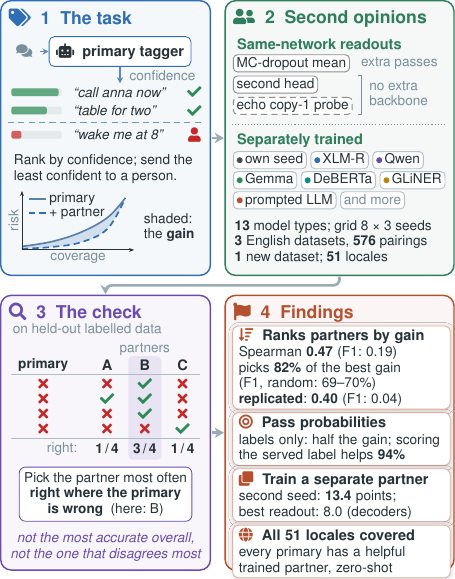}
  \caption{\textbf{The best second opinion is right where the primary is wrong: held-out accuracy on the primary's errors ranks candidate partners by their test gain.}
Panels 1--3: the task, the candidates and the check (the grid is a schematic).
Panel 4: main results (\S\S\ref{sec:curve}--\ref{sec:rq3}).
Spearman: rank correlation with test gain (in brackets: held-out span F1); replicated: the frozen test on a new dataset.
Gains are in points (\% of the margin-only \eaurc{}); readouts: probes and heads on a decoder primary.}
  \label{fig:hero}
\end{figure}

\begin{figure*}[t]
  \centering
  \includegraphics[width=\textwidth]{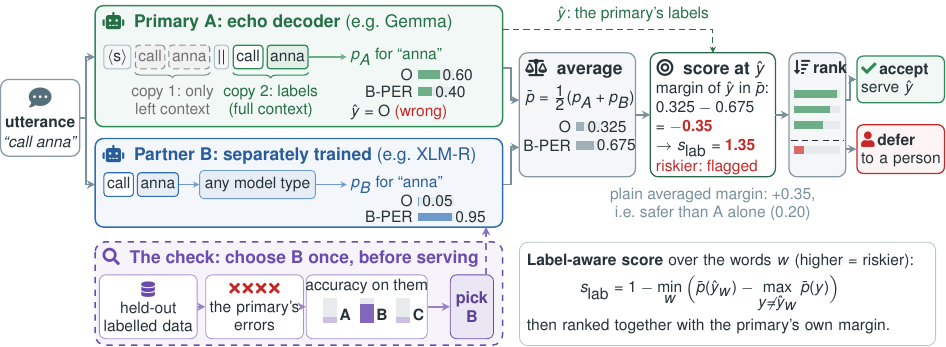}
  \caption{\textbf{A second opinion re-ranks the primary's outputs without changing its labels.}
Here the primary is an echo decoder: it reads the utterance twice and labels each word from copy~2; a separately trained partner~B tags the same words.
Their word label distributions are averaged and scored at the primary's label $\hat y$.
When the average prefers another label, the utterance is flagged as risky; the plain averaged margin would call it safer (illustrative numbers).
The most confident utterances are served with $\hat y$; the rest go to a person.
Dashed box: the check chooses B once, on held-out labelled data.}
  \label{fig:method}
\end{figure*}

A tagger in a search or assistant pipeline should know which of its outputs not to trust.
The system can then defer the riskiest ones to a person (selective prediction; \citealp{geifman2017,xin2021}).
The usual risk signal is the tagger's own confidence.
Here it is the \emph{margin}: the smallest gap between the top two label probabilities over the words of an utterance.
A second model sharpens this ranking.
We average its word probabilities with those of the \emph{primary}, the tagger whose labels are served, and rank by the result.
The primary's labels stay as they are (Figure~\ref{fig:method}).

Second opinions come in two kinds (Figure~\ref{fig:hero}, panel~2).
\emph{Same-network readouts} reuse the primary: the mean of Monte-Carlo (MC) dropout passes \citep{gal2016}, a second output head, or a probe on an earlier reading of the input.
\emph{Separately trained partners} cost a second model: another seed, another model type or a prompted LLM.

Which one should a practitioner choose?
Classical ensemble work asks whether models make complementary errors.
\citet{brill1998} measured this for part-of-speech taggers as the \emph{complementary rate}, and diversity measures such as the double-fault rate build on the same idea \citep{giacinto2001,kuncheva2003,brown2005}.
We turn this idea into a decision rule for a fixed primary: deploy the candidate that is most often right on the primary's errors.
Global accuracy tells you how good a model is; conditional correctness tells you how useful it is to this primary.
The quantity is classical.
What is new is showing that it, not accuracy, is what to select on, and validating this prospectively and across 51 locales.
We ask three questions:

\begin{description}[leftmargin=2.6em,labelwidth=2.2em,labelsep=0.4em,itemsep=0pt,topsep=2pt,font=\normalfont\bfseries]
\item[RQ1] Which second opinion helps? (\S\ref{sec:curve})
\item[RQ2] How should its output be used? (\S\ref{sec:using})
\item[RQ3] Does it hold on new data, in other languages and under serving cost? (\S\ref{sec:rq3})
\end{description}

\noindent All tests follow pre-specified, hashed protocols; the prospective one was frozen before its dataset was chosen.
We use three seeds per tagger, paired and two-way cluster bootstraps, and Holm correction.
A separately written second pipeline re-derived every number, and Appendix Table~\ref{tab:guide} maps each claim to its evidence.
Our contributions are:
\begin{enumerate}[leftmargin=*,itemsep=1pt,topsep=2pt]
\item \textbf{Partner quality is conditional and directional} (\S\ref{sec:curve}). Choosing the partner by held-out accuracy on the primary's errors recovers 82\% of the best partner's gain; span F1, utterance accuracy or a random pick recover 69--70\% (Table~\ref{tab:policies}).
\item \textbf{A prospective replication} (\S\ref{sec:prospective}). Frozen before the dataset was chosen, the rule met all three pre-declared criteria (Spearman 0.40, against 0.04 for span F1).
\item \textbf{The predictive distribution is the second opinion} (\S\ref{sec:using}). Hard labels discard half of a partner's value, and scoring the label that will be served improves 94\% of 576 pairings. Frozen in advance, this score also replicated on the new dataset (+2.7 points). For LoRA-tuned decoders, trained partners beat every readout of the primary.
\item \textbf{Selection under real constraints} (\S\ref{sec:rq3}). Across four English datasets and 51 locales, the preferred partner follows competence. Inside a serving budget, the same rule recovers 85\% of the best affordable gain, against 72\% at random.
\end{enumerate}

\section{Setup}
\label{sec:setup}

\paragraph{Data.}
We use CoNLL-2003 English NER \citep{tjongkimsang2003}, MASSIVE en-US slot filling \citep{fitzgerald2023} and SNIPS slot filling \citep{coucke2018} in the split of \citet{goo2018} (sizes: Appendix Table~\ref{tab:splits}).
Each has a 2,000-utterance \emph{fit slice}, held out of training, for temperatures and combiners; dev selects checkpoints (Appendix~\ref{app:data}).
A fourth dataset, chosen by a frozen rule, serves the prospective test (\S\ref{sec:prospective}), and 51 MASSIVE locales test transfer (\S\ref{sec:shift}).

\paragraph{Taggers.}
We use 13 tagger types in three groups (Appendix Table~\ref{tab:models}).
An \emph{echo tagger} is a decoder that reads its input twice and labels each word from the second copy (\citealp{kukic2026,ewais2026}, after \citealp{springer2025echo}).
\emph{Main:} XLM-R-large \citep{conneau2020}, fully fine-tuned, and two echo taggers trained with LoRA \citep{hu2022lora}: Gemma-4-E2B \citep{gemma4} and Qwen2.5-3B \citep{qwen2025}.
We call them XLM-R, Gemma and Qwen-3B.
\emph{Grid:} the main three plus DeBERTa-v3-large, mDeBERTa-v3-base \citep{he2023debertav3}, XLM-R-large with LoRA, and Qwen2.5-0.5B and Qwen2.5-7B echo taggers, eight types in total.
\emph{Generalisation:} GLiNER-large \citep{zaratiana2024} fine-tuned and zero-shot, XLM-R-XL \citep{goyal2021xlmrxl}, and two Qwen2.5-3B taggers that read the input once.

The echo decoders are as accurate as XLM-R on CoNLL (test span F1 0.930 against 0.922) and SNIPS, and about 2 F1 points behind on MASSIVE (Appendix Table~\ref{tab:f1}).

\paragraph{How a second opinion enters (Figure~\ref{fig:method}).}
Let $p_A$ and $p_B$ be the temperature-scaled \citep{guo2017} word label distributions of the primary $A$ and the second opinion $B$, and $\bar p=\tfrac12(p_A+p_B)$.
Write $p^{(1)}$ and $p^{(2)}$ for the top two probabilities of a word.
The primary's risk score is its margin, $u_A=1-\min_w\big(p_A^{(1)}-p_A^{(2)}\big)$ over the words $w$; higher means riskier.
The second opinion enters through the \emph{averaged margin} $\savg(B)=1-\min_w\big(\bar p^{(1)}-\bar p^{(2)}\big)$, or through the \emph{label-aware score}, which scores the primary's own label $\hat y_w$:
\[\slab(B)=1-\min_w\Big(\bar p(\hat y_w)-\max_{y\neq\hat y_w}\bar p(y)\Big).\]
A logistic regression on the rank-mapped margin and $\savg$ or $\slab$, fit on the fit slice, gives the final score (Appendix~\ref{app:signals}).
Our partner-selection tests use $\savg$; \S\ref{sec:using} tests $\slab$.

\paragraph{What we measure.}
An utterance is wrong if its predicted span set differs from the gold one.
We rank utterances by the combined score and measure the excess area under the risk--coverage curve (\eaurc{}; \citealp{geifman2019}); 0 is perfect.
The \emph{gain} of a second opinion is the drop in \eaurc{} when it is added.
We report it in \emph{points}: percent of the margin-only \eaurc{} (curves: Appendix Figure~\ref{fig:riskcov}).
A partner's \emph{accuracy on the primary's errors} is the share of the primary's wrong utterances on which the partner, used alone, is fully right.
This is \citeauthor{brill1998}'s complementary rate at the utterance level; for a fixed primary, it ranks partners exactly as the double-fault rate does.
\emph{The check} picks the partner with the highest held-out accuracy on the primary's errors.
We score any way of choosing by the share of the best partner's test gain that its choice misses.

\paragraph{Protocol and statistics.}
Every experiment followed a protocol written before it ran, most of them hashed (SHA-256) at writing time (Appendix~\ref{app:prereg}).
For the out-of-sample test, we froze the selection rule and a rule for choosing a new dataset, then scored once on the dataset it chose (\S\ref{sec:prospective}).
Intervals are 95\% bootstrap intervals with 10,000 replicates: gains resample seeds and test utterances, and partner rankings resample primary and partner types.
Confirmatory families use Holm correction.
A second, separately written pipeline re-derived every number (Appendix~\ref{app:recompute}).
Code, protocols and per-item outputs will be released (Appendix~\ref{app:repro}).

\section{RQ1: The best partner fixes the primary's errors}
\label{sec:curve}

\begin{figure*}[t]
  \centering
  \includegraphics[width=\textwidth]{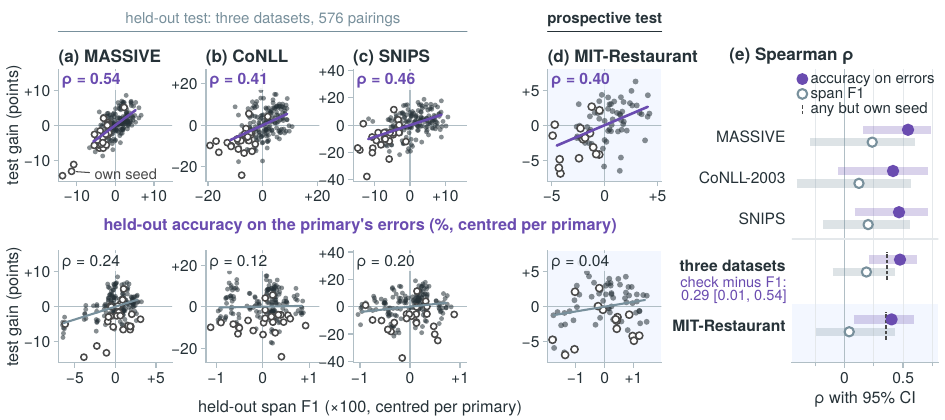}
  \caption{\textbf{Held-out accuracy on the primary's errors ranks a primary's partners by test gain; held-out span F1 does so only weakly.}
Dot: one primary seed paired with one partner (ring: its own second seed); axes centred on the primary's mean over its partners; gain in points; lines are visual aids.
(a)--(c): three datasets, 576 pairings; (d): MIT-Restaurant, 75 pairings, scored once with the rule frozen before the dataset was chosen.
(e): within-primary Spearman $\rho$, averaged over primaries, with 95\% intervals (MIT-Restaurant: the frozen protocol's interval).}
  \label{fig:central}
\end{figure*}

\begin{table}[t]
\centering\footnotesize\setlength{\tabcolsep}{2.1pt}
\begin{tabular}{@{}lrrcrr@{}}
\toprule
 & \multicolumn{3}{c}{3 datasets} & \multicolumn{2}{c}{new dataset}\\
\cmidrule(lr){2-4}\cmidrule(l){5-6}
partner chosen by & $\rho$ & missed & vs.\ check & $\rho$ & missed\\
\midrule
\multicolumn{6}{@{}l}{\emph{no labels on the target dataset}}\\
own seed & -- & 54\% & +36 [10, 43] & -- & 41\%\\
random partner & -- & 30\% & +12 [4, 21] & -- & 24\%\\
any but own seed & .36 & 26\% & +9 [1, 17] & .35 & 20\%$^\dagger$\\
type prior & .37 & 19\% & +1 [$-$10, 16] & .42 & 15\%$^\dagger$\\
pairing prior & .42 & 16\% & $-$2 [$-$10, 7] & \textbf{.57} & \textbf{6\%}$^\dagger$\\
\addlinespace[2pt]
\multicolumn{6}{@{}l}{\emph{partner accuracy (ignores the primary)}}\\
span F1 & .19 & 31\% & +13 [1, 23] & .04 & 26\%\\
utterance acc. & .25 & 30\% & +12 [1, 22] & .07 & 26\%$^\dagger$\\
word accuracy & .33 & 18\% & 0 [$-$4, 9] & .12 & 23\%$^\dagger$\\
\addlinespace[2pt]
\multicolumn{6}{@{}l}{\emph{symmetric diversity (ignores which model is the primary)}}\\
error correlation $\phi$ & .31 & 25\% & +8 [$-$2, 15] & .36 & 24\%\\
disagreement & .23 & 29\% & +11 [$-$2, 19] & .29 & 30\%\\
\addlinespace[2pt]
\multicolumn{6}{@{}l}{\emph{complementarity on the primary's errors}}\\
margin drop & .41 & 20\% & +2 [$-$7, 8] & .37 & 8\%\\
\rowcolor{checkrow}accuracy on errors & .47 & 18\% & & .40 & 11\%\\
type prior + check & .48 & \textbf{15\%} & $-$3 [$-$7, 4] & .49 & 11\%$^\dagger$\\
gain on dev$^{*}$ & \textbf{.51} & 18\% & 0 [$-$8, 7] & .49 & 13\%\\
\midrule
hindsight best type & -- & 15\% & & -- & 10\%\\
\bottomrule
\end{tabular}
\caption{\textbf{Measures conditioned on the primary's errors miss 15--20\% of the best gain on three datasets; span F1, utterance accuracy and disagreement miss 29--31\%, as much as a random pick (30\%).} Missed: share of the best partner's test gain lost (ties split); vs.\ check: difference from the check (shaded) in percentage points, 95\% interval over primary and partner types; $\rho$: within-primary Spearman with test gain; bold: best per column (hindsight excluded). Priors: a partner type's mean gain on the other datasets (pairing prior: same primary type). New dataset: MIT-Restaurant, check frozen in advance; $^\dagger$post hoc. $^{*}$Needs a combiner per candidate. Hindsight: always Gemma or Qwen-3B. Full table: Appendix Table~\ref{tab:policiesfull}.}
\label{tab:policies}
\end{table}

\paragraph{Why the primary's errors matter.}
The primary's labels stay fixed, so a second opinion helps only by reordering risk.
Let $\Delta S=\savg(B)-u_A$ be how much averaging raises an utterance's risk score above the primary's margin.
A partner helps when, on average, it raises risk more on the primary's wrong outputs than on its right ones: $\mathbb{E}[\Delta S\mid A\text{ wrong}]-\mathbb{E}[\Delta S\mid A\text{ right}]>0$.
A useful partner therefore perturbs the risk ranking preferentially where the primary is wrong.
Correctness on the primary's residual errors is a cheap, discrete proxy for this asymmetry: a correct partner moves probability mass away from the primary's wrong served label.
The averaged margin converts that signal into risk only imperfectly, because the averaged distribution can become confident in a different label; Section~\ref{sec:using} shows that scoring the served label directly realises it more faithfully.
The gap itself is the \emph{expected margin drop}; as a selection rule it misses 20\% of the best gain, against 18\% for the check (Table~\ref{tab:policies}).

\paragraph{Held-out accuracy on the primary's errors ranks partners by test gain.}
Each grid type, with three seeds, serves as a primary.
Its candidate partners are its own seed (a second seed of the same type) and one seed of each other type.
This gives 576 pairings over three datasets.
We measure every candidate quantity on held-out data and ask whether it orders each primary's partners like their test gains.

The check does (Figure~\ref{fig:central}a--c).
Its within-primary Spearman correlation is $\rho=0.47$, against 0.19 for held-out span F1, a difference of 0.29 [0.01, 0.54].
It leads on every dataset (Appendix Table~\ref{tab:g2}).

\paragraph{Global performance is not the partner-selection objective.}
Global accuracy tells you how good a model is; conditional correctness tells you how useful it is to this primary.
Picking the candidate with the best held-out span F1 or utterance accuracy misses 30--31\% of the best partner's gain (Table~\ref{tab:policies}).
That is as much as a random pick, and 12--13 percentage points more than the check (18\%); both differences exclude zero.
Word accuracy ties the check on two datasets but falls close to random on the other two (Appendix Table~\ref{tab:fourdata}).

\paragraph{Every measure conditioned on the primary's errors works, so use the cheapest.}
The expected margin drop, gain on dev \citep{caruana2004} and priors from other datasets all miss 15--20\% of the best gain (Table~\ref{tab:policies}).
None differs significantly from the check.
Correlation-normalised diversity measures, error correlation $\phi$ and disagreement, miss 25--29\%.
The check is the simplest: it needs no combiner per candidate and no partner history, only about 100 labelled errors of the primary (Appendix Table~\ref{tab:budget}).

\paragraph{Partner value is directional.}
How much partner B helps primary A says little about how much A helps B.
Over the type pairs of each dataset, the two gains are negatively correlated, on all four datasets (Appendix Figure~\ref{fig:gainmat}).
On CoNLL, Gemma adds 32 points to an XLM-R primary, while XLM-R adds only 8 to a Gemma primary.
Although the double-fault rate is symmetric as a pairwise statistic, second-opinion utility is role-conditioned: the primary determines which residual errors matter.

\paragraph{Never default to the own seed.}
The primary's own seed misses 54\% of the best gain, more than any other rule in Table~\ref{tab:policies}.
With the own seed set aside, the check still misses less than span F1, 18\% against 27\% (Appendix~\ref{app:policies}).

\paragraph{A prior keeps the choice robust to unfamiliar partners.}
The prior matters most for a label-only partner, which passes labels rather than probabilities.
With the prompted LLM of \S\ref{sec:using} among the candidates, the check alone misses 38\% of the best gain, against 35\% for a random pick.
Adding a prior built from each type's gains on the other datasets cuts this to 15\% (Appendix Table~\ref{tab:mixedpools}).

\begin{takeaway}
\textbf{RQ1 answer:} partner selection is conditional, not global. Choose the partner that is right on the primary's errors: every held-out measure of this, and a prior from other datasets, misses 15--20\% of the best gain. Span F1 and utterance accuracy miss 30--31\%, as much as a random pick.
\end{takeaway}

\section{RQ2: The predictive distribution is the second opinion}
\label{sec:using}

\begin{figure}[t]
  \centering
  \includegraphics[width=\columnwidth]{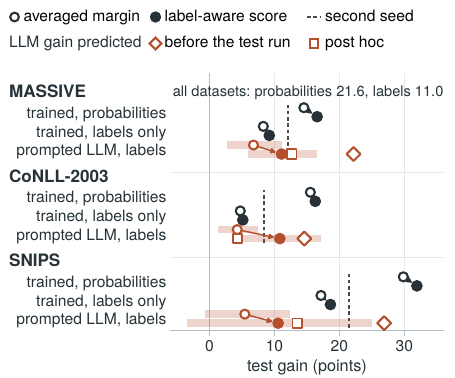}
  \caption{\textbf{Partners help most when they pass probabilities and the score checks the served label.}
Mean test gain with the averaged margin (hollow) and the label-aware score (filled), for trained partners (576 pairings) and a prompted LLM; bands: 95\% intervals; dashed: a second seed.
Diamond: the LLM's gain predicted before its test run; square: a post hoc prediction from label-only partners; top right: trained partners over all datasets.}
  \label{fig:llm}
\end{figure}

\paragraph{Score the label that will be served.}
Selective confidence must evaluate the decision the system will actually serve, not the ensemble's preferred decision.
Plain averaging ignores which label the primary chose.
In Figure~\ref{fig:method}, averaging makes the utterance look safer than the primary alone, although the served label O is wrong.
The label-aware score $\slab$ measures the margin of the served label instead, so the utterance is flagged.
It raises the gain in 94\% of the 576 pairings, by 1.6 points on average [1.2, 2.2] (Appendix~\ref{app:labelaware}).

\paragraph{The label-aware score replicated on a new dataset.}
We fixed $\slab$ and its pass rule before MIT-Restaurant was chosen for the frozen test of \S\ref{sec:prospective}.
We then scored it once there, on every pairing not scored with $\slab$ before.
The test passes: $\slab$ raises the gain by 2.7 points [1.7, 4.0], and 68 of 70 pairings improve (Appendix~\ref{app:slabmitr}).

\paragraph{Hard labels discard half of a partner's value.}
When each trained partner passes only its labels, as one-hot distributions, its gain falls from 21.6 to 11.0 points, a difference of $-10.6$ [$-14.4$, $-7.6$] (Figure~\ref{fig:llm}).

\paragraph{A prompted LLM is a useful label-only partner.}
We prompt Qwen2.5-7B-Instruct \citep{qwen2025} with 8 retrieved training examples.
Its span F1 is lower than that of every trained tagger.
Yet with $\slab$ it matches a second seed on MASSIVE and exceeds it on CoNLL (Figure~\ref{fig:llm}; Appendix Table~\ref{tab:llm}).
On MASSIVE and SNIPS, its gain follows the label-only trained partners, not the probability partners.
So compare candidates of one output kind: probabilities or labels.

\begin{figure}[t]
  \centering
  \includegraphics[width=\columnwidth]{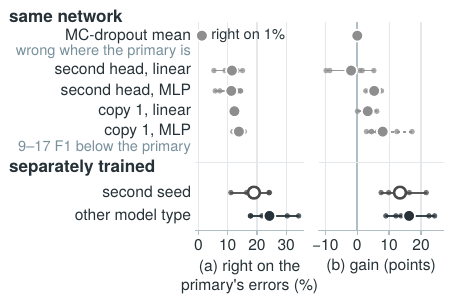}
  \caption{\textbf{For LoRA-tuned echo primaries, separately trained partners beat readouts of the primary network.}
(a) Share of the primary's wrong utterances on which each second opinion is fully right; (b) gain in points.
Small dots: one echo primary on one dataset (mean of 3 seeds); large dots: mean of the six.
Copy~1 is the echo tagger's first reading of its input (Appendix Figure~\ref{fig:echoschem}); probes on it tag 9--17 F1 points below the primary.}
  \label{fig:samenet}
\end{figure}

\paragraph{A second readout is not a second opinion.}
\label{sec:same}
For the LoRA-tuned echo primaries, readouts of the primary network are wrong where the primary is wrong (Figure~\ref{fig:samenet}a).
Their gains follow (Figure~\ref{fig:samenet}b).
The best readout, an MLP probe on the first copy of the input, gains 8.0 points.
A second seed gains 13.4, and a partner of another type 16.3.
MC dropout, stronger dropout and the fully fine-tuned XLM-R are covered in Appendix~\ref{app:same}.

\paragraph{Trained partners help reliably.}
All 9 pre-specified primary--partner tests on MASSIVE and 12 of 18 overall are significant under both the bootstrap and a three-seed $t$ interval.
Partners also help on top of a richer single-model score \citep{vazhentsev2022} (Appendix~\ref{app:partners}).

\begin{takeaway}
\textbf{RQ2 answer:} the predictive distribution is the second opinion. Pass probabilities: they give twice the gain of labels. Score their average at the served label: +1.6 points in 94\% of pairings, and the frozen test on a new dataset replicated it. Spend the second forward pass on a separately trained model, not on a readout of a LoRA-tuned primary.
\end{takeaway}

\section{RQ3: Complementarity generalises across data, languages and budgets}
\label{sec:rq3}

\subsection{A frozen test on a new dataset}
\label{sec:prospective}

We froze the rule in a hashed protocol, together with five tagger types, their recipes, three success criteria and a rule for choosing the dataset.
That rule then chose MIT-Restaurant \citep{liu2013asgard} from licence and download metadata alone (Appendix~\ref{app:prospective}; timeline in Figure~\ref{fig:timeline}).
We trained the five types with 3 seeds each and scored the test set once.
All three pre-declared criteria hold (Figure~\ref{fig:central}d,e).
The check reaches $\rho=0.40$ [0.08, 0.59], against 0.04 for held-out span F1.
Its top-ranked partner misses less than half as much of the best gain as a random partner.
The frozen label-aware test of \S\ref{sec:using} also passed here.

\subsection{Across 51 locales}
\label{sec:shift}

\begin{figure*}[t]
  \centering
  \includegraphics[width=\textwidth]{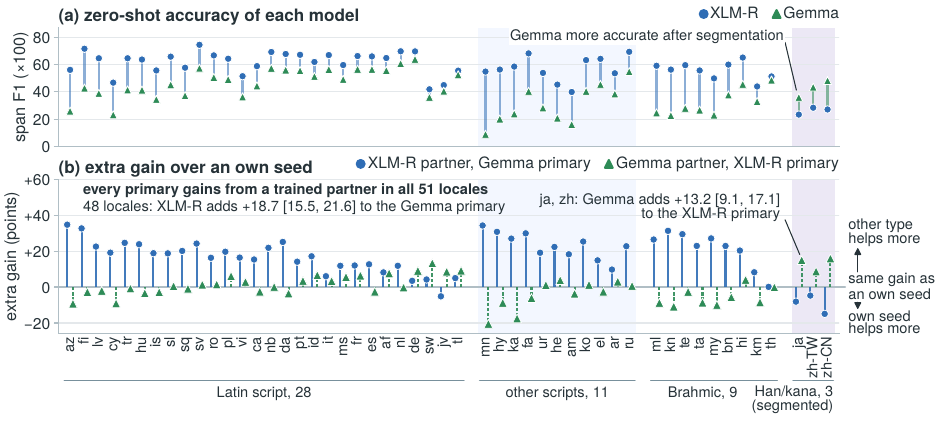}
  \caption{\textbf{In 50 of 51 locales, a partner of the more accurate model type adds more than an own seed} (exception: jv-ID).
en-US taggers applied zero-shot to the other MASSIVE locales, grouped by writing system (test-set analysis; Japanese and Chinese after word segmentation).
(a) Span F1 of XLM-R (circles) and Gemma (triangles).
(b) Gain from the other-type partner minus gain from an own seed (mean of 3 seeds; solid: XLM-R partner, dashed: Gemma partner), with pooled estimates and 95\% intervals.}
  \label{fig:crossling}
\end{figure*}

We apply the en-US taggers zero-shot to the 51 other locales of MASSIVE v1.1 \citep{fitzgerald2023}, with temperatures and combiners fit on en-US.

\paragraph{Every primary has a helpful partner in every locale.}
For every primary in all 51 locales, at least one separately trained partner improves selective ranking (Appendix Table~\ref{tab:locales}).

\paragraph{Competence, not architecture, decides the preferred partner.}
When the stronger model type changes across languages, the preferred partner changes with it.
In the 48 locales other than Japanese and Chinese, XLM-R is far more accurate than either decoder (Figure~\ref{fig:crossling}a).
It is also the better partner: a Gemma primary gains 18.7 points [15.5, 21.6] more from XLM-R than from its own seed.
An XLM-R primary gains no more from Gemma than from its own seed.

\paragraph{Japanese and Chinese reverse the direction.}
After word segmentation (Appendix~\ref{app:cjk}), Gemma is the more accurate model (42.6 F1, against 26.3 for XLM-R).
It then becomes the better partner: the XLM-R primary gains 13.2 points [9.1, 17.1] more from Gemma than from its own seed (Figure~\ref{fig:crossling}b).
This holds in every Japanese and Chinese locale under every calibration condition, and gold segmentation agrees (Appendix~\ref{app:cjk}).

\paragraph{The two regimes agree.}
When candidates have similar standalone quality, residual-error complementarity distinguishes useful partners.
Under large cross-lingual quality gaps, overall competence and residual complementarity become strongly aligned, so either guides the choice (Appendix~\ref{app:mlexp}).

\subsection{Serving cost}
\label{sec:systems}

\begin{figure}[t]
  \centering
  \includegraphics[width=\columnwidth]{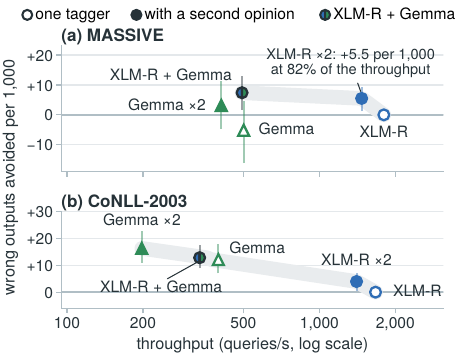}
  \caption{\textbf{A second XLM-R seed avoids 5.5 wrong outputs per 1,000 at 82\% of XLM-R's throughput; on CoNLL, serving the more accurate Gemma avoids 12.2.}
Wrong outputs avoided among the 80\% most confident per 1,000 queries, against XLM-R alone (bars: 95\% intervals), and throughput on one H100 (batch 32).
Hollow: one tagger; filled: with a second opinion ($\times2$: a second seed); split marker: XLM-R with Gemma; grey band: not beaten on both axes.}
  \label{fig:pareto}
\end{figure}

At 80\% coverage, we count wrong outputs avoided per 1,000 queries against XLM-R alone; a pair serves the averaged labels (Figure~\ref{fig:pareto}).
On CoNLL, where Gemma is the more accurate tagger, serving Gemma alone avoids 12.2 (Appendix~\ref{app:systems}).

On MASSIVE, a second XLM-R seed avoids 5.5 [1.2, 9.4] at 82\% of XLM-R's throughput.
Adding Gemma instead avoids 7.4, at about a quarter of the throughput.

\paragraph{Fix the serving budget, then choose the most complementary model inside it.}
The best partner of every encoder primary is a decoder.
When the budget rules out decoders, the check nearly halves the miss rate among encoder partners: 15\%, against 28\% at random (Appendix Table~\ref{tab:cost}).

\begin{takeaway}
\textbf{RQ3 answer:} complementarity generalises. Both frozen tests passed on a new dataset. In all 51 locales, every primary has a helpful trained partner. Inside a cost class, the check misses 15\% of the class's best gain, against 28\% at random.
\end{takeaway}

\section{Related work}
\label{sec:related}

\paragraph{Complementary errors and diversity.}
The complementary rate of \citet{brill1998} counts where a second tagger is right and the first is wrong; tagger combination goes back to part-of-speech tagging and CoNLL-2003 NER \citep{vanhalteren2001,florian2003,wu2003}.
Pairwise diversity measures such as the double-fault rate \citep{giacinto2001} track ensemble accuracy only weakly \citep{kuncheva2003,brown2005}.
Ensemble selection picks members by held-out gain \citep{caruana2004}.
\emph{We} turn complementarity into a partner-selection rule, show that its error-conditioned form beats accuracy and symmetric diversity (Table~\ref{tab:policies}), and test it prospectively.

\paragraph{Ensembles, uncertainty and selective prediction.}
Independent ensembles give better uncertainty than single models \citep{lakshminarayanan2017,ovadia2019,fort2019}, and members trained more differently make less correlated errors \citep{gontijolopes2022}; cheaper variants share weights or distil the ensemble \citep{wen2020,havasi2021,malinin2020}.
Tagger confidence has been improved with richer single-model signals \citep{vazhentsev2022,shelmanov2021}, calibration \citep{guo2017,desai2020,jagannatha2020} and selective prediction \citep{kamath2020,varshney2022sel}.
\emph{These} works ask whether a second model helps, or improve one tagger's confidence.
We ask which second model to deploy, and our partners help beyond the richer baseline.

\paragraph{Decoders and serving.}
Repeating the input lets a causal decoder's second copy attend to the whole input, for embeddings \citep{springer2025echo} and tagging \citep{kukic2026,ewais2026}; learning to defer trains a model to hand inputs to an expert \citep{madras2018,mozannar2020}, and cascades and routers decide when to call a larger model \citep{wang2022committees,chen2023frugal}.
These works decide \emph{when} to defer or escalate; \emph{we} decide \emph{which} model to pair with the primary, within a cost class (extended discussion: Appendix~\ref{app:related}).

\section{Conclusion}
\label{sec:conclusion}

Global accuracy tells you how good a model is; conditional correctness tells you how useful it is to this primary.
Across tagger types, datasets, languages and serving budgets, useful partners correct the primary's residual errors and expose this through their predictive distributions.
The rule that finds them, and the label-aware score, replicated on a dataset chosen after both were frozen.
This turns second-opinion selection from a generic ensembling choice into a directional, measurable deployment decision.
For practitioners:
\begin{enumerate}[leftmargin=*,itemsep=0pt,topsep=2pt]
\item Pick the partner most often right on the primary's errors; about 100 labelled errors suffice, and gain on dev or a prior works as well. Never default to the own seed.
\item Pass probabilities, not labels.
\item Score their average at the label the primary serves.
\item Prefer a separately trained partner to readouts of the primary, and compare candidates of one output kind and cost class.
\end{enumerate}

\section*{Limitations}

\paragraph{One frozen dataset.}
We formulated the selection rule and the label-aware score on our three datasets, including their test sets, and then froze both. The prospective tests cover one English slot-filling dataset and five tagger types.
All three pre-declared criteria of the selection test held; the comparison with span F1 was declared on point estimates (difference 0.36 [$-0.09$, 0.62]), and in-sample its interval excludes zero (0.29 [0.01, 0.54]).
One MIT-Restaurant primary had been scored with the label-aware score before its test, so the test left it out (Appendix~\ref{app:slabmitr}).
Future work: frozen tests on MultiCoNER and MIT-Movie with pre-declared competitors.

\paragraph{Equivalent measures.}
Gain on dev, the expected margin drop, word accuracy and priors from other datasets perform on par with the check; we treat them as measurements of one principle and recommend the cheapest available.
With fewer than about 100 labelled errors of the primary, or with an unfamiliar label-only partner in the pool, the check should be combined with a prior (Appendix~\ref{app:priorcheck}).

\paragraph{Where the edge over F1 lies.}
The check's lead over span F1 comes from encoder primaries and from never picking the own seed.
For decoder primaries, the check, span F1, word and utterance accuracy and gain on dev perform alike (18--21\% of the best gain missed, against 28\% for a random partner and 42\% for the own seed).
Intervals resample 8 model types (5 on MIT-Restaurant).

\paragraph{Effect sizes and cost.}
Gains are reported relative to the margin: in-language partners remove 6--32 points of the margin's \eaurc{} on MASSIVE and CoNLL ($1$--$9\times10^{-3}$ absolute), and the XLM-R pairs of \S\ref{sec:systems} avoid 4.0--12.8 wrong outputs per 1,000 queries at 80\% coverage.
The best partner of an encoder primary is a decoder, which lowers XLM-R's throughput from 1,801 to 494 queries per second; inside the encoder class the check still nearly halves the miss rate.

\paragraph{Scope.}
In-language data are English.
Cross-lingual results apply English-trained taggers zero-shot and are test-set analyses.
The Japanese--Chinese reversal rests on two languages and automatic segmentation (gold segmentation agrees), and one primary (Qwen-3B on zh-CN) breaks the pattern.
Abstention is per utterance with argmax decoding, and latency comes from one eager PyTorch server on one GPU type.
Future work: span-level abstention, constrained decoding, jointly trained readout heads, LoRA ensembles and more prompted LLMs.

\paragraph{Protocol records.}
Protocols are plain-text files, most hashed at writing time, without an external timestamp.
Three components were specified after earlier results were seen: the rank-mapped combiner, zero-shot calibration as the main cross-lingual condition, and the label-aware score.
The last was specified after seeing the three original datasets, then confirmed once on MIT-Restaurant pairings never scored with it.
The prior-plus-check rule missed its pre-specified 400-utterance criterion, and on CoNLL, dev served as the held-out slice and also selected checkpoints.
Appendix~\ref{app:prereg} lists every protocol, deviation and outcome.

\section*{Ethics Statement}

We use public datasets under their terms: CoNLL-2003 (research use), MASSIVE (CC BY 4.0), SNIPS (CC0 1.0), CrossNER, WNUT-17, and MIT-Restaurant, which has no formal licence but is distributed by its creators to support research in speech and language processing.
We use public model checkpoints; the Qwen2.5 licences carry use restrictions, which our use respects.
No new data was collected and nothing was annotated.
Better failure ranking lets systems pass uncertain outputs to people.
Its quality varies strongly across locales in our results (Appendix Table~\ref{tab:locales}), so deferral thresholds tuned on English may not protect users of other languages equally.
Compute is reported in Appendix~\ref{app:compute}, and the reproducibility checklist in Appendix~\ref{app:repro}.
AI assistants were used for coding, analysis and editing; the authors take full responsibility for the content.

\appendix
\renewcommand{\topfraction}{0.95}\renewcommand{\dbltopfraction}{0.95}\renewcommand{\bottomfraction}{0.8}
\renewcommand{\textfraction}{0.05}\renewcommand{\floatpagefraction}{0.95}\renewcommand{\dblfloatpagefraction}{0.95}
\setcounter{topnumber}{4}\setcounter{dbltopnumber}{4}\setcounter{bottomnumber}{3}\setcounter{totalnumber}{8}

\section{Guide to the appendix}
\label{app:overview}

Table~\ref{tab:guide} maps each claim of the main text to its evidence.

\begin{table*}[tp]
\centering\small\setlength{\tabcolsep}{4pt}
\begin{tabular}{>{\raggedright\arraybackslash}p{0.40\textwidth}>{\raggedright\arraybackslash}p{0.36\textwidth}>{\raggedright\arraybackslash}p{0.17\textwidth}}
\toprule
claim (main text) & evidence (appendix) & main-text picture\\
\midrule
Held-out accuracy on the primary's errors ranks partners by test gain; it beats F1; double fault is identical (\S\ref{sec:curve}) & Tables~\ref{tab:predfull}, \ref{tab:preddiff}, \ref{tab:g2}; Appendix~\ref{app:oos} & Figure~\ref{fig:central}; Table~\ref{tab:policies}\\
Global performance is not the selection objective; every measure of complementarity works; label budget; cost classes (\S\ref{sec:curve}, \S\ref{sec:systems}) & Appendix~\ref{app:policies}; Tables~\ref{tab:policiesfull}, \ref{tab:fourdata}, \ref{tab:budget}, \ref{tab:cost}; Figures~\ref{fig:labelbudget}, \ref{fig:costpolicies}, \ref{fig:gainmat} & Table~\ref{tab:policies}\\
Where the rule holds: encoder vs decoder primaries, own seed, new types & Tables~\ref{tab:g2}, \ref{tab:mixedpools}, \ref{tab:anomalies} & Table~\ref{tab:policies}\\
The frozen rule replicated on MIT-Restaurant (\S\ref{sec:prospective}) & Appendix~\ref{app:prospective}; Tables~\ref{tab:prosp_models}, \ref{tab:prosp_pred}; Figure~\ref{fig:timeline} & Figure~\ref{fig:central}d,e\\
Partner prior and labelled check; new partner types & Appendix~\ref{app:priorcheck}; Figure~\ref{fig:priorcheck} & Table~\ref{tab:policies}\\
Label-aware score (\S\ref{sec:using}) & Appendix~\ref{app:labelaware}; Figure~\ref{fig:labelaware}; Tables~\ref{tab:labelaware}, \ref{tab:readoutsci} & Figure~\ref{fig:method}\\
The frozen label-aware score replicated on MIT-Restaurant (\S\ref{sec:using}) & Appendix~\ref{app:slabmitr}; Table~\ref{tab:slabmitr} & --\\
Prompted LLM as a label-only partner (\S\ref{sec:using}) & Appendix~\ref{app:llm}; Table~\ref{tab:llm} & Figure~\ref{fig:llm}\\
Same-network readouts (\S\ref{sec:same}) & Tables~\ref{tab:readoutsci}, \ref{tab:equiv}, \ref{tab:mcstrong}, \ref{tab:mcrank}, \ref{tab:singlecopyh2h}; Figures~\ref{fig:curve}, \ref{fig:mcdropout}; Appendix~\ref{app:curve} & Figures~\ref{fig:echoschem}, \ref{fig:samenet}\\
Trained partners help; one-property partners (\S\ref{sec:same}) & Tables~\ref{tab:secC1}, \ref{tab:bfamily}, \ref{tab:combfree}, \ref{tab:power}, \ref{tab:afamily} & Figure~\ref{fig:samenet}\\
Larger models and GLiNER & Tables~\ref{tab:abe}, \ref{tab:gliner}; Figure~\ref{fig:newpartners} & --\\
Worked examples, risk--coverage curves, gains of every pair of types & Appendix~\ref{app:visuals}; Figures~\ref{fig:riskcov}--\ref{fig:gainmat} & Figure~\ref{fig:method}\\
Cross-lingual results (\S\ref{sec:shift}) & Tables~\ref{tab:xloverlap}, \ref{tab:xlingsens}, \ref{tab:xlmain}, \ref{tab:cjk}, \ref{tab:locales}, \ref{tab:mfamily}; Figure~\ref{fig:localegain} & Figure~\ref{fig:crossling}\\
Serving (\S\ref{sec:systems}) & Table~\ref{tab:serving}; Figure~\ref{fig:load} & Figure~\ref{fig:pareto}\\
Evaluation choices that reversed conclusions (\S\ref{sec:setup}) & Appendix~\ref{app:lessons}; Tables~\ref{tab:lessons}, \ref{tab:noise} & --\\
\bottomrule
\end{tabular}
\caption{Where to find the evidence for each claim. In every table, a positive gain means the second opinion helps. Points: percent of the margin-only \eaurc{}; most appendix tables give absolute \eaurc{} $\times10^{-3}$, as marked.}
\label{tab:guide}
\end{table*}

\section{Worked examples and further views}
\label{app:visuals}

Gains are in points, as in the main text, and use the averaged-margin score of the pre-specified tests unless a caption says otherwise.

\subsection{Risk--coverage curves}
\label{app:riskcov}

\begin{figure*}[t]
  \centering
  \includegraphics[width=\textwidth]{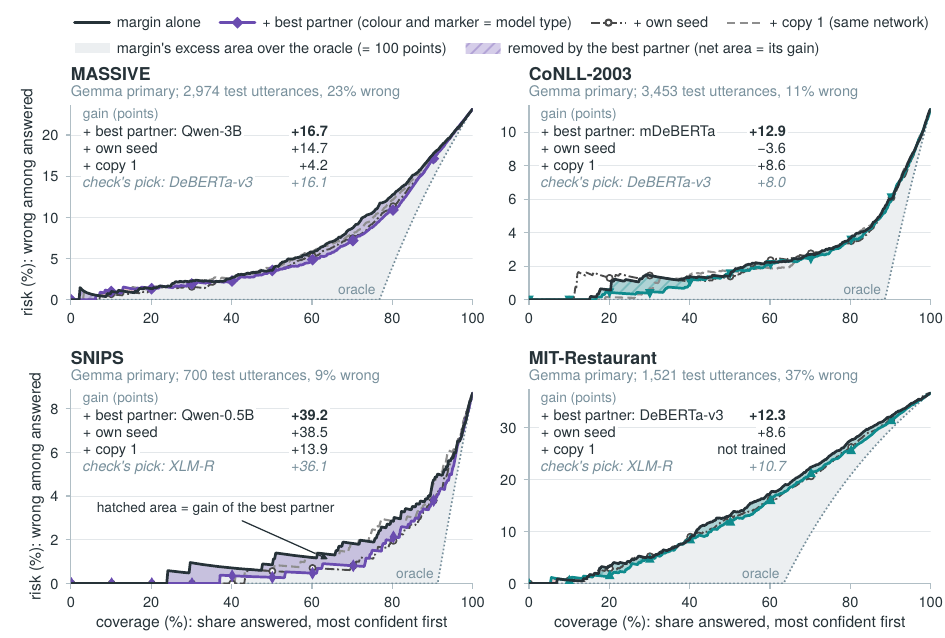}
  \caption{\textbf{Risk--coverage curves: the area between two curves is the gain.} One primary per dataset (the Gemma echo tagger, seed~0, fixed in advance). Answering the most confident utterances first, each curve gives the share of wrong answers among those answered. The grey region is the excess area of the primary's margin over the oracle, which answers every right utterance first; this area is the margin's eAURC, so it is worth 100 points. The hatched band is the area that the best partner (best by test gain) removes; minus the thin stretches where its curve lies above the margin's, it equals that partner's gain. Numbers in each panel are gains in points; copy~1 was not trained on MIT-Restaurant. Averaged-margin score, as in the pre-specified tests.}
  \label{fig:riskcov}
\end{figure*}

Figure~\ref{fig:riskcov} draws the risk--coverage curves behind the eAURC for one primary per dataset (the Gemma echo tagger, seed~0, fixed before plotting).
On MASSIVE, SNIPS and MIT-Restaurant, 79--91\% of that gain comes from coverage above 40\%; on MASSIVE at 80\% coverage, for example, the share of wrong answers falls from 12.7\% to 10.8\% with Qwen-3B as the partner.
On CoNLL, two thirds of the gain comes from coverage below 40\%, where the partner moves a few confident errors out of the most confident answers.
The partner that the held-out check picks misses 0.6--3.1 points of the best gain on three datasets; on CoNLL it picks DeBERTa-v3 (8.0 points) instead of mDeBERTa (12.9).

\subsection{Worked examples}
\label{app:examples}

\begin{figure*}[p]
  \centering
  \includegraphics[width=\textwidth]{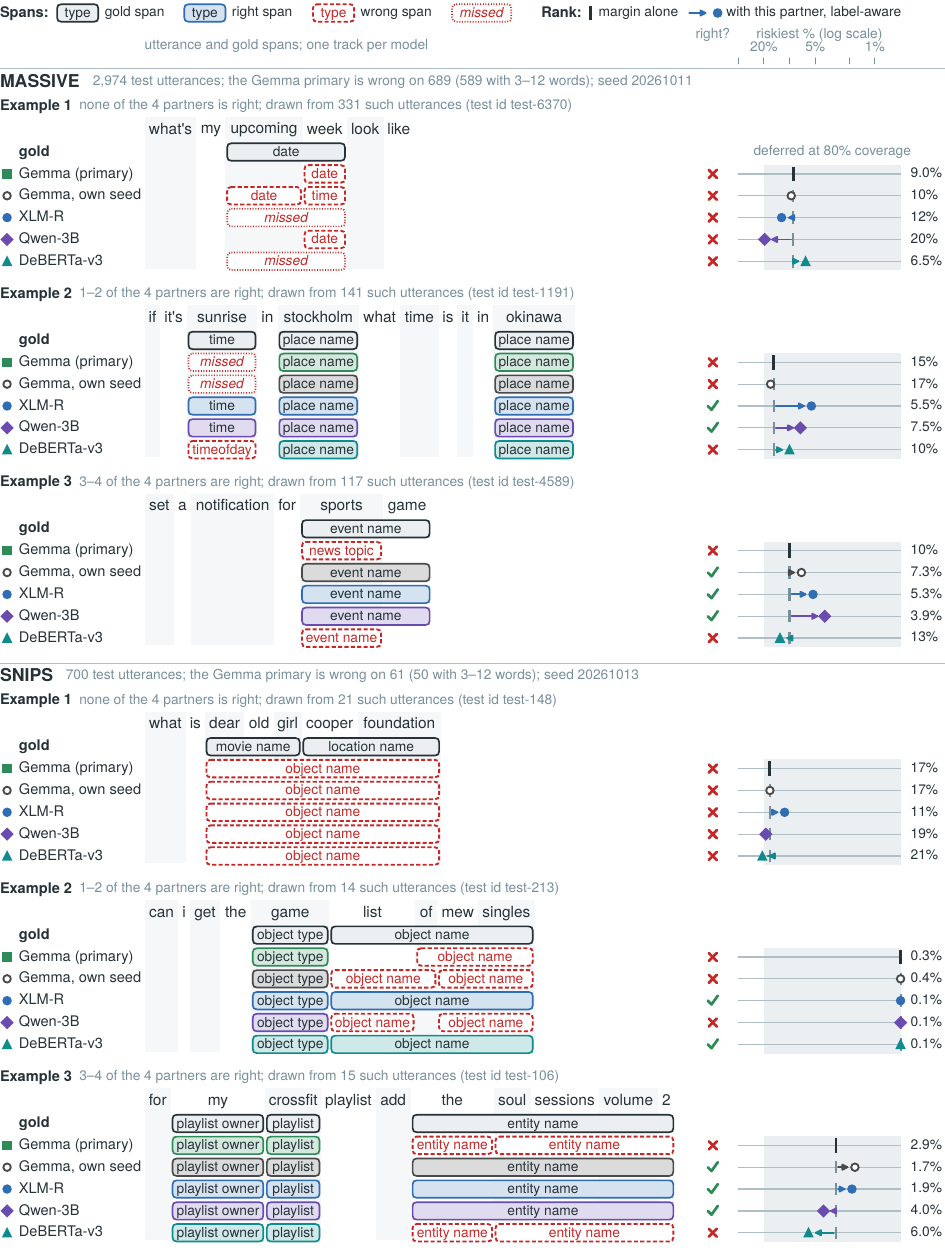}
  \caption{\textbf{Worked examples on MASSIVE and SNIPS: partners that are right where the primary is wrong usually move the utterance towards deferral.} Per dataset, three test utterances on which the Gemma primary is wrong, drawn with a fixed seed, one from each stratum of how many of the four partners are fully right; none was picked by hand. Chips are predicted spans: solid if exactly right, dashed red if wrong, dotted red outline if a gold span was missed. The strip ranks the utterance among the dataset's test utterances (log scale; ranks riskier than 0.5\% sit at the right end): the bar is its rank by the primary's margin alone, and the arrow shows where it moves once that partner is added with the label-aware score. The shaded end is deferred at 80\% coverage.}
  \label{fig:ex-massive-snips}
\end{figure*}

\begin{figure*}[p]
  \centering
  \includegraphics[width=\textwidth]{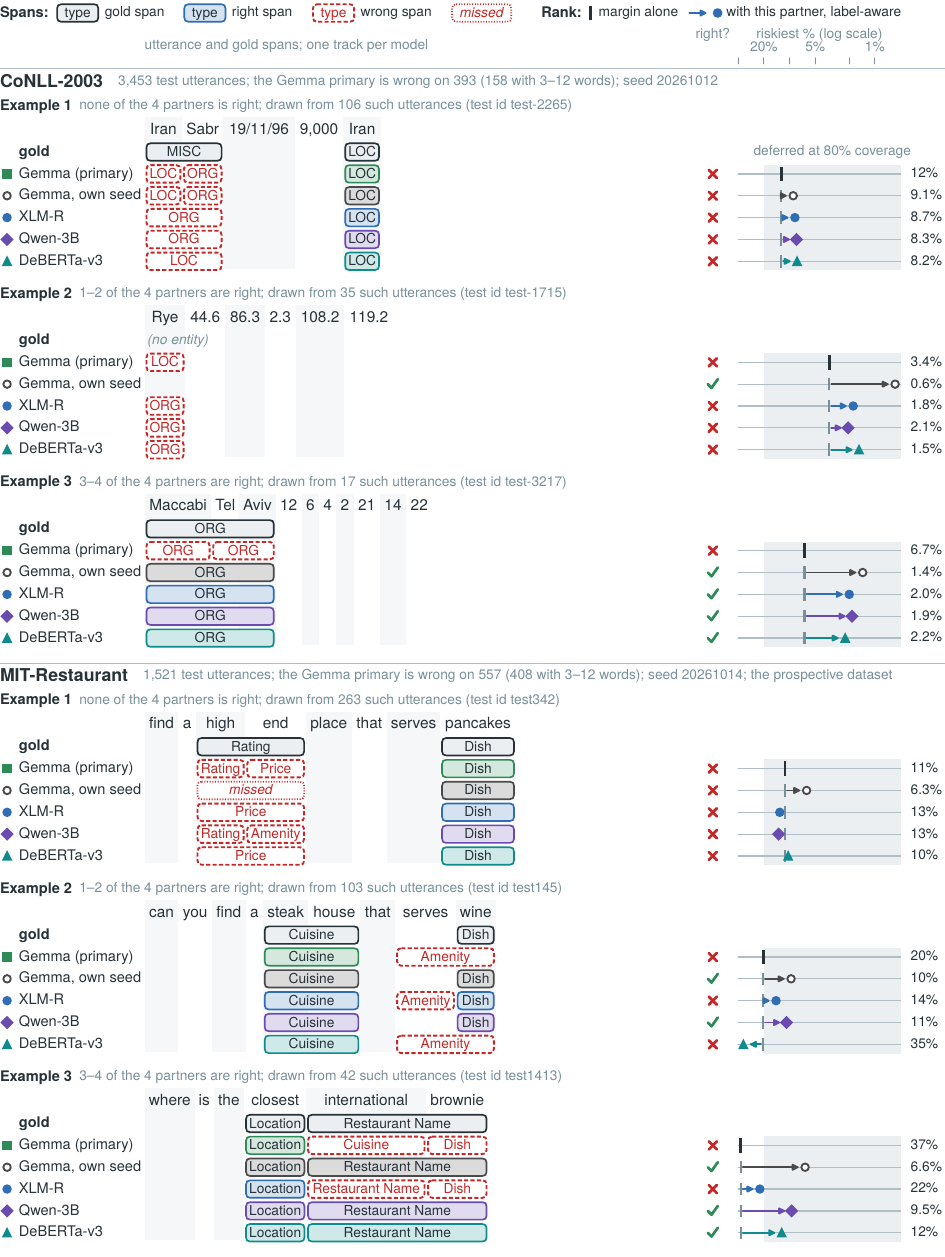}
  \caption{\textbf{Worked examples on CoNLL-2003 and MIT-Restaurant: the same pattern holds.} Drawn and shown as in Figure~\ref{fig:ex-massive-snips}. All three CoNLL examples are table rows from news wires: a rice-shipping list (vessel, date, tonnes, destination), a grain-statistics table (``Rye'' is the grain, so the gold annotation has no entity) and a football league table. This MIT-Restaurant primary was left out of the frozen label-aware test (Appendix~\ref{app:slabmitr}), so its ranks here are descriptive.}
  \label{fig:ex-conll-mitr}
\end{figure*}

Figures~\ref{fig:ex-massive-snips} and~\ref{fig:ex-conll-mitr} show test utterances of 3--12 words on which the Gemma primary (seed~0) is wrong, one for each number of the four partners (Gemma's second seed, XLM-R, Qwen-3B and DeBERTa-v3) that tag them fully right.
Beyond the drawn examples, over all of the primary's wrong test utterances, a right partner shrinks the utterance's risk percentile (the share of test utterances ranked at least as risky) by a median factor of 1.6--2.8 and moves it towards the risky end in 79--95\% of cases, while a wrong partner leaves the median unchanged (factor 0.92--1.06) and moves it either way.

\subsection{Which partner helps which primary}
\label{app:gainmat}

\begin{figure*}[tp]
  \centering
  \includegraphics[width=\textwidth]{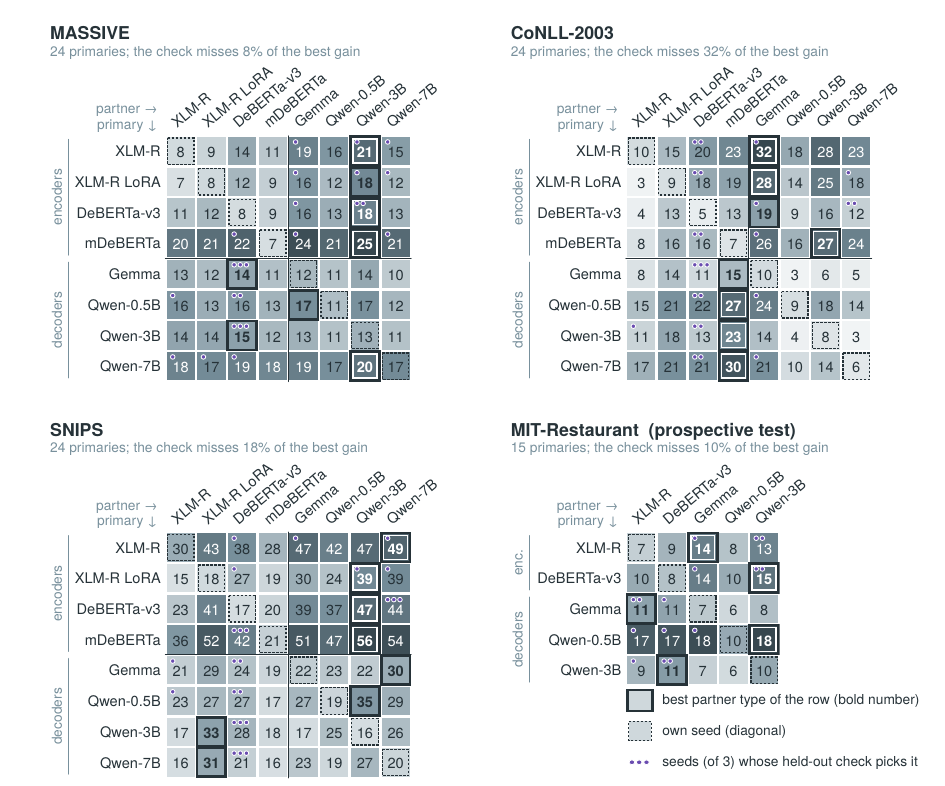}
  \caption{\textbf{The best partner type usually has the other architecture (24 of 29 rows).} Each cell is the mean test gain (points) of a primary of the row type paired with a partner of the column type, over three primary seeds; the diagonal pairs it with another seed of the same type. Purple dots count the seeds whose held-out check (highest held-out accuracy on the primary's errors) picks that partner. Darker means a larger gain (scaled per dataset); no cell is a loss on average.}
  \label{fig:gainmat}
\end{figure*}

Figure~\ref{fig:gainmat} breaks the gains down by primary type and partner type.
For every encoder primary type on every dataset (14 of 14 rows), the best partner type is a decoder; for decoder primaries, the best is an encoder in 10 of 15 rows.
The own seed is the weakest partner type in 14 of 29 rows and one of the two weakest in 22.
The held-out check picks the row's best type for only 21 of 87 primaries, but its picks gain almost as much: it misses 8--18\% of the best gain on MASSIVE, SNIPS and MIT-Restaurant.
CoNLL is the exception (32\%): there the check picks DeBERTa-v3 for 15 of 24 primaries, while the best partner type is mDeBERTa or Gemma in 7 of 8 rows.

\paragraph{Directionality.} Over the type pairs of each dataset, the gain of B for primary A and the gain of A for primary B are negatively correlated: Spearman $-0.47$ (MASSIVE), $-0.22$ (CoNLL), $-0.61$ (SNIPS) and $-0.44$ (MIT-Restaurant).

\subsection{Timeline of the prospective test}
\label{app:timeline}

Figure~\ref{fig:timeline} puts the prospective test of \S\ref{sec:prospective} (details in Appendix~\ref{app:prospective}) on a clock.
The protocol, with the decision rule, the five tagger types and their recipes, and the three success criteria, was hashed at 04:11 UTC, before any candidate dataset was looked at.
Training took 24 of the 33 minutes between the freeze and the decision, and each tagger wrote its test outputs once, after training; test gains were computed only after all 15 taggers had finished.
No deviation from the protocol was logged; the one restart (three runs that could not find their weights) came before any test gain existed.
The label-aware score and its pass rule had been hashed at 04:06, before this protocol (Appendix~\ref{app:slabmitr}).

\begin{figure*}[tp]
  \centering
  \includegraphics[width=\textwidth]{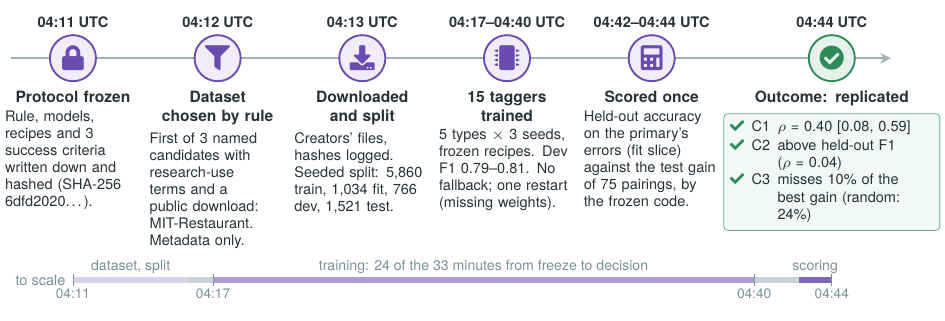}
  \caption{\textbf{The prospective test ran in a fixed order, and the protocol was not changed after the freeze.} Times (UTC) come from the protocol hash file, the dataset-choice log, the download stamp, the training logs and the result-file times. The bottom bar is drawn to scale.}
  \label{fig:timeline}
\end{figure*}

\section{Protocols, tests and deviations}
\label{app:prereg}

\subsection{How the protocols were kept}
Before each experiment we wrote a plain-text protocol in its folder: hypotheses, the exact tests, the decision rules, and what would count as a failure.
For most protocols we stored the file's SHA-256 hash at the time of writing; later changes were appended as dated addenda.
There is no external registry and no trusted timestamp, so the record shows internal consistency, not independent proof of timing.
Most experiments re-used trained checkpoints or re-analysed saved outputs, so most protocols were written knowing earlier results.
This matters most for the test-set curve, the cross-lingual re-analysis, the in-language encoder/decoder asymmetry and the label-aware score, whose effects we had seen before their protocols were written; and for the held-out test of \S\ref{sec:curve}, written after the test gains were known but before any held-out value was computed.
We call all of these \emph{pre-specified, not blind}.
Four tests were fixed before any of their outcomes existed: the prospective MIT-Restaurant test, the label-aware test on the MIT-Restaurant pairings not scored with it before (Appendix~\ref{app:slabmitr}), the LLM's predicted gain, and the choice of the LLM's prompt (fixed on training utterances only).
The protocol files, their hashes and the file times are released with the code.

\subsection{What the protocols specified, and what happened}
\begin{description}[leftmargin=0pt,itemsep=2pt,font=\normalfont\bfseries]
\item[Copy-1 disagreement.] Disagreement between copy 1 and copy 2 adds information beyond the margin, in at least 2 of 3 seeds. Failed on CoNLL (0 of 9 cells), MASSIVE (0 of 3) and SNIPS (1 of 3), and under domain shift in all 162 (domain, seed, probe seed) cells, where the gain's interval lies above 0 in 3 cells and below 0 in 2.
\item[Partner on MASSIVE.] A partner of another model type beats the margin and a learned router in at least 2 of 3 seeds. Passed.
\item[Calibration law.] Across locales, the gain of a second model grows with calibration error, beyond the error rate. Failed (Appendix~\ref{app:mlexp}).
\item[One-property partners.] An axis ``matters'' if its partner beats the own seed after Holm on at least 2 of 3 datasets. None did; the rule had low power (Table~\ref{tab:power}).
\item[Curve analysis.] All second opinions lie on one curve of gain against relative F1. Failed. Accuracy on the primary's errors was named as a secondary covariate without a reading rule; its test-set analysis is exploratory.
\item[Held-out test (D1--D4).] D1 (pooled within-primary Spearman positive with an interval excluding 0, above held-out F1 on at least 2 of 3 datasets): holds, 3 of 3. D2 (beats error correlation, $\kappa$, Q and double fault): beats all but double fault, which is identical by construction. D3 (a label-free proxy recovers at least half): fails. D4 (the partner's AUROC adds out-of-sample $R^2$): fails. The 90\%-of-full-slice sample-size rule is not met.
\item[Prospective test (C1--C3).] On a new dataset chosen by a fixed rule: C1 as D1's first part; C2 above held-out F1 by point estimate; C3 the top-ranked partner loses less than a random one. All three hold.
\item[Label-aware score (D-LA).] Recommend $\slab$ if its pooled gain over $\savg$ has a two-way interval above 0 and D1 computed with $\slab$ gains stays above 0.424, that is, at most 0.05 below its value with $\savg$ (0.474). Met. Applied once more to MIT-Restaurant: both conditions hold (Appendix~\ref{app:slabmitr}).
\item[Prompted LLM (B1--B4).] B1 the LLM helps; B2 its gain matches the hashed prediction (residual interval includes 0); B3 ranking with the LLM among the candidates; B4 label-aware vs averaged margin. B1 holds on MASSIVE and CoNLL; B2 fails on MASSIVE and SNIPS.
\item[Prior plus check (D-PC).] Prior plus check beats the check alone at 400 labelled utterances (interval below 0) and is no worse on the full slice. Not met.
\end{description}

\subsection{Confirmatory families and multiplicity}
Families A (one-property partners and averaging controls, Table~\ref{tab:afamily}), B (partners on top of richer baselines, Table~\ref{tab:bfamily}) and M (cross-lingual, Table~\ref{tab:mfamily}) were declared before the corresponding models were trained or scored. The secondary family C1 (18 primary--partner tests, Table~\ref{tab:secC1}) and the 12 head-to-heads are also pre-specified. Holm's correction \citep{holm1979} is applied within each family.
\emph{Multiplicity.} We correct within families, not across the paper. Across the paper we ran about a dozen families and several decision rules; we report every one, including failures (D3, D4, the sample-size rule, B2, D-PC, the calibration law and the one-property rule), and we do not claim significance for any result outside a family or decision rule.

\begin{table*}[tp]\centering\scriptsize\setlength{\tabcolsep}{3.5pt}
\begin{tabular}{lllrrrrrrl}\toprule
id & question & dataset & rank gain & 95\% CI & Holm $p$ & raw gain & 95\% CI & Holm $p$ & per seed (rank)\\\midrule
A1 & pretraining & MASSIVE & +1.1 & [-0.7, +2.8] & 1.000 & +1.1 & [-1.0, +3.2] & 1.000 & +1.8 / +1.7 / -0.2\\
A2 & pretraining & CoNLL & +2.6 & [+1.3, +4.0] & 0.002$^{*}$ & +2.4 & [+1.1, +3.7] & 0.002$^{*}$ & +3.5 / +2.8 / +1.5\\
A3 & pretraining & SNIPS & +1.0 & [-0.9, +3.3] & 1.000 & +1.1 & [-1.2, +3.7] & 1.000 & +0.3 / +2.1 / +0.7\\
A4 & architecture & MASSIVE & +3.4 & [+1.7, +5.2] & 0.002$^{*}$ & +3.6 & [+1.6, +5.6] & 0.008$^{*}$ & +4.0 / +2.4 / +3.8\\
A5 & architecture & CoNLL & +2.3 & [+0.1, +5.1] & 0.420 & +2.3 & [-0.1, +5.3] & 0.758 & +0.9 / +5.1 / +0.8\\
A6 & architecture & SNIPS & +1.7 & [-0.6, +4.1] & 1.000 & +2.0 & [-0.5, +4.6] & 1.000 & +2.9 / +2.5 / -0.2\\
A7 & recipe & MASSIVE & +0.5 & [-0.8, +1.8] & 1.000 & +0.6 & [-0.8, +2.1] & 1.000 & +0.1 / +0.8 / +0.5\\
A8 & recipe & CoNLL & +1.4 & [+0.4, +2.6] & 0.042$^{*}$ & +1.6 & [+0.6, +2.7] & 0.003$^{*}$ & +1.4 / +2.1 / +0.8\\
A9 & recipe & SNIPS & +1.9 & [-0.3, +4.6] & 0.892 & +2.2 & [-0.1, +5.2] & 0.741 & +3.1 / +2.5 / +0.2\\
\addlinespace[2pt]
A10 & copy-1 avg & MASSIVE & +1.3 & [+0.0, +2.5] & 0.535 & -1.5 & [-4.0, +0.9] & 1.000 & +1.6 / +0.5 / +1.8\\
A11 & MC-dropout avg & MASSIVE & +0.0 & [+0.0, +0.1] & 1.000 & +0.0 & [-0.1, +0.1] & 1.000 & +0.0 / +0.0 / +0.0\\
A12 & copy-1 avg & CoNLL & +0.5 & [-0.5, +1.5] & 1.000 & -2.1 & [-3.8, -0.4] & 0.158 & +0.2 / +1.0 / +0.2\\
A13 & MC-dropout avg & CoNLL & +0.1 & [+0.0, +0.2] & 1.000 & +0.1 & [+0.0, +0.2] & 1.000 & +0.1 / +0.1 / +0.0\\
A14 & copy-1 avg & SNIPS & +0.6 & [-0.5, +2.0] & 1.000 & -0.5 & [-2.2, +1.9] & 1.000 & +1.3 / -0.3 / +0.8\\
A15 & MC-dropout avg & SNIPS & +0.0 & [-0.1, +0.1] & 1.000 & +0.0 & [-0.2, +0.2] & 1.000 & +0.0 / -0.1 / +0.0\\
\bottomrule\end{tabular}
\caption{Confirmatory family A (A: one-property partners and averaging controls; 15 tests, Holm within family; $^{*}$ = rejected at 0.05). Gains are $\text{eAURC}$ reductions in $\times10^{-3}$ \eaurc{}, positive = helps. A1--A9: XLM-R primary, gain of partner $B$ minus gain of an own-seed partner, for a partner $B$ that differs in pretraining (DeBERTa-v3-large on CoNLL and SNIPS, mDeBERTa-v3-base on MASSIVE), architecture at matched size (Qwen2.5-0.5B echo) or recipe (XLM-R with LoRA). A10--A15: Gemma primary, gain over the margin of the copy-1 average and the MC-dropout mean. rank = rank-mapped combiner (post hoc deviation); raw = pre-specified logistic. Hierarchical bootstrap with separate item draws per seed, 10k replicates. A8 under rank-ECDF is Monte-Carlo fragile (Holm $p$ 0.042 here, 0.060 in the re-computation's replicates).}\label{tab:afamily}\end{table*}

\begin{table*}[tp]\centering\scriptsize\setlength{\tabcolsep}{3.5pt}
\begin{tabular}{llllrrrrc}\toprule
id & dataset & primary & contrast & raw gain & 95\% CI & Holm $p$ & rank gain (Holm $p$) & seeds helped / hurt (raw)\\\midrule
B1 & MASSIVE & XLM-R & own seed on top of all signals & +2.4$^{*}$ & [+1.4, +3.5] & 0.0028 & +2.1 (0.0140) & 3/0\\
B2 & MASSIVE & Gemma & own seed on top of all signals & +3.3$^{*}$ & [+1.6, +5.0] & 0.0048 & +2.9 (0.0128) & 3/0\\
B3 & MASSIVE & Qwen-3B & own seed on top of all signals & +3.5$^{*}$ & [+2.1, +5.0] & 0.0028 & +3.2 (0.0028) & 3/0\\
B4 & MASSIVE & XLM-R & Gemma on top of all signals & +4.5$^{*}$ & [+2.7, +6.2] & 0.0028 & +4.9 (0.0028) & 3/0\\
B5 & MASSIVE & Gemma & XLM-R on top of all signals & +4.4$^{*}$ & [+2.6, +6.3] & 0.0028 & +3.3 (0.0040) & 3/0\\
B6 & MASSIVE & Qwen-3B & XLM-R on top of all signals & +4.3$^{*}$ & [+2.7, +5.9] & 0.0028 & +3.6 (0.0040) & 3/0\\
B7 & MASSIVE & Gemma & MC dropout (32 passes) minus own seed & -4.5$^{*}$ & [-6.5, -2.6] & 0.0028 & -4.4 (0.0028) & 0/3\\
B8 & CoNLL & XLM-R & own seed on top of all signals & +2.7$^{*}$ & [+0.9, +6.1] & 0.0042 & +2.8 (0.0320) & 3/0\\
B9 & CoNLL & Gemma & own seed on top of all signals & +1.8 & [+0.3, +4.0] & 0.0790 & +1.4 (0.1974) & 2/0\\
B10 & CoNLL & Qwen-3B & own seed on top of all signals & +0.9 & [+0.0, +1.6] & 0.2160 & +1.8 (0.0252) & 2/0\\
B11 & CoNLL & XLM-R & Gemma on top of all signals & +7.8$^{*}$ & [+4.7, +12.7] & 0.0028 & +7.6 (0.0028) & 3/0\\
B12 & CoNLL & Gemma & XLM-R on top of all signals & +0.3 & [-0.5, +1.3] & 0.8991 & +1.1 (0.1974) & 0/0\\
B13 & CoNLL & Qwen-3B & XLM-R on top of all signals & +0.3 & [-0.5, +1.1] & 0.8991 & +2.8 (0.1256) & 0/0\\
B14 & CoNLL & Gemma & MC dropout (32 passes) minus own seed & -1.9 & [-4.2, +1.0] & 0.5873 & -1.9 (0.1974) & 0/2\\
\bottomrule\end{tabular}
\caption{Confirmatory family B (B: partners on top of richer single-model scores; 14 tests, Holm). All signals = margin plus every richer single-model signal of Appendix~\ref{app:signals} that exists for the primary. B1--B6/B8--B13: does a partner add ranking value on top of all signals? B7, B14: MC dropout ($k=32$) versus an own-seed partner (negative = MC worse). Gains are eAURC reductions in $\times10^{-3}$ \eaurc{}, positive = helps. Raw combiner: 9 of 14 rejected; rank-mapped (post hoc): 10 of 14. $^{*}$ rejected under raw.}\label{tab:bfamily}\end{table*}

\begin{table*}[tp]\centering\scriptsize\setlength{\tabcolsep}{2.5pt}
\begin{tabular}{llrrl}\toprule
id & test (48 locales other than ja/zh; raw combiner; few-shot calibration) & est. & 95\% CI & Holm $p$\\\midrule
M1 & Gemma primary, mean $D$ & $+0.0268$ & [$+0.0227$, $+0.0310$] & $<10^{-4}$\\
M2 & XLM-R primary, mean $D$ & $-0.0015$ & [$-0.0038$, $+0.0009$] & 0.21\\
M3 & Qwen-3B primary, mean $D$ & $+0.0861$ & [$+0.0727$, $+0.0993$] & $<10^{-4}$\\
M4 & decoder-primary coefficient given F1 gap & $-0.0398$ & [$-0.0513$, $-0.0306$] & $<10^{-4}$ (not interpretable)\\
\midrule
\multicolumn{5}{l}{exploratory regressions of $D$ (locale-cluster bootstrap):}\\
\multicolumn{5}{l}{no control: decoder $+0.0579$ [$+0.0487$, $+0.0676$]}\\
\multicolumn{5}{l}{+ F1 gap: decoder $-0.0398$; F1 gap $+0.216$ [$+0.197$, $+0.239$]}\\
\multicolumn{5}{l}{+ F1 gap + change in calibrated AUROC + change in P(partner wrong $|$ primary wrong): F1 gap $+0.079$ [$+0.049$, $+0.110$];}\\
\multicolumn{5}{l}{\quad change in calibrated AUROC $+0.353$ [$+0.284$, $+0.418$], change in P(partner wrong $|$ primary wrong) $-0.108$ [$-0.174$, $-0.052$]}\\
\bottomrule\end{tabular}
\caption{Confirmatory family M (M: cross-lingual; Holm over 4) and the exploratory mediation regressions, under the few-shot calibration that the protocol specified. $D$ as in Table~\ref{tab:xlmain} (other-type minus own-seed gain, positive = the other type helps more); regression coefficients are in the same direction. The mediators are measured after treatment. M4 is formally significant but not interpretable: decoder and encoder primaries have almost no overlap in F1 gap, and the adjusted coefficient changes sign with the functional form (Appendix~\ref{app:mlexp}); so we do not read it as a sign reversal.}\label{tab:mfamily}\end{table*}

\subsection{Deviations}
\begin{description}[leftmargin=0pt,itemsep=2pt,font=\normalfont\bfseries]
\item[Combiner.] The first protocols specified a logistic combiner on standardised raw features. The rank-ECDF combiner was introduced after those results were seen. Re-running all 52 earlier decision rules under both combiners changes one verdict (SNIPS ``partner helps'' for Gemma: 2 of 3 seeds under rank-ECDF, 1 of 3 under raw).
\item[Cross-lingual main condition.] The cross-lingual protocols specified few-shot calibration (temperatures and combiners fit on 300 locale utterances). The main text uses zero-shot calibration (fit on en-US), chosen in a later protocol after the few-shot results were seen; Table~\ref{tab:cjk} gives both.
\item[CoNLL fit slice is shifted.] A protocol stated that the fit slices are not shifted. For CoNLL this is false (fit-slice utterance error 0.075--0.092, test 0.111--0.134), so the held-out test uses dev for CoNLL.
\item[Bootstrap size and scheme.] Protocols specified 1,000 replicates; we use 10,000. Some earlier tests drew a separate item resample per seed; all were re-run with one shared draw (Table~\ref{tab:crossed}), and the main text quotes the shared-draw results. Family A keeps separate item draws.
\item[Score rounding.] Confidence scores are rounded to $10^{-9}$ before ranking (Appendix~\ref{app:gliner}); this changes non-GLiNER results by less than $10^{-6}$.
\item[Diverged 7B seed.] Qwen2.5-7B on MASSIVE seed 1 diverged after its first dev check (dev F1 0.744 to 0.16, recovered to 0.58). No fallback had been declared, so the best-dev checkpoint was kept (test F1 0.725). The larger-model comparison is reported with and without it (Table~\ref{tab:abe}).
\item[Snapshots and missing weights.] Snapshot ensembles come from re-trained Gemma runs with the same recipe and seed. CoNLL XLM-R seed 0 has no saved weights, so it lacks hidden-state signals and MC dropout.
\item[Timing with stand-in weights.] mDeBERTa and Qwen-0.5B were timed with randomly initialised weights of the same architecture; timing does not depend on weight values.
\item[Prospective test.] None. Three Qwen2.5-0.5B runs crashed at load because a local cache lacked the weights file; they were re-run from scratch with the public weights at the pinned revision, before any test gain was computed.
\end{description}

\subsection{Crossed bootstrap}
All seeds of a model are scored on the same test utterances, so a bootstrap over seeds and utterances should apply one utterance resample to all seeds and models in a replicate.
Table~\ref{tab:crossed} re-runs every seed-averaged test with one shared draw and applies the rule that a claim counts only if this bootstrap and the 2-df $t$ interval \citep{knapp2003} agree.
\begin{table*}[tp]
\centering\footnotesize\setlength{\tabcolsep}{2.5pt}
\begin{tabular}{@{}>{\raggedright\arraybackslash}p{3.0cm}>{\raggedright\arraybackslash}p{1.75cm}>{\raggedright\arraybackslash}p{3.45cm}>{\raggedright\arraybackslash}p{4.35cm}>{\raggedright\arraybackslash}p{2.25cm}@{}}
\toprule
test family & earlier bootstrap & earlier result & crossed bootstrap, both intervals & verdict\\
\midrule
partner vs margin, 18 tests (Holm) & shared item draw & 14 of 18 & 12 of 18 pass both the crossed-Holm and the $t_2$ interval (14 under the bootstrap alone; CoNLL Qwen with an XLM-R or Gemma partner fails $t_2$) & lead with 12\\
other type vs own seed, 12 tests (Holm) & shared item draw & XLM-R primary 4 of 4; decoders 0 of 8 & same, under both intervals & unchanged\\
copy-1 average, pooled & separate draws & $+0.8$, $p=0.02$ (v7), $p=0.07$ (v7 re-computation, shared draw) & $+0.8$ [$-0.1$, 1.7], $p\approx0.07$; 0.26 of a seed's gain & unchanged ($p\approx.07$)\\
MC-dropout mean, equivalence at 0.001 & separate draws & equivalent on every dataset & equivalent on every dataset (90\% CI within $\pm0.15\times10^{-3}$) & unchanged\\
second head, CoNLL & separate draws & $-2.0$, bootstrap significant, $t_2$ borderline & $-2.0$ [$-3.8$, $-0.2$]; 95\% $t_2$ [$-5.0$, 1.1] & not established under both\\
readouts vs curve (Appendix~\ref{app:curve}) & Knapp--Hartung & second head below under 3 of 4 forms & below under 4 of 6 forms with both intervals, 6 of 6 crossed & stronger ``below''\\
partner types left out, accuracy-on-errors curve & Knapp--Hartung & own seed $-0.5$ [$-1.0$, $-0.03$]; other type $+0.5$ [$-0.1$, 1.1] & own seed $-0.7$ [$-1.4$, $-0.3$]; other family $+0.6$ [$+0.1$, 1.5], crossed only & own seed below; other family not significant\\
GLiNER G1--G6 (Holm) & separate draws & G1--G3 help; G4--G6 n.s. & same & unchanged\\
pair vs Qwen2.5-7B (Holm over 8) & separate draws & MASSIVE: bootstrap significant, $t_2$ not & $t_2$ includes 0 in all 8 & unchanged (no difference detected)\\
cross-lingual, other type $-$ own seed & locale clusters & XLM-R primary: $-3.5\times10^{-3}$ (raw, $p=0.04$) & XLM-R primary, rank-ECDF, relative: $-1.1\%$ [$-3.5$, 1.5] & drop the raw-only claim\\
serving rows & shared item draw & -- & re-derived exactly & unchanged\\
\bottomrule
\end{tabular}
\caption{Every seed-averaged test re-run with the crossed bootstrap: all seeds are scored on the same test utterances, so one utterance resample is applied to all seeds and models. The family tests and the serving rows already used one shared draw, so for them this is the same procedure re-run, not a robustness check; the copy-1, MC-mean, second-head, GLiNER and 7B tests had drawn a separate resample per seed. A claim counts as significant only if the crossed bootstrap (with Holm where a family applies) and the $t$ interval with 2 df agree. Gains $\times10^{-3}$ \eaurc{}, positive = helps.}
\label{tab:crossed}
\end{table*}

\section{How the numbers were re-computed}
\label{app:recompute}

Each experiment had two separate code bases.
The first trained models, saved per-word logits and computed the results.
The second was written afterwards in a separate session, with access to the protocol and the saved files but not to the first pipeline's analysis code, and re-derived the numbers.
It re-implemented the span decoder, temperature fitting, the signals, both combiners, the tie-aware metrics, the bootstraps, the Knapp--Hartung intervals and Holm's correction; only data loading and file locations were shared.
Both were written with an AI coding assistant.
Point estimates agreed to better than $10^{-6}$ and bootstrap intervals to within Monte-Carlo error; each disagreement was resolved by reading both code paths.
The checks caught errors that would otherwise have reached the paper, among them a learned router that looked better than the margin only because of the combiner, a spurious ``XLM-R partner is worse'' produced by ranking scores from differently fitted folds together, a seeding bug that gave two datasets the same seed draws, and sentences that went beyond the saved results.
The full list is released with the code.

\section{Data}
\label{app:data}

\begin{table}[tbp]
\centering\footnotesize\setlength{\tabcolsep}{1.8pt}
\begin{tabular}{@{}lrrrrl@{}}
\toprule
dataset & train & fit & dev & test & licence\\
\midrule
MASSIVE en-US & 9,514 & 2,000 & 2,033 & 2,974 & CC BY 4.0\\
CoNLL-2003 & 12,041 & 2,000 & 3,250 & 3,453 & research\\
SNIPS & 11,084 & 2,000 & 700 & 700 & CC0 1.0\\
MIT-Restaurant & 5,860 & 1,034 & 766 & 1,521 & research use$^\ast$\\
\bottomrule
\end{tabular}
\caption{Splits. The fit slice is held out of training and used only for temperatures and combiners (and as the held-out slice of \S\ref{sec:curve} for MASSIVE, SNIPS and MIT-Restaurant). Dev selects checkpoints. MIT-Restaurant's dev and fit slices are seeded random subsets of its 7,660 training queries. $^\ast$No formal licence; the creators state that the corpora are publicly available to support research in speech and language processing.}
\label{tab:splits}
\end{table}

\paragraph{Locales.} The zero-shot test sets are the full MASSIVE test splits of the 51 non-English locales. In the zero-shot condition (main) temperatures and combiners are fit on 2,000 en-US utterances; in the few-shot condition on 300 utterances from the locale's validation split.
For Japanese and Chinese our first analysis used one character per word, as the official MASSIVE baselines do; the main text re-segments them with SudachiPy \citep{takaoka2018sudachi} (mode A) and jieba.
Thai, Khmer and Burmese keep MASSIVE's whitespace chunks as words: 3.1--4.0 chunks per utterance, 8.6--9.4 characters per chunk, against a median of 6.25 chunks per utterance in the other locales (Japanese and Chinese excluded). These units are phrase-like. Dropping the three locales changes no result.

\paragraph{Domain shift.} CoNLL taggers were also applied to the five CrossNER domains \citep{liu2021crossner} and WNUT-17 \citep{derczynski2017}, with target types mapped to PER, ORG, LOC and MISC following the CoNLL guidelines (mapping released with the code). An XLM-R partner improves the Gemma primary's ranking in 19 of 27 pooled cells, while the copy-1 signal does so in 2; the cells share data, so the effective sample is the 3 tagger seeds.

\paragraph{Duplicates.} Exact-text duplicates between test and train inherited from the standard splits are 0.5\%, 3.7\% and 3.0\% of test (MASSIVE, CoNLL, SNIPS); every model sees them.

\section{Models and training}
\label{app:models}

\begin{table*}[tp]
\centering\footnotesize\setlength{\tabcolsep}{3pt}
\begin{tabular}{@{}>{\raggedright\arraybackslash}p{3.3cm}ll>{\raggedright\arraybackslash}p{0.50\textwidth}@{}}
\toprule
model & type & params & training\\
\midrule
XLM-R-large & encoder & 560M & full fine-tuning; lr 2e-5, 3 epochs (CoNLL), 3e-5, 5 epochs (MASSIVE, SNIPS); batch 16\\
Gemma-4-E2B echo & decoder & $\sim$5B (2B eff.) & LoRA r32/$\alpha$64/dropout 0.05 on q,k,v,o,gate,up,down; lr 1e-4; batch 16\\
Qwen2.5-3B echo & decoder & 3.1B & as Gemma\\
DeBERTa-v3-large & encoder & 435M & XLM-R's recipe\\
mDeBERTa-v3-base & encoder & 280M & XLM-R's recipe\\
Qwen2.5-0.5B echo & decoder & 0.5B & Qwen-3B's recipe\\
XLM-R-large + LoRA & encoder & 560M & LoRA r32/$\alpha$64; lr 3e-4 (dev-selected); head fully trained\\
Qwen2.5-7B echo & decoder & 7B & Qwen-3B's recipe\\
GLiNER-large-v2.1 & span model & 0.45B & fine-tuned on each training split; epochs chosen on dev\\
GLiNER-large-v2.1 & span model & 0.45B & zero-shot, threshold 0.5\\
XLM-R-XL & encoder & 3.5B & XLM-R's recipe; lr from \{1e-5, 2e-5\} (CoNLL) or \{1e-5, 3e-5\} (MASSIVE) on seed-0 dev F1\\
Qwen2.5-3B, one copy, no causal mask & decoder & 3.1B & Qwen-3B's recipe; padding-only attention mask\\
Qwen2.5-3B, one copy, causal & decoder & 3.1B & Qwen-3B's recipe\\
\bottomrule
\end{tabular}
\caption{Taggers. Echo taggers read ``\texttt{<s>} $Q$ \texttt{||} $Q$'' (Qwen uses \texttt{<|endoftext|>} as start token) and use a linear head with dropout 0.1 on the copy-2 first non-space piece of each word; 6\% warmup with linear decay, gradient clipping 1, bf16, dev F1 every half epoch, best checkpoint kept. Encoders label the first subword. Labels are decoded word by word (argmax), without a CRF or BIO constraint. Every trained tagger has seeds 0, 1, 2 and one temperature per model, fit on fit-slice word NLL. The MIT-Restaurant taggers use the MASSIVE/SNIPS recipe unchanged.}
\label{tab:models}
\end{table*}

\begin{table*}[tp]
\centering\small\setlength{\tabcolsep}{4pt}
\begin{tabular}{lccc}
\toprule
model & MASSIVE & CoNLL & SNIPS\\
\midrule
XLM-R-large & .8161 / .8145 / .8139 & .9195 / .9254 / .9202 & .9649 / .9627 / .9431\\
Gemma-4-E2B echo & .7824 / .8035 / .7945 & .9307 / .9314 / .9288 & .9594 / .9535 / .9642\\
Qwen2.5-3B echo & .8091 / .7909 / .7891 & .9301 / .9324 / .9287 & .9526 / .9622 / .9613\\
DeBERTa-v3-large & .8176 / .8156 / .8179 & .9259 / .9269 / .9273 & .9674 / .9699 / .9674\\
mDeBERTa-v3-base & .7665 / .7854 / .7863 & .9179 / .9210 / .9181 & .9542 / .9536 / .9509\\
Qwen2.5-0.5B echo & .7882 / .7852 / .7926 & .9175 / .9162 / .9209 & .9489 / .9493 / .9536\\
XLM-R-large LoRA & .8127 / .8176 / .8175 & .9192 / .9252 / .9220 & .9555 / .9542 / .9567\\
Qwen2.5-7B echo & .7997 / .7250 / .8059 & .9337 / .9272 / .9321 & .9564 / .9554 / .9614\\
GLiNER-large fine-tuned & .8280 / .8325 / .8313 & .9341 / .9348 / .9304 & --\\
GLiNER-large zero-shot & .1997 & .5508 & --\\
XLM-R-XL & .7733 / .7664 / .7652 & .9144 / .9120 / .9137 & --\\
Qwen2.5-3B one copy, no mask & .7924 (.7882--.7960) & .9228 (.9210--.9246) & --\\
Qwen2.5-3B one copy, causal & .7056 & .7449 & --\\
\bottomrule
\end{tabular}
\caption{Test span F1 per seed (0 / 1 / 2), or seed mean (range). Word calibration error after temperature scaling: XLM-R 0.015--0.017, Gemma 0.010--0.012, Qwen 0.007--0.011 on MASSIVE; 0.004--0.008 on CoNLL.}
\label{tab:f1}
\end{table*}

\begin{figure}[tbp]
  \centering
  \includegraphics[width=\columnwidth]{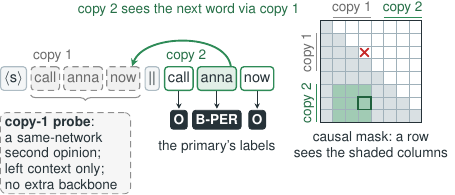}
  \caption{\textbf{An echo tagger labels each word from the second copy of its input, which sees the whole sentence.}
The causal decoder reads ``\texttt{<s>} $Q$ \texttt{||} $Q$''.
A word in copy~2 can attend to every word of copy~1, including the words to its right (right: the attention mask, where each row attends to the shaded columns; ringed cell); a word in copy~1 cannot (crossed cell).
A probe on copy~1 is a same-network second opinion: it sees only left context and adds no backbone cost (\S\ref{sec:same}).}
  \label{fig:echoschem}
\end{figure}

\paragraph{Echo taggers.} Figure~\ref{fig:echoschem} shows how an echo tagger reads its input and where the copy-1 probe sits.

\paragraph{Same-network readouts.}
All probes and heads are trained on the frozen states of the fine-tuned primary, saved from the training split, with early stopping on 10\% of training sentences.
Copy-1 readouts: a linear or MLP (one hidden layer of 512, GELU) probe on the copy-1 last-piece state of each word, or the primary's own entity head applied to the copy-1 state.
The second head is a linear or MLP head on the copy-2 states that the primary's head reads.
The MC-dropout mean averages 8 stochastic passes (32 in sweeps).
Each readout is tempered on the fit slice and averaged with the primary exactly as a partner is.
\paragraph{MC dropout settings.} At the trained rate, dropout is active in the LoRA adapters (0.05) and the head (0.1). S1: LoRA dropout 0.10 on all seven LoRA modules, attention-probability dropout 0.10, head 0.10. S2: 0.20, 0.20 and 0.10. S3: S1 plus dropout 0.10 on the output of every attention and MLP block.
\paragraph{Single-copy decoders.} The bidirectional variant passes a padding-only attention mask to every layer with causal masking disabled; a check confirmed that the output at position 0 changes when later tokens change.

\section{Signals, combiners, metrics and inference}
\label{app:signals}
\paragraph{Signals.} Margin $=1-\min_w(\text{top1}-\text{top2})$ over words. Averaged margin $\savg(B)=1-\min_w\text{margin}(\tfrac12(p_A+p_B))$; label-aware score $\slab(B)=1-\min_w[\bar p(\hat y_w)-\max_{y\neq\hat y_w}\bar p(y)]$, where $\hat y_w$ is the primary's label (a second variant, $1-\min_w\bar p(\hat y_w)$, gains less: $+0.17\times10^{-3}$, 66\% of pairings improve). An own-seed partner of seed $s$ uses seed $(s+1)\bmod 3$; partners of other types use seed $s$. The richer baseline adds sequence and mean token NLL, mean and maximum entropy, the product of span maximum probabilities, a Mahalanobis distance on hidden states \citep{lee2018maha}, and a hidden-state error predictor trained out of fold on the fit slice \citep{vazhentsev2022}.
\paragraph{Combiners.} Raw: standardised features, logistic regression ($C=1$). Rank-ECDF: each feature mapped through a piecewise-linear mid-rank ECDF of its fit-slice values, then the same logistic model; the margin-only baseline is the one-feature version. Combiner-free checks rank by $\savg(B)$ alone or by an equal-weight average of the two ranks.
\paragraph{Metrics.} With $n$ utterances ranked by score and $E_k$ the expected number of errors among the $k$ most confident (ties broken uniformly at random, in closed form), $\text{AURC}=\frac1n\sum_k E_k/k$, $\text{eAURC}=\text{AURC}-\text{AURC}_{\text{opt}}$ and $\text{AUGRC}=\frac1n\sum_k E_k/n$ \citep{traub2024}. Scores are rounded to $10^{-9}$ before ranking. The gain is $\text{eAURC}(\text{margin})-\text{eAURC}(\text{margin}+\savg(B))$; the relative gain (in points) is 100 times this divided by $\text{eAURC}(\text{margin})$. Accuracy on the primary's errors is $1-P(B\text{ wrong}\mid A\text{ wrong})$ at the utterance level. ``Share of the best gain lost'' is mean regret divided by the mean best gain (a ratio of means).
\paragraph{Inference.} The seed-and-item bootstrap draws 3 seeds with replacement and one item resample shared by all seeds and models, then averages (10,000 replicates); $p=2\min(P^*(\Delta\ge0),P^*(\Delta\le0))$, floored at $10^{-4}$. With three seeds the seed stage understates between-seed variance, so we also report Knapp--Hartung intervals ($t$ with 2 df; \citealp{knapp2003}). Equivalence uses two one-sided tests \citep{schuirmann1987}. Regressions over pairings use cluster-robust errors with wild cluster bootstraps \citep{cameron2008} with Webb weights \citep{webb2023}. Cross-lingual tests cluster on locales.

\section{Compute}
\label{app:compute}
All work used four NVIDIA H100 80GB GPUs on one machine.
The training script logged its own GPU time for 71 runs: 12.8 GPU-hours in total, 5--16 minutes per echo-tagger run; the DeBERTa, mDeBERTa and XLM-R LoRA partners took 4.7 GPU-hours, and the 15 MIT-Restaurant taggers 1.4 GPU-hours (3--8 minutes per run). The prompted LLM generated all fit, dev and test outputs in about 4 minutes on one GPU.
Inference cost per query (Table~\ref{tab:flops}) counts $2\times$ non-embedding parameters $\times$ input tokens; echo taggers process the doubled input. FLOPs are not time: on short MASSIVE inputs Qwen2.5-7B (0.87\,ms per query at batch 32) is faster than Qwen2.5-3B (1.07\,ms), which has more layers (36 against 28).

\begin{table}[tbp]
\centering\footnotesize\setlength{\tabcolsep}{3.5pt}
\begin{tabular}{@{}lrrrr@{}}
\toprule
 & \multicolumn{2}{c}{CoNLL} & \multicolumn{2}{c}{MASSIVE}\\
model & tokens & GFLOPs & tokens & GFLOPs\\
\midrule
mDeBERTa-v3-base & 22.4 & 3.8 & 10.7 & 1.8\\
DeBERTa-v3-large & 18.2 & 11.0 & 9.3 & 5.6\\
XLM-R-large & 21.2 & 12.9 & 10.0 & 6.1\\
GLiNER-large & 27.2 & 17.1 & 171.3 & 107.5\\
Qwen2.5-0.5B echo & 43.7 & 31.3 & 16.7 & 12.0\\
XLM-R-XL & 21.2 & 120.3 & 10.0 & 57.0\\
Gemma-4-E2B echo & 42.4 & 159.2 & 16.7 & 62.5\\
Qwen2.5-3B echo & 43.7 & 242.8 & 16.7 & 92.8\\
Qwen2.5-7B echo & 43.7 & 570.9 & 16.7 & 218.2\\
\bottomrule
\end{tabular}
\caption{Mean input tokens and forward GFLOPs per test query. GLiNER's MASSIVE inputs are long because its prompt lists all slot types.}
\label{tab:flops}
\end{table}

\section{The held-out test in detail}
\label{app:oos}

\paragraph{Design.} The 576 pairings of the trained grid keep their test gains (rank-ECDF \eaurc{}). Every candidate quantity is recomputed from saved logits on a held-out slice: the fit slice for MASSIVE and SNIPS (2,000 utterances each) and dev for CoNLL (3,250). Within-primary Spearman correlations are averaged over the 72 primaries. Out-of-sample $R^2$ refits a within-primary linear model with some partner types or datasets left out.

\paragraph{Two intervals.} The pre-specified interval resamples the 8 primary types and replaces each test gain by a shared item-bootstrap replicate (10,000 draws). Because the replicate gains add noise, it sits around an attenuated statistic (bootstrap mean 0.38). The main text therefore reports a two-way interval that resamples primary and partner types instead (Table~\ref{tab:g2}).

\begin{table*}[tp]
\centering\scriptsize\setlength{\tabcolsep}{2.6pt}
\begin{tabular}{@{}lcccccccc@{}}
\toprule
held-out quantity & \hd{$\rho$\\(95\% CI)} & \hd{$R^2$ types\\(95\% CI)} & \hd{$R^2$ data\\(95\% CI)} & \hd{$R^2$ type\\(95\% CI)} & regret & top-1 & top-3 & \hd{test-slice $R^2$\\(M/C/S)}\\
\midrule
accuracy on primary's errors & .47 [.23, .52] & .26 [.05, .34] & .22 [.04, .29] & .51 [$-$.53, .58] & 1.13 & 0.22 & 0.56 & .38/.32/.15\\
\quad double fault (same ranking) & .47 [.23, .52] & .26 [.05, .34] & .30 [.06, .38] & .52 [$-$.66, .58] & 1.13 & 0.22 & 0.56 & .37/.32/.15\\
error correlation $\phi$ & .31 [.09, .42] & .13 [$-$.04, .21] & .17 [$-$.04, .25] & $-$.11 [$-$.77, .17] & 1.61 & 0.19 & 0.51 & .07/.30/.12\\
Cohen's $\kappa$ & .31 [.09, .42] & .11 [$-$.05, .20] & .17 [$-$.04, .25] & $-$.11 [$-$.73, .17] & 1.62 & 0.18 & 0.54 & .06/.29/.11\\
Yule's Q & .29 [.08, .40] & .05 [$-$.09, .15] & .08 [$-$.07, .18] & $-$.14 [$-$.97, .14] & 1.75 & 0.18 & 0.51 & .02/.27/.10\\
partner wrong where primary right & $-$.02 [$-$.18, .12] & $-$.12 [$-$.22, $-$.00] & $-$.10 [$-$.33, .01] & $-$.17 [$-$.58, .01] & 2.75 & 0.06 & 0.22 & .01/.06/.00\\
overall F1 gap & .19 [$-$.02, .30] & $-$.04 [$-$.21, .07] & .05 [$-$.18, .12] & .03 [$-$.25, .17] & 1.94 & 0.04 & 0.42 & .09/.01/.02\\
AUROC of partner's margin & $-$.10 [$-$.20, .05] & $-$.12 [$-$.18, $-$.00] & $-$.02 [$-$.12, .01] & $-$.15 [$-$.47, .08] & 2.65 & 0.01 & 0.31 & .08/.06/.06\\
\quad within primary-margin deciles & .24 [.06, .34] & .08 [$-$.06, .23] & .13 [$-$.00, .26] & .21 [$-$.31, .42] & 1.56 & 0.10 & 0.50 & .61/.41/.24\\
expected margin drop & .41 [.13, .54] & .20 [$-$.01, .33] & .21 [.01, .31] & .44 [$-$.67, .51] & 1.24 & 0.22 & 0.62 & .67/.22/.52\\
disagreement, no labels & .23 [.02, .38] & .02 [$-$.12, .12] & $-$.20 [$-$.46, .04] & $-$.24 [$-$.79, .09] & 1.84 & 0.20 & 0.46 & .02/.23/.09\\
gain measured on dev (reference) & .51 [.27, .58] & .36 [.10, .47] & .26 [$-$.11, .43] & .57 [$-$.68, .61] & 1.15 & 0.11 & 0.60 & .56/.24/.47\\
\bottomrule
\end{tabular}
\caption{Held-out predictors of test gain in full (primary held-out slices; outcome rank-ECDF \eaurc{} gain, positive = helps). $\rho$: pooled within-primary Spearman; its pre-specified two-stage interval resamples primary types and replaces each test gain by a shared item-bootstrap replicate, which adds noise and pulls the correlation down (bootstrap mean 0.38 for the first row), so the point estimate sits near the top of the interval; the interval is conservative for a test of $\rho>0$. $R^2$: out-of-sample within-primary $R^2$ leaving out partner types (types), datasets (data) or partner types (own seed, same pretraining, other family); partner-type intervals are wide (8 types). Regret: mean over 72 primaries of the best partner's test gain minus the chosen partner's ($\times10^{-3}$ \eaurc{}; ties split uniformly, as everywhere; the first tied partner gives 1.16 for the first two rows; random partner 1.90, always own seed 3.43, mean best gain 6.35). Top-1/top-3: share of primaries whose best partner is ranked first / in the top three (chance 0.125 / 0.375). Test-slice $R^2$: in-sample within-primary $R^2$ with the quantity computed on the test set (not out of sample). Double fault and accuracy on errors rank partners identically, so their $\rho$, regret and top-$k$ columns agree by construction; their slope-transfer $R^2$ differ.}
\label{tab:predfull}
\end{table*}

\begin{table*}[tp]
\centering\scriptsize\setlength{\tabcolsep}{3pt}
\begin{tabular}{@{}lccccc@{}}
\toprule
accuracy on errors minus & \hd{pooled $\rho$\\difference} & MASSIVE & CoNLL & SNIPS & \hd{$R^2$ difference, partner types left out\\(other minus accuracy on errors)}\\
\midrule
double fault (same ranking) & +.00 [.00, .00] & +.00 [.00, .00] & +.00 [.00, .00] & +.00 [.00, .00] & $-$.00 [$-$.01, .01]\\
error correlation $\phi$ & +.16 [.02, .23] & +.32 [.04, .48] & +.03 [$-$.15, .18] & +.14 [$-$.07, .26] & $-$.13 [$-$.25, .05]\\
Cohen's $\kappa$ & +.16 [.01, .23] & +.31 [.03, .47] & +.02 [$-$.15, .17] & +.16 [$-$.07, .27] & $-$.15 [$-$.27, .04]\\
Yule's Q & +.19 [.03, .25] & +.33 [.03, .50] & +.04 [$-$.17, .21] & +.18 [$-$.06, .30] & $-$.21 [$-$.33, .00]\\
partner wrong where primary right & +.50 [.20, .63] & +.51 [.07, .81] & +.54 [.28, .76] & +.45 [$-$.05, .65] & $-$.38 [$-$.50, $-$.10]\\
overall F1 gap & +.29 [.08, .42] & +.31 [$-$.02, .60] & +.29 [.14, .45] & +.26 [$-$.09, .48] & $-$.30 [$-$.38, $-$.12]\\
AUROC of partner's margin & +.57 [.25, .66] & +.60 [.16, .80] & +.61 [.27, .84] & +.50 [$-$.05, .72] & $-$.38 [$-$.46, $-$.10]\\
within primary-margin deciles & +.23 [.03, .34] & +.31 [.01, .46] & +.24 [$-$.03, .42] & +.15 [$-$.23, .44] & $-$.18 [$-$.28, .03]\\
expected margin drop & +.07 [$-$.09, .20] & +.14 [$-$.14, .30] & +.07 [$-$.12, .27] & $-$.01 [$-$.29, .28] & $-$.06 [$-$.16, .11]\\
disagreement, no labels & +.25 [.02, .34] & +.46 [.08, .66] & +.01 [$-$.20, .17] & +.27 [$-$.16, .49] & $-$.24 [$-$.38, $-$.00]\\
gain measured on dev (reference) & $-$.04 [$-$.16, .08] & $-$.05 [$-$.21, .11] & $-$.05 [$-$.16, .06] & +.00 [$-$.29, .26] & +.10 [$-$.03, .25]\\
\bottomrule
\end{tabular}
\caption{Head to head on the same bootstrap draws. Left: within-primary Spearman of accuracy on errors minus that of each other held-out quantity (positive = accuracy on errors ranks partners better). Right: out-of-sample $R^2$ (partner types left out) of each quantity minus that of accuracy on errors. The double-fault difference is zero by construction. Against error correlation, $\kappa$ and Yule's Q the pooled lead excludes zero, but per dataset only on MASSIVE.}
\label{tab:preddiff}
\end{table*}

\begin{table*}[tp]\centering\footnotesize\setlength{\tabcolsep}{3.5pt}
\begin{tabular}{@{}>{\raggedright\arraybackslash}p{3.3cm}ccccccc@{}}\toprule
 & & \multicolumn{3}{c}{all 8 partners} & \multicolumn{3}{c}{without the own seed}\\\cmidrule(lr){3-5}\cmidrule(l){6-8}
primaries & $n$ & acc.\ on errors & F1 & difference & acc.\ on errors & F1 & difference\\\midrule
all & 72 & .47 [.21, .61] & .19 & .29 [.01, .54] & .37 [.09, .55] & .21 & .15 [$-$.08, .39]\\
encoders: XLM-R, XLM-R LoRA, DeBERTa-v3, mDeBERTa & 36 & .60 [.23, .76] & .12 & .48 [.02, .88] & .48 [.05, .72] & .18 & .30 [$-$.08, .64]\\
decoders: Gemma, Qwen-3B, Qwen-0.5B, Qwen-7B & 36 & .35 [.02, .56] & .26 & .09 [$-$.15, .38] & .25 [$-$.05, .48] & .24 & .01 [$-$.21, .25]\\
Gemma only & 9 & .14 [$-$.18, .45] & .23 & $-$.09 [$-$.32, .15] & .16 [$-$.25, .53] & .25 & $-$.08 [$-$.35, .18]\\
\midrule
MASSIVE & 24 & .54 [.16, .74] & .24 & .31 [$-$.10, .76] & .45 [.06, .69] & .26 & .18 [$-$.20, .64]\\
CoNLL & 24 & .41 [$-$.06, .71] & .12 & .29 [.02, .56] & .28 [$-$.24, .70] & .15 & .14 [$-$.10, .43]\\
SNIPS & 24 & .46 [.09, .71] & .20 & .26 [$-$.11, .60] & .37 [$-$.09, .70] & .23 & .14 [$-$.24, .51]\\
\bottomrule\end{tabular}
\caption{\textbf{The held-out rule is strongest for encoder primaries ($\rho=0.60$) and leads span F1 on every dataset} (post hoc). Mean within-primary Spearman correlation of held-out accuracy on the primary's errors and of held-out F1 with test gain, with 95\% intervals from a two-way bootstrap that resamples primary types and partner types (test gains at their observed values; for Gemma primaries only partner families vary). The pre-specified interval, which resamples primary types and test utterances, is [0.23, 0.52] for the first row.}
\label{tab:g2}\end{table*}

\paragraph{Robustness.} The pooled Spearman of held-out accuracy on errors is 0.42 on the secondary slices (MASSIVE dev, the shifted CoNLL fit slice, SNIPS dev), 0.47 with the raw combiner's gain as outcome and 0.51 with AUGRC. Split-half reliability (Spearman--Brown) is 0.82, 0.86 and 0.81 on the three datasets. Leaving out one partner type at a time gives 0.43--0.49; removing one type both as primary and as partner gives 0.44--0.52 (held-out F1: 0.10--0.24). Out of sample, the check also predicts the gain of partner types left out of the fit ($R^2$ 0.26; F1: $-0.04$).

\paragraph{Selection policies.} Ties are split uniformly throughout (the expected result of random tie-breaking); four of the 72 primaries have a tie at the top of the check. Picking the partner with the highest held-out accuracy on errors has regret $1.13\times10^{-3}$ \eaurc{} (1.16 if the first tied partner is taken), against 1.90 for a random partner (difference $-0.76$ [$-1.36$, $-0.25$], primary and partner types resampled; with the first tied partner and primary types and test utterances resampled instead, $-0.74$ [$-1.17$, $-0.31$]) and 3.43 for always the own seed. The own seed is the best partner for 3\% of primaries. The rule picks the best partner for 22\% (chance 12.5\%). Two fixed types do better in hindsight: always Gemma (0.94) and always Qwen-3B (0.99). Choosing one type from its average held-out accuracy on errors gives 1.56.

\begin{table*}[tp]
\centering\small\setlength{\tabcolsep}{3.5pt}
\begin{tabular}{llccc}
\toprule
pairings & dataset & accuracy on errors & + AUROC & + margin drop\\
\midrule
decoder primary, XLM-R partner & MASSIVE & $-$.42 [$-$1.38, +.56] & $-$.49 [$-$1.39, +.40] & $-$.67 [$-$1.63, +.27]\\
decoder primary, XLM-R partner & CoNLL & $-$.82 [$-$1.74, +.13] & $-$1.42 [$-$2.28, $-$.59] & $-$.68 [$-$1.49, +.15]\\
decoder primary, XLM-R partner & SNIPS & $-$.67 [$-$1.49, +.05] & $-$.37 [$-$1.20, +.45] & $-$.16 [$-$.90, +.60]\\
\addlinespace[2pt]
decoder primary, any encoder partner & MASSIVE & $-$.38 [$-$.93, +.14] & $-$.38 [$-$.93, +.14] & $-$.43 [$-$.98, +.11]\\
decoder primary, any encoder partner & CoNLL & +.55 [$-$.20, +1.42] & +.30 [$-$.20, +.90] & +.64 [+.07, +1.31]\\
decoder primary, any encoder partner & SNIPS & $-$.24 [$-$.61, +.10] & $-$.08 [$-$.42, +.26] & $-$.07 [$-$.39, +.25]\\
\addlinespace[2pt]
fine-tuned GLiNER partner & MASSIVE & $-$1.41 [$-$2.59, $-$.22] & $-$1.15 [$-$2.57, +.39] & $-$.33 [$-$1.66, +1.01]\\
fine-tuned GLiNER partner & CoNLL & $-$2.73 [$-$4.39, $-$1.52] & $-$8.47 [$-$18.83, $-$.16] & $-$2.10 [$-$5.72, +.38]\\
\addlinespace[2pt]
XLM-R-XL partner & MASSIVE & $-$2.91 [$-$4.14, $-$1.65] & $-$2.46 [$-$4.28, $-$.53] & $-$2.16 [$-$3.38, $-$.92]\\
XLM-R-XL partner & CoNLL & +1.61 [+.30, +3.29] & +.20 [$-$1.22, +1.62] & +1.86 [+.82, +2.92]\\
\bottomrule
\end{tabular}
\caption{Residuals (observed minus predicted test gain, $\times10^{-3}$ \eaurc{}, 95\% model-type cluster bootstrap) under within-primary models fitted to the trained grid with held-out predictors: accuracy on errors alone, plus the AUROC of the partner's margin, or plus the expected margin drop (exploratory). Decoder primaries with an XLM-R partner sit below the line on all three datasets, none significantly; adding AUROC does not remove this and makes the CoNLL residual significant. Fine-tuned GLiNER falls short on both datasets; its held-out margin drop is low (0.011 against a grid mean of 0.038 on MASSIVE, 0.048 against 0.096 on CoNLL), and adding it removes the MASSIVE shortfall but only a small part of the CoNLL one, whose interval widens. XLM-R-XL lands off the curve in opposite directions on the two datasets. Zero-shot GLiNER is not shown: its margin drop is negative and far outside the grid, so models using it extrapolate.}
\label{tab:anomalies}
\end{table*}

\paragraph{Anomalies and new types.} Decoder primaries with an XLM-R partner sit below the line on all three datasets, none significantly (Table~\ref{tab:anomalies}); the AUROC of XLM-R's margin on decoder errors does not explain this, and its cause is open. New partner types scatter by up to about $3\times10^{-3}$ (absolute \eaurc{}) around the grid line: fine-tuned GLiNER $-1.4$ and $-2.7$ (MASSIVE, CoNLL), XLM-R-XL $-2.9$ and $+1.6$, zero-shot GLiNER $-2.9$ and $-0.1$. The label-aware score leaves all of these unchanged (Appendix~\ref{app:labelaware}).

\section{Ways of choosing a partner, in full}
\label{app:policies}

\paragraph{Definitions.} Table~\ref{tab:policiesfull} lists every policy, with and without the own seed among the candidates. Partner accuracy (span F1, word accuracy, utterance accuracy) is measured on the same held-out slices as the check. The type prior scores a partner type (own seed or a named type) by its mean test gain on the other datasets, over primaries of other types; the pairing prior uses primaries of the same type; for MIT-Restaurant both come from the three old datasets. Intervals resample primary and partner types with test gains fixed (10,000 draws); on MIT-Restaurant only 5 types are resampled, so its intervals are wide. Every number in this section is post hoc.
\paragraph{Encoder and decoder primaries.} With the own seed set aside, the check misses 16\% of the best gain for the 36 encoder primaries, half of span F1's 33\% (random partner: 26\%); for the 36 decoder primaries the two tie (20\% and 19\%; random: 26\%).

\begin{table*}[tp]
\centering\footnotesize\setlength{\tabcolsep}{2.4pt}
\begin{tabular}{@{}l rr rr r rr rr@{}}
\toprule
 & \multicolumn{5}{c}{3 datasets, 72 primaries} & \multicolumn{4}{c}{MIT-Restaurant, 15 primaries}\\
\cmidrule(lr){2-6}\cmidrule(l){7-10}
 & $\rho$ & missed & \multicolumn{2}{c}{minus check} & pairs & $\rho$ & missed & \multicolumn{2}{c}{minus check}\\
\cmidrule(lr){4-5}\cmidrule(l){9-10}
partner chosen by & & & $\rho$ & missed (pp) & & & & $\rho$ & missed (pp)\\
\midrule
\multicolumn{10}{@{}l}{\emph{own seed among the candidates}}\\
own seed, always & -- & 54\% & -- & +36 [10, 43] & .42 & -- & 41\% & -- & +30 [$-$3, 37]\\
random partner & -- & 30\% & -- & +12 [4, 21] & .50 & -- & 24\% & -- & +13 [$-$2, 21]\\
any but the own seed & .36 & 26\% & $-$.11 [$-$.27, .32] & +9 [1, 17] & .58 & .35 & 20\% & $-$.05 [$-$.62, .63] & +9 [$-$4, 18]\\
type prior & .37 & 19\% & $-$.11 [$-$.41, .20] & +1 [$-$10, 16] & .64 & .42 & 15\% & +.02 [$-$.47, .50] & +5 [$-$14, 16]\\
pairing prior & .42 & 16\% & $-$.05 [$-$.35, .24] & $-$2 [$-$10, 7] & .67 & .57 & 6\% & +.17 [$-$.27, .53] & $-$5 [$-$16, 10]\\
span F1 & .19 & 31\% & $-$.29 [$-$.54, $-$.01] & +13 [1, 23] & .58 & .04 & 26\% & $-$.36 [$-$1.03, .21] & +15 [$-$6, 26]\\
utterance accuracy & .25 & 30\% & $-$.23 [$-$.46, .01] & +12 [1, 22] & .59 & .07 & 26\% & $-$.33 [$-$.90, .23] & +16 [$-$5, 27]\\
word accuracy & .33 & 18\% & $-$.14 [$-$.34, .07] & 0 [$-$4, 9] & .62 & .12 & 23\% & $-$.28 [$-$.77, .27] & +12 [$-$7, 21]\\
error corr.\ $\phi$ & .31 & 25\% & $-$.16 [$-$.33, .03] & +8 [$-$2, 15] & .62 & .36 & 24\% & $-$.04 [$-$.48, .44] & +13 [$-$11, 25]\\
disagreement (no labels) & .23 & 29\% & $-$.25 [$-$.49, .04] & +11 [$-$2, 19] & .59 & .29 & 30\% & $-$.11 [$-$.69, .49] & +19 [$-$11, 33]\\
expected margin drop & .41 & 20\% & $-$.07 [$-$.26, .16] & +2 [$-$7, 8] & .66 & .37 & 8\% & $-$.03 [$-$.34, .34] & $-$3 [$-$10, 7]\\
\textbf{accuracy on errors} & .47 & 18\% &  &  & .68 & .40 & 11\% &  & \\
type prior + check & .48 & 15\% & +.01 [$-$.15, .17] & $-$3 [$-$7, 4] & .69 & .49 & 11\% & +.09 [$-$.32, .51] & 0 [$-$11, 12]\\
gain on dev & .51 & 18\% & +.04 [$-$.14, .22] & 0 [$-$8, 7] & .70 & .49 & 13\% & +.09 [$-$.39, .73] & +2 [$-$13, 12]\\
\addlinespace[3pt]
\multicolumn{10}{@{}l}{\emph{own seed removed from the candidates}}\\
random partner & -- & 26\% & -- & +9 [1, 17] & .50 & -- & 19\% & -- & +9 [$-$4, 18]\\
type prior & .22 & 19\% & $-$.15 [$-$.53, .24] & +1 [$-$10, 16] & .58 & .33 & 15\% & +.06 [$-$.89, .86] & +5 [$-$14, 16]\\
pairing prior & .32 & 16\% & $-$.05 [$-$.40, .28] & $-$2 [$-$10, 7] & .63 & .59 & 5\% & +.32 [$-$.50, .91] & $-$5 [$-$16, 4]\\
span F1 & .21 & 27\% & $-$.15 [$-$.39, .08] & +9 [0, 16] & .59 & .05 & 21\% & $-$.22 [$-$.97, .33] & +11 [$-$10, 17]\\
utterance accuracy & .28 & 25\% & $-$.09 [$-$.29, .10] & +8 [0, 12] & .60 & .11 & 19\% & $-$.16 [$-$.80, .40] & +8 [$-$10, 16]\\
word accuracy & .35 & 16\% & $-$.02 [$-$.20, .16] & $-$2 [$-$5, 6] & .64 & .08 & 19\% & $-$.19 [$-$.72, .67] & +8 [$-$12, 18]\\
error corr.\ $\phi$ & .15 & 25\% & $-$.22 [$-$.42, .04] & +7 [$-$2, 15] & .55 & .23 & 23\% & $-$.04 [$-$.69, .75] & +13 [$-$10, 26]\\
disagreement (no labels) & .04 & 28\% & $-$.32 [$-$.62, .06] & +11 [$-$2, 19] & .52 & .15 & 30\% & $-$.11 [$-$1.13, .89] & +19 [$-$12, 34]\\
expected margin drop & .32 & 19\% & $-$.05 [$-$.29, .22] & +1 [$-$7, 8] & .63 & .36 & 7\% & +.09 [$-$.53, .69] & $-$3 [$-$12, 7]\\
\textbf{accuracy on errors} & .37 & 18\% &  &  & .64 & .27 & 10\% &  & \\
type prior + check & .38 & 15\% & +.01 [$-$.17, .20] & $-$3 [$-$7, 4] & .64 & .46 & 10\% & +.19 [$-$.53, .83] & 0 [$-$11, 13]\\
gain on dev & .47 & 18\% & +.10 [$-$.13, .31] & 0 [$-$8, 7] & .67 & .39 & 12\% & +.12 [$-$.68, .90] & +2 [$-$13, 14]\\
\bottomrule
\end{tabular}
\caption{All ways of choosing a partner, with and without the own seed among the candidates. $\rho$: within-primary Spearman with test gain, averaged over primaries; missed: share of the best candidate's test gain that the chosen one misses (ties split uniformly); pairs: share of candidate pairs ordered correctly (chance .50). Minus check: the policy's value minus the check's, with a 95\% interval from a two-way bootstrap over primary types and partner types (test gains fixed; 8 types on the three datasets, 5 on MIT-Restaurant, so the MIT-Restaurant intervals are wide). Positive missed differences mean the policy misses more than the check. Priors use test gains on the other datasets only. On MIT-Restaurant, the priors, word and utterance accuracy, prior + check and ``any but the own seed'' were computed after its data were seen. The pre-specified MIT-Restaurant computation took the first tied partner (check: 9.7\% missed); with ties split it is 10.9\%.}
\label{tab:policiesfull}
\end{table*}

\paragraph{Four datasets.} Table~\ref{tab:fourdata} pools the 72 primaries of the old datasets with the 15 of MIT-Restaurant. Word accuracy ties the check on MASSIVE and SNIPS and is close to random on CoNLL and MIT-Restaurant; pooled, the check's lead over it is not significant, while its lead over span F1 and utterance accuracy is (on the share missed). A crossed random-effects model gives the same picture.

\begin{table}[tbp]
\centering\footnotesize\setlength{\tabcolsep}{2.2pt}
\begin{tabular}{@{}lrrrrr@{}}
\toprule
 & MAS & CoN & SNI & MIT & pooled\\
\midrule
\multicolumn{6}{@{}l}{\emph{$\rho$ with test gain}}\\
\quad \textbf{check} & .54 & .41 & .46 & .40 & .46\\
\quad gain on dev & .60 & .47 & .46 & .49 & .51\\
\quad word accuracy & .46 & .12 & .41 & .12 & .29\\
\quad utterance accuracy & .29 & .14 & .31 & .07 & .22\\
\quad span F1 & .24 & .12 & .20 & .04 & .16\\
\multicolumn{6}{@{}l}{\emph{share of the best gain missed}}\\
\quad \textbf{check} & 8\% & 31\% & 16\% & 11\% & 16\%\\
\quad gain on dev & 12\% & 28\% & 16\% & 13\% & 17\%\\
\quad word accuracy & 9\% & 37\% & 10\% & 23\% & 19\%\\
\quad utterance accuracy & 22\% & 38\% & 33\% & 26\% & 29\%\\
\quad span F1 & 23\% & 38\% & 33\% & 26\% & 29\%\\
\midrule
\multicolumn{6}{@{}l}{\emph{check minus \ldots, pooled (95\% crossed type bootstrap)}}\\
 & \multicolumn{3}{c}{$\rho$} & \multicolumn{2}{c}{missed (pp)}\\
\quad gain on dev & \multicolumn{3}{r}{$-$.05 [$-$.24, .13]} & \multicolumn{2}{r}{$-$1 [$-$6, 7]}\\
\quad word accuracy & \multicolumn{3}{r}{+.17 [$-$.05, .34]} & \multicolumn{2}{r}{$-$3 [$-$10, 3]}\\
\quad utterance accuracy & \multicolumn{3}{r}{+.25 [$-$.002, .50]} & \multicolumn{2}{r}{$-$13 [$-$21, $-$1]}\\
\quad span F1 & \multicolumn{3}{r}{+.30 [.003, .59]} & \multicolumn{2}{r}{$-$13 [$-$22, $-$1]}\\
\bottomrule
\end{tabular}
\caption{The check against partner accuracy on all four datasets (own seed among the candidates). Word accuracy does as well as the check on MASSIVE and SNIPS (better on SNIPS in share missed) but is close to random on CoNLL and MIT-Restaurant; span F1 and utterance accuracy are weak everywhere. Pooled: the 72 primaries of the three datasets and the 15 of MIT-Restaurant. The crossed bootstrap resamples the 8 model types as primaries and as partners, with the same draw in every dataset; also resampling the 4 datasets widens the check-minus-word-accuracy interval to +.17 [$-$.11, .42] for $\rho$. A crossed random-effects model of the per-primary $\rho$ difference (random intercepts for primary type and dataset; $t$ interval with 7 df) gives +.17 [$-$.01, .35] against word accuracy and +.30 [.07, .53] against span F1. Without the own seed, the pooled check leads span F1 by +.16 [$-$.07, .41] and word accuracy by +.05 [$-$.18, .23].}
\label{tab:fourdata}
\end{table}

\begin{figure*}[tp]
  \centering
  \includegraphics[width=\textwidth]{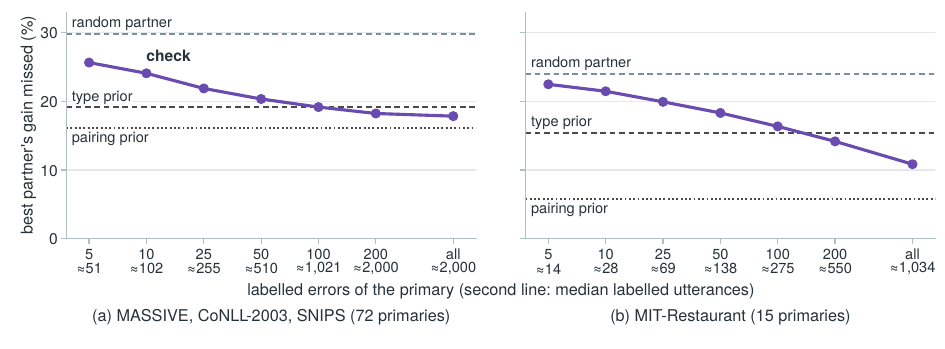}
  \caption{\textbf{About 100 labelled errors of the primary bring the check to the level of a prior learned on other datasets; with no labels, the prior alone recovers 81\% of the best gain.}
Share of the best partner's gain missed when each primary's partner is chosen by accuracy on $k$ of its labelled held-out errors (2,000 random draws per $k$).
Second tick line: median number of labelled utterances that contain $k$ errors, capped at the held-out slice size; all: every held-out error (170--586 per primary; 353--397 on MIT-Restaurant).
Dashed and dotted lines need no labels from the dataset: a random partner, and the partner type (type prior) or primary--partner pairing (pairing prior) with the highest mean gain on the other datasets.
The pairing prior is not beaten at any sample size. Post hoc analysis of saved outputs.
}
  \label{fig:labelbudget}
\end{figure*}

\paragraph{Label budget.} Table~\ref{tab:budget} and Figure~\ref{fig:labelbudget} count the labels the check needs in the primary's errors, the only utterances it uses. About 100 labelled errors (roughly 1,000 utterances at our error rates) match the type prior; the full slices hold 170--586 errors.

\begin{table}[tbp]
\centering\small\setlength{\tabcolsep}{3pt}
\begin{tabular}{@{}rrrrrr@{}}
\toprule
labelled & utterances & \multicolumn{2}{c}{3 datasets} & \multicolumn{2}{c}{MIT-Restaurant}\\
\cmidrule(lr){3-4}\cmidrule(l){5-6}
primary errors & (median) & missed & $\rho$ & missed & $\rho$\\
\midrule
5 & 51 & 26\% & .15 & 22\% & .08\\
10 & 102 & 24\% & .20 & 21\% & .12\\
25 & 255 & 22\% & .28 & 20\% & .17\\
50 & 510 & 20\% & .35 & 18\% & .23\\
100 & 1,021 & 19\% & .41 & 16\% & .30\\
200 & 2,000 & 18\% & .46 & 14\% & .38\\
all (170--586) & 2,000 & 18\% & .47 & 11\% & .40\\
\midrule
\multicolumn{2}{@{}l}{random partner} & 30\% & -- & 24\% & --\\
\multicolumn{2}{@{}l}{any but the own seed} & 26\% & .36 & 20\% & .35\\
\multicolumn{2}{@{}l}{type prior (no labels)} & 19\% & .37 & 15\% & .42\\
\multicolumn{2}{@{}l}{pairing prior (no labels)} & 16\% & .42 & 6\% & .57\\
\bottomrule
\end{tabular}
\caption{The label budget of the check, counted in the primary's labelled errors. For each primary we draw $k$ of its held-out errors (2,000 draws), rank candidates by their accuracy on those $k$, and choose the top one (ties split). Utterances: $k$ divided by the primary's held-out error rate (median over primaries; capped at the slice size). The full slices hold 408--586 primary errors on MASSIVE, 182--264 on CoNLL, 170--229 on SNIPS and 353--397 on MIT-Restaurant. With about 100 primary errors the check matches the type prior on the three datasets.}
\label{tab:budget}
\end{table}

\paragraph{Mixed pools and new types.} The type prior of a new partner type is built from that type's test gains on other datasets, so it presumes history a truly new type does not have. Table~\ref{tab:mixedpools} replaces it with a back-off prior that treats an unseen type as average. Prior plus check then misses 18\% (GLiNER-FT and XLM-R-XL added) and 20\% (prompted LLM added), against 30\% and 38\% for the check alone and 35\% for a random partner.

\begin{table}[tbp]
\centering\footnotesize\setlength{\tabcolsep}{1.5pt}
\begin{tabular}{@{}lrrr@{}}
\toprule
chosen by & \begin{tabular}[b]{@{}c@{}}grid\\(8 types)\end{tabular} & \begin{tabular}[b]{@{}c@{}}+ GLiNER-FT\\+ XLM-R-XL\end{tabular} & \begin{tabular}[b]{@{}c@{}}+ prompted\\LLM\end{tabular}\\
\midrule
random partner & 30\% & 35\% & 35\%\\
check alone & 18\% & 30\% & 38\%\\
type prior (history) & 19\% & 20\% & 19\%\\
type prior + check & 15\% & 17\% & 15\%\\
back-off prior & 19\% & 20\% & 19\%\\
back-off prior + check & 15\% & 18\% & 20\%\\
\bottomrule
\end{tabular}
\caption{Mixed candidate pools: share of the best gain missed (ties split). The type prior of a new type (GLiNER-FT, XLM-R-XL, the prompted LLM) uses that type's test gains on the other datasets, so it assumes partner history. The back-off prior gives a type with no history the mean prior of the primary's other candidates except its own seed. Pools: MASSIVE and CoNLL for GLiNER-FT and XLM-R-XL (48 primaries, 10 candidates); all three datasets for the LLM (72 primaries, 9 candidates). In the LLM pool the check alone misses more than a random partner (difference +3 [$-$19, 9] pp, two-way type bootstrap); back-off prior + check misses less than random ($-$15 [$-$25, $-$1] pp).}
\label{tab:mixedpools}
\end{table}

\begin{figure*}[tp]
  \centering
  \includegraphics[width=\textwidth]{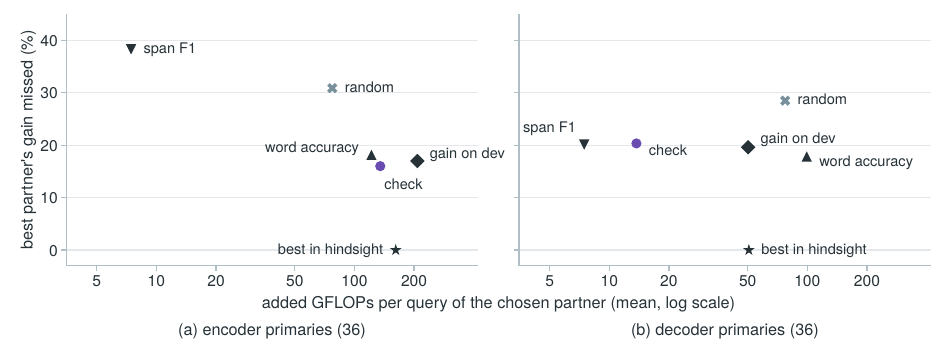}
  \caption{\textbf{For encoder primaries, the good ways of choosing mostly pick expensive decoder partners; for decoder primaries, the check almost always picks a cheap encoder.}
Share of the best partner's gain missed against the mean added compute of the chosen partner (all 8 partner types are candidates).
Choosing by span F1 is cheap for encoder primaries only because it never picks a decoder, and it misses much more gain.
Restricting the candidates to cheaper partners is analysed in Table~\ref{tab:cost}.
Three datasets, post hoc; SNIPS costs use MASSIVE's.
}
  \label{fig:costpolicies}
\end{figure*}

\paragraph{Cost classes.} Table~\ref{tab:cost} repeats the selection inside cost classes, with inference FLOPs per query as the cost; Figure~\ref{fig:costpolicies} plots the gain missed against the chosen partner's compute.

\begin{table}[tbp]
\centering\small\setlength{\tabcolsep}{2.4pt}
\begin{tabular}{@{}llrrrr@{}}
\toprule
candidates & chosen by & \multicolumn{2}{c}{missed} & FLOP & decoder\\
\cmidrule(lr){3-4}
 & & of class & of all & ratio & chosen\\
\midrule
\multicolumn{6}{@{}l}{\emph{encoder primaries (36)}}\\
\quad all 8 & check & 16\% & 16\% & 22.9 & 65\%\\
\quad  & random & 31\% & 31\% & 14.4 & 50\%\\
\quad  & hindsight & 0\% & 0\% & 31.7 & 100\%\\
\quad encoders only & check & 15\% & 35\% & 1.4 & 0\%\\
\quad  & random & 28\% & 45\% & 1.3 & 0\%\\
\quad  & hindsight & 0\% & 24\% & 1.4 & 0\%\\
\quad $\leq2\times$ FLOPs & check & 16\% & 49\% & 1.0 & 6\%\\
\quad  & random & 24\% & 53\% & 0.9 & 7\%\\
\quad  & hindsight & 0\% & 38\% & 1.1 & 19\%\\
\addlinespace[2pt]
\multicolumn{6}{@{}l}{\emph{decoder primaries (36)}}\\
\quad all 8 & check & 20\% & 20\% & 0.3 & 4\%\\
\quad  & random & 28\% & 28\% & 1.4 & 50\%\\
\quad  & hindsight & 0\% & 0\% & 2.0 & 39\%\\
\quad encoders only & check & 16\% & 20\% & 0.2 & 0\%\\
\quad  & random & 18\% & 23\% & 0.1 & 0\%\\
\quad  & hindsight & 0\% & 5\% & 0.1 & 0\%\\
\quad $\leq2\times$ FLOPs & check & 19\% & 21\% & 0.2 & 1\%\\
\quad  & random & 27\% & 28\% & 0.4 & 39\%\\
\quad  & hindsight & 0\% & 2\% & 0.3 & 22\%\\
\bottomrule
\end{tabular}
\caption{Choosing a partner within a cost class (three datasets). FLOP ratio: the chosen partner's inference FLOPs per query over the primary's (mean; SNIPS uses the MASSIVE token counts). Missed: shortfall from the best partner in the class as a share of that partner's gain (of class), or from the best of all 8 candidates as a share of its gain (of all). For every encoder primary the best partner is a decoder (36/36); the best encoder partner gains 5.52 against 7.29$\times10^{-3}$ \eaurc{} for the best decoder. Inside the encoder class the check stays ahead of random by point estimate ($\rho=.41$; regret 0.80 against 1.52$\times10^{-3}$ for random, difference $-$.72 [$-$1.44, .35]). For decoder primaries the check already picks an encoder partner in 34.5 of 36 cases (ties split).}
\label{tab:cost}
\end{table}

\section{Prospective replication on MIT-Restaurant}
\label{app:prospective}

\paragraph{What was frozen.} Before any data of the new dataset was downloaded, we wrote a protocol that fixes the dataset-choice rule, the splits, the five tagger types and their recipes, the selection rule of \S\ref{sec:curve}, the combiner and metric, and three success criteria, and stored its SHA-256 hash (6dfd2020acbc\ldots) together with the hash of the decision code. The protocol also records what we knew: on the three old datasets, the same five-type design would have passed on MASSIVE ($\rho=0.55$) and CoNLL (0.59) and failed the interval criterion on SNIPS (0.39 [$-0.07$, 0.63]).

\paragraph{Dataset choice.} The rule took the first of MIT-Restaurant, MultiCoNER~v2 (English) and MIT-Movie whose terms permit research use and that can be downloaded anonymously, judged from licence and availability metadata only. MIT-Restaurant \citep{liu2013asgard} qualified on its creators' research-use statement. It has 7,660 training and 1,521 test queries with 8 slot types; we carved a seeded dev set (766) and fit slice (1,034) from the training file, leaving 5,860 for training.

\paragraph{Models.} XLM-R-large and DeBERTa-v3-large (full fine-tuning) and Gemma-4-E2B, Qwen2.5-3B and Qwen2.5-0.5B echo taggers, with the MASSIVE/SNIPS recipes and 3 seeds each. Each primary has five partners (its second seed and the other four types): 75 pairings. All 15 runs trained without a fallback (dev span F1 0.788--0.813).

\begin{table}[tbp]\centering\small\setlength{\tabcolsep}{4pt}
\begin{tabular}{@{}lcc@{}}\toprule
tagger & test span F1 & utterance error (\%)\\\midrule
XLM-R-large & 0.802--0.816 & 35.4--36.9\\
Gemma-4-E2B echo & 0.799--0.803 & 36.4--36.9\\
Qwen2.5-3B echo & 0.792--0.801 & 36.2--37.6\\
DeBERTa-v3-large & 0.806--0.810 & 35.5--37.2\\
Qwen2.5-0.5B echo & 0.771--0.798 & 37.7--41.0\\
\bottomrule\end{tabular}
\caption{MIT-Restaurant taggers on test (range over 3 seeds).}\label{tab:prosp_models}
\end{table}

\paragraph{The pre-declared test.} C1: the pooled within-primary Spearman correlation between held-out accuracy on the primary's errors and test gain is positive with a 95\% interval excluding zero (resampling the five tagger types and the test utterances). C2: it exceeds that of held-out F1. C3: picking the top-ranked partner loses less than a random partner. Result: $\rho=0.40$ [0.08, 0.59] (C1 holds); held-out F1 0.04 [$-0.24$, 0.43], difference 0.36 [$-0.09$, 0.62] (C2 holds by point estimate); loss 10\% against 24\% for a random partner, difference $1.56\times10^{-3}$ [0.40, 2.78] (C3 holds; the decision code took the first tied partner, and with ties split, the rule used elsewhere, the loss is 11\%). The outcome is a replication. The rule picked the best partner for 27\% of the 15 primaries (chance 20\%); always taking the second seed loses 41\%. The check's advantage over the second seed was not a pre-declared criterion; under the same resampling it is $3.44\times10^{-3}$ [0.87, 6.21] (post hoc). Resampling partner types as well (Table~\ref{tab:policiesfull}) widens this and the random-partner interval to include zero.

\begin{table}[tbp]\centering\small\setlength{\tabcolsep}{4pt}
\begin{tabular}{@{}lccc@{}}\toprule
held-out quantity (fit slice) & $\rho$ & 95\% CI & lost\\\midrule
accuracy on A's errors (the rule) & .40 & [.08, .59] & 10\%\\
\quad double fault & .40 & [.08, .59] & 10\%\\
error correlation $\phi$ & .36 & [.01, .54] & 24\%\\
B wrong where A is right & $-$.22 & [$-$.42, .14] & --\\
F1 gap & .04 & [$-$.24, .43] & 26\%\\
AUROC of B's margin & .09 & [$-$.28, .35] & 24\%\\
margin drop & .37 & [.06, .57] & 8\%\\
unlabelled disagreement & .29 & [$-$.05, .48] & 30\%\\
gain on dev (reference) & .49 & [.11, .64] & 13\%\\
\bottomrule\end{tabular}
\caption{Prospective test, all candidate quantities (descriptive except the first row). Lost: shortfall from the best partner's gain when picking the top-ranked partner (first tied partner, as pre-specified; with ties split the first two rows lose 11\%).}\label{tab:prosp_pred}
\end{table}

\paragraph{Partners help.} 68 of 75 pairings have a gain whose 95\% interval lies above zero and none below. Mean gains are 8 points (second seed), 12 (same pretraining) and 11 (other type).

\paragraph{Robustness.} With dev as the held-out slice the correlation is 0.33 (F1: 0.34). Leaving out one partner type, out-of-sample $R^2$ is 0.20 for accuracy on errors and $-0.66$ for F1. Under the raw combiner, AUGRC, and both, the outcome is also a replication (0.40, 0.38, 0.42). By primary type the correlation ranges from 0.15 (Qwen-3B) to 0.79 (DeBERTa).
Dropping any one of the five types keeps $\rho$ at 0.30--0.46.

\paragraph{Other rules.} The expected margin drop, another error-conditioned measure, misses 8\%, and always pairing with Qwen2.5-3B, a choice made in hindsight, misses 10\%. The frozen rule beats a random partner, the own seed and held-out F1.

\subsection{The frozen label-aware test}
\label{app:slabmitr}

\paragraph{What was frozen, and when.} $\slab$ and its decision rule D-LA were hashed at 04:06 UTC (SHA-256 627bbe14\ldots), before MIT-Restaurant was chosen or downloaded. D-LA passes if (i) the mean extra gain of $\slab$ over $\savg$ has a two-way 95\% interval above zero, and (ii) the check's Spearman correlation with $\slab$ gains is at least 0.424.

\paragraph{Application rule.} Before scoring, a hashed application file (SHA-256 45ec9f78\ldots) copied the score and the rule verbatim, kept every setting of the prospective test, and named the pairings that count. One primary, Gemma seed 0, had already been scored with $\slab$ for Figure~\ref{fig:ex-conll-mitr}, so the test uses the other 70 pairings. As a gate, the $\savg$ gains reproduced the prospective test's saved gains and bootstrap replicates exactly.

\begin{table}[tbp]\centering\footnotesize\setlength{\tabcolsep}{2.5pt}
\begin{tabular}{@{}lcccc@{}}\toprule
pairings & $\savg$ & $\slab$ & extra [95\% CI] & better\\\midrule
70 unseen (test) & 11.03 & 13.73 & 2.70 [1.70, 3.96] & 68/70\\
all 75 & 10.90 & 13.50 & 2.60 [1.63, 3.82] & 73/75\\
5 seen (Gemma s0) & 9.02 & 10.29 & 1.26 [$-$0.53, 3.10] & 5/5\\
\midrule\multicolumn{5}{@{}l}{\emph{the 70, by primary type}}\\
\quad XLM-R & 10.15 & 12.09 & 1.93 [1.01, 2.88] & 15/15\\
\quad Gemma & 8.35 & 10.79 & 2.45 [1.24, 3.79] & 10/10\\
\quad Qwen-3B & 8.56 & 11.00 & 2.43 [1.18, 3.78] & 14/15\\
\quad DeBERTa & 11.25 & 13.51 & 2.26 [1.29, 3.32] & 15/15\\
\quad Qwen-0.5B & 15.96 & 20.30 & 4.34 [2.97, 5.82] & 14/15\\
\midrule\multicolumn{5}{@{}l}{\emph{the 70, by partner}}\\
\quad own seed & 8.48 & 10.07 & 1.59 [0.46, 2.73] & 12/14\\
\quad same pretraining & 11.99 & 15.47 & 3.48 [1.54, 5.47] & 6/6\\
\quad other type & 11.63 & 14.55 & 2.92 [1.77, 4.33] & 50/50\\
\bottomrule\end{tabular}
\caption{Frozen label-aware test on MIT-Restaurant. Mean gain in points with the averaged margin ($\savg$) and the label-aware score ($\slab$), and the extra gain from $\slab$ (two-way cluster bootstrap over primary and partner types). Better: pairings that improve. Unseen: never scored with $\slab$ before. Gemma s0: Gemma, seed 0. The first row is the pre-declared test; the others are descriptive.}
\label{tab:slabmitr}
\end{table}

\paragraph{Result.} Both parts of D-LA hold (Table~\ref{tab:slabmitr}). On the 70 pairings, $\slab$ adds 2.70 points [1.70, 3.96], or $2.05\times10^{-3}$ [1.26, 3.09] in absolute \eaurc{}. All 50 pairings with a partner of another type improve; the two that do not are own-seed pairings ($-1.12$ and $-0.03$ points). For (ii), the correlation is 0.46 [0.14, 0.64], against 0.39 with $\savg$. The score and the absolute success threshold were fixed before dataset selection. A later application file, hashed before scoring, restated criterion (ii) relative to $\savg$ on the same pairings; the observed result satisfies both the original absolute threshold and the later relative reading. A separately written verifier, with its own \eaurc{} code, reproduced every number.

\section{Partner prior and labelled check}
\label{app:priorcheck}

\begin{figure*}[tp]
  \centering
  \includegraphics[width=\textwidth]{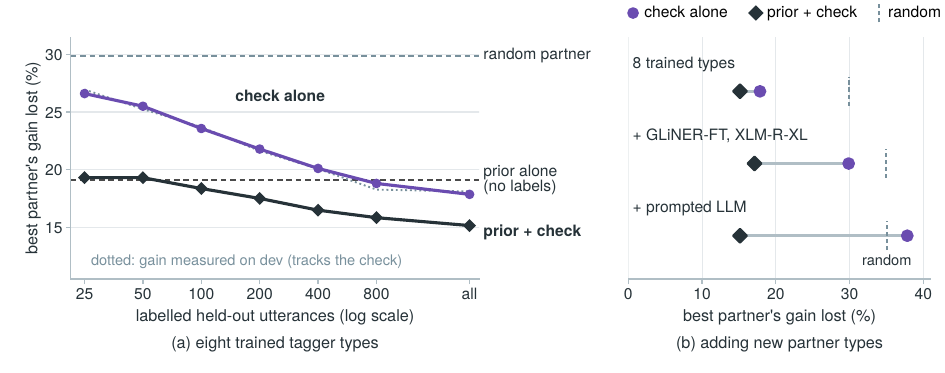}
  \caption{\textbf{A prior learned on other datasets makes the check robust, most of all when new partner types join the pool.}
(a) With more labelled held-out utterances, the check alone loses less of the best partner's gain; combining its ranks with the prior loses less than the check alone at every size, significantly so at 25--50 utterances (Appendix~\ref{app:prereg} lists the pre-specified 400-utterance criterion).
(b) All held-out utterances: when the candidates include types whose accuracy on errors overstates their gain, the check alone loses much more (with the LLM among the candidates, more than a random choice), while prior plus check does not; the prior uses the new types' gains on the other datasets.
Averaged-margin gains; 72 primaries (48 in the middle row of b); dashed ticks in (b): a random partner.
}
  \label{fig:priorcheck}
\end{figure*}

\paragraph{Policies.} The \emph{prior} scores each partner type (own seed, or a named type) by its mean test gain on the other two datasets, over primaries of other types, so it uses no labels of the target dataset. The \emph{check} is held-out accuracy on the primary's errors measured on $n$ labelled utterances drawn at random from the held-out slice (200 subsets per size). \emph{Prior plus check} averages the two within-primary ranks with equal weight, as fixed in the protocol. Each policy picks its top-scoring partner; ties count as a random choice among the tied partners.

\paragraph{Results by sample size.} With no labels, the prior misses 19\% of the best partner's gain ($\rho=0.37$). The check alone misses 27\% at 25 utterances and 18\% on all held-out utterances, and is never significantly better than the prior. Prior plus check misses 19\% at 25 and 15\% on all held-out utterances. Its advantage over the check is significant at 25 utterances ($-0.46\times10^{-3}$ [$-0.92$, $-0.03$]) and 50 ($-0.39$ [$-0.78$, $-0.02$]) but not at 400 ($-0.23$ [$-0.54$, $+0.08$]), so the pre-specified rule failed. Under the label-aware outcome, the advantage is also significant at 100 and 200 utterances, but still not at 400 ($-0.24$ [$-0.54$, $+0.06$]). Measuring the gain on dev tracks the check at every size.

\paragraph{New partner types.} With fine-tuned GLiNER and XLM-R-XL as extra candidates (MASSIVE and CoNLL), the check alone has regret 2.11 against 1.21 for prior plus check (difference $-0.90$ [$-1.74$, $-0.20$]); with the prompted LLM as an extra candidate, 2.40 against 0.96 ($-1.44$ [$-2.20$, $-0.73$]), and the check alone does worse than a random partner (2.23). The prior of a new type uses that type's gains on the other datasets; a back-off prior without such history is in Appendix~\ref{app:policies}.

\section{The label-aware score}
\label{app:labelaware}

\begin{figure*}[tp]
  \centering
  \includegraphics[width=\textwidth]{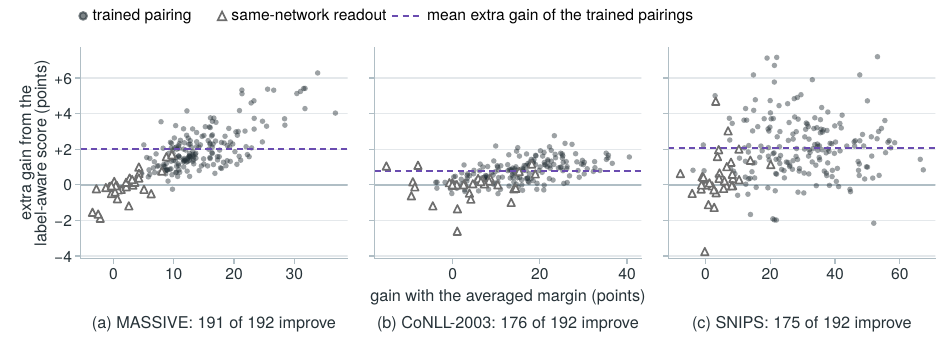}
  \caption{\textbf{Scoring the primary's own label raises the gain of almost every trained partner.}
Dots: the 576 trained primary--partner pairings; triangles: same-network readouts.
x: gain with the averaged margin; y: extra gain when the label-aware score replaces it (both in points); dashed line: mean extra gain of the trained pairings.
542 of 576 pairings improve; the readouts scatter around zero.
}
  \label{fig:labelaware}
\end{figure*}

\paragraph{Protocol.} The protocol fixed $\slab$, a second variant, the decision rule D-LA, and the analyses below before any number of this analysis was computed; we had seen the score help in 39 pairings, so the test is pre-specified but not blind. The frozen test on MIT-Restaurant is in Appendix~\ref{app:slabmitr}. The unsigned pipeline reproduces the earlier grid gains and the held-out statistic (0.474) exactly.

\begin{table}[tbp]\centering\small\setlength{\tabcolsep}{2pt}
\begin{tabular}{@{}lccc@{}}\toprule
subset & $n$ & extra gain [95\% CI] & improve\\\midrule
all & 576 & $+0.41$ [0.30, 0.53] & 94\%\\
MASSIVE & 192 & $+0.83$ [0.56, 1.13] & 99\%\\
CoNLL & 192 & $+0.18$ [0.10, 0.28] & 92\%\\
SNIPS & 192 & $+0.23$ [0.11, 0.37] & 91\%\\
own-seed partners & 72 & $+0.26$ [0.14, 0.38] & 93\%\\
other pretraining & 432 & $+0.45$ [0.32, 0.59] & 94\%\\
encoder primaries & 288 & $+0.37$ [0.19, 0.58] & 91\%\\
decoder primaries & 288 & $+0.46$ [0.34, 0.60] & 97\%\\
\bottomrule\end{tabular}
\caption{Extra gain from the label-aware score ($\times10^{-3}$ \eaurc{}; two-way cluster bootstrap). The held-out test statistic with $\slab$ gains is 0.483 (with $\savg$: 0.474); AUGRC $+0.36$ [0.27, 0.46].}
\label{tab:labelaware}
\end{table}

\paragraph{What changes.} The averaged distribution overrules the primary on 27--43\% of its wrong utterances, and on 41--48\% of those it looks more confident than the primary alone. The label-aware score flags these utterances. In points, it adds 2.0, 0.8 and 2.1 on MASSIVE, CoNLL and SNIPS.

\paragraph{What does not change.} Partner rankings (Spearman between the two scores 0.97, minimum 0.83; the best partner is unchanged for 83\% of primaries). Same-network verdicts (Table~\ref{tab:readoutsci}): the linear copy-1 probe goes from $p=0.07$ to $0.14$, the linear second head stays at or below zero, the MC-dropout mean stays at zero. The new-type residuals and the decoder-primary $\times$ XLM-R anomaly are unchanged. One-hot versions of the grid partners gain $+0.21$ [0.13, 0.31] from the label-aware score, but only about half as much as the same partners with probabilities.

\begin{table*}[tp]\centering\small\setlength{\tabcolsep}{4pt}
\begin{tabular}{llcccc}\toprule
 & & \multicolumn{2}{c}{averaged margin $\savg$} & \multicolumn{2}{c}{label-aware $\slab$}\\\cmidrule(lr){3-4}\cmidrule(l){5-6}
primary & second opinion & gain [95\% CI] & $p$ & gain [95\% CI] & $p$\\\midrule
Gemma & MC-dropout mean & +0.02 [$-$0.04, 0.08] & 0.505 & +0.02 [$-$0.04, 0.09] & 0.458\\
 & second head, linear & $-$0.68 [$-$1.59, 0.24] & 0.136 & $-$0.77 [$-$1.72, 0.17] & 0.107\\
 & second head, MLP & +0.97 [$-$0.13, 2.16] & 0.086 & +1.04 [$-$0.14, 2.33] & 0.091\\
 & copy 1, linear probe & +0.78 [$-$0.07, 1.70] & 0.071 & +0.76 [$-$0.26, 1.84] & 0.143\\
 & copy 1, MLP probe & +1.63 [0.74, 2.62] & 0.000 & +1.67 [0.66, 2.77] & 0.002\\
 & copy 1, primary's own head & +0.92 [0.28, 1.64] & 0.004 & +0.86 [0.15, 1.64] & 0.020\\
 & second seed (reference) & +3.00 [1.68, 4.52] & 0.000 & +3.28 [1.90, 4.87] & 0.000\\
\midrule
Qwen-3B & MC-dropout mean & +0.00 [$-$0.08, 0.09] & 0.974 & $-$0.00 [$-$0.09, 0.08] & 0.977\\
 & second head, linear & $-$0.28 [$-$1.20, 0.72] & 0.537 & $-$0.27 [$-$1.30, 0.82] & 0.590\\
 & second head, MLP & +1.24 [0.50, 2.00] & 0.000 & +1.37 [0.58, 2.20] & 0.000\\
 & copy 1, linear probe & +0.43 [$-$0.22, 1.06] & 0.185 & +0.41 [$-$0.32, 1.14] & 0.264\\
 & copy 1, MLP probe & +1.76 [1.08, 2.52] & 0.000 & +1.90 [1.15, 2.72] & 0.000\\
 & copy 1, primary's own head & +0.39 [$-$0.10, 0.92] & 0.120 & +0.26 [$-$0.34, 0.89] & 0.419\\
 & second seed (reference) & +2.63 [1.85, 3.42] & 0.000 & +2.92 [2.08, 3.77] & 0.000\\
\bottomrule\end{tabular}
\caption{\textbf{Same-network readouts gain at most two thirds of a second seed's gain, and the label-aware score changes no verdict.} Readouts of the two echo primaries as second opinions, with intervals (gain $\times10^{-3}$ \eaurc{}, positive = helps; mean over MASSIVE, CoNLL and SNIPS and 3 seeds). Probes and heads are trained on the frozen states of the fine-tuned primary. The label-aware score changes no verdict at the 5\% level. The MC-dropout mean gains nothing; the MLP readouts gain a third to two thirds of a second seed's gain.}
\label{tab:readoutsci}\end{table*}

\section{The prompted LLM partner}
\label{app:llm}

\paragraph{Setup.} Qwen2.5-7B-Instruct \citep{qwen2025}, served with vLLM \citep{kwon2023vllm} in bf16 with greedy decoding. The system prompt gives the task, the allowed types (with CoNLL's type definitions) and the output format, a JSON list of \{text, type\}. Eight training utterances, retrieved by TF-IDF similarity (word 1--2-grams), are given as worked examples, most similar last. Output spans are aligned to the input by exact case-insensitive word match (left to right, first unused match); at most 2.1\% of output spans are dropped. The partner's word distribution is one-hot on its labels. The prompt was fixed after a pilot on 150 training utterances per dataset, before any fit, dev or test utterance was sent to the model. The fit-slice and dev outputs were then used to write and hash the predicted test gain, from the within-primary line of test gain on held-out accuracy on errors fitted to the trained partners; only then were test utterances sent.

\begin{table*}[tp]\centering\small\setlength{\tabcolsep}{3.5pt}
\begin{tabular}{lccccc}\toprule
 & & & \multicolumn{3}{c}{test gain $\times10^{-3}$ \eaurc{} (label-aware score)}\\\cmidrule(l){4-6}
dataset & LLM F1 & acc.\ on errors: LLM / best trained & LLM [95\% CI] & predicted in advance & second seed\\\midrule
MASSIVE & 0.71 & 0.28 / 0.26 & 4.49 [2.39, 6.78] & 8.93 & 4.89\\
CoNLL & 0.84 & 0.27 / 0.37 & 2.54 [1.16, 4.11] & 3.53 & 2.01\\
SNIPS & 0.84 & 0.29 / 0.41 & 1.38 [$-$0.23, 3.19] & 3.42 & 2.66\\
\bottomrule\end{tabular}
\caption{\textbf{A prompted LLM with a lower span F1 than every trained tagger matches a second seed on MASSIVE and exceeds it on CoNLL.} The LLM is a label-only partner (Qwen2.5-7B-Instruct, 8 retrieved examples; means over 24 primaries). Accuracy on errors is measured on the held-out slice; ``best trained'' is the best grid partner of each primary, averaged. Pooled gain: 2.80 [1.73, 3.97] with the label-aware score and 1.50 [0.80, 2.28] with the averaged margin. The prediction was written and hashed before any test utterance was sent to the LLM.}
\label{tab:llm}\end{table*}

\paragraph{Results.} Table~\ref{tab:llm}. The LLM beats a second seed for 46\%, 67\% and 12\% of primaries (MASSIVE, CoNLL, SNIPS). The label-aware score adds $+1.30\times10^{-3}$ [0.82, 1.80] over the averaged margin for the LLM, more than for any trained partner. With the LLM among 9 candidates the plain rule picks it for 29\% of primaries although it is never the best; regret rises from 1.17 to 2.22. Post hoc, a line fitted to one-hot versions of the trained partners predicts 5.1 on MASSIVE (observed 4.5) and 1.7 on SNIPS (1.4), but 1.0 on CoNLL (2.5); ranking the LLM against this line and probability partners against the usual one restores regret 1.17.

\section{The test-set curve}
\label{app:curve}

\begin{figure*}[tp]
  \centering
  \includegraphics[width=\textwidth]{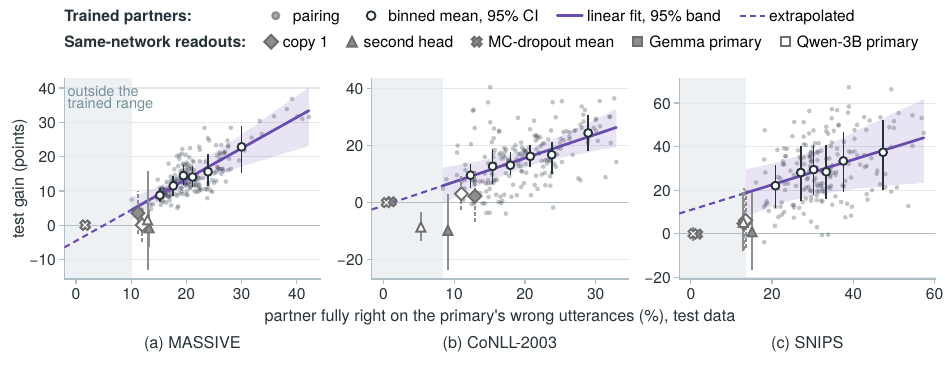}
  \caption{\textbf{On test data, gain rises with how often the partner is right on the primary's errors; same-network readouts sit at the low end.}
Dots: the 576 trained primary--partner pairings; open circles: binned means with 95\% intervals; line and band: linear fit with a 95\% cluster-bootstrap band over the range of trained partners, extended (dashed) into the shaded range they do not cover.
Grey markers: same-network readouts of the two echo primaries (filled: Gemma; hollow: Qwen-3B), with 95\% intervals.
This is a test-set relation; the held-out test is in Figure~\ref{fig:central}.
}
  \label{fig:curve}
\end{figure*}

\begin{table}[tbp]
\centering\small\setlength{\tabcolsep}{4pt}
\begin{tabular}{lccc}
\toprule
curve model & MASSIVE & CoNLL & SNIPS\\
\midrule
F1 gap (spline) & .09 & .08 & .02\\
accuracy on errors (linear) & .38 & .32 & .15\\
both & .49 & .36 & .19\\
\bottomrule
\end{tabular}
\caption{Share of the variation in test gain explained for a fixed primary (within-primary $R^2$), with both quantities measured on the test set. Partner F1 differs from the primary's by at most 9.3, 1.6 and 2.4 points, so part of the contrast is range.}
\label{tab:curvefit}
\end{table}

\paragraph{Readouts and the curve.} Copy 1 is not distinguishable from the accuracy-on-errors curve under any curve form with an intercept; only a curve forced through the origin puts it below, and that form fits SNIPS worse than no curve. The linear second head is below the curve under four of six forms (six of six with the crossed bootstrap); the MLP head is not distinguishable from it. The MC mean is outside the trained range (right on 0.4--1.7\% of errors; trained partners 8--57\%), so its place depends only on the curve's form.

\section{Same-network readouts in detail}
\label{app:same}

\begin{table*}[tp]
\centering\small\setlength{\tabcolsep}{4pt}
\begin{tabular}{lrrrrl}
\toprule
test (Gemma primary unless stated) & MASSIVE & CoNLL & SNIPS & pooled & 95\% CI\\
\midrule
copy-1 average vs margin, rank-ECDF & $+1.28$ & $+0.47$ & $+0.60$ & $+0.78$ & [$+0.13$, $+1.46$]\\
\quad same, one item resample shared by all seeds & & & & & [$-0.06$, $+1.69$]\\
\quad as share of an own-seed gain & & & & 0.26 & [0.05, 0.49]\\
\quad raw combiner & & & & $-1.37$ & [$-2.54$, $-0.15$]\\
\quad fixed equal weights (no fitting) & & & & $-0.66$ & [$-1.76$, $+0.45$]\\
\quad Qwen-3B primary (exploratory) & & & & $+0.43$ & [$-0.05$, $+0.89$]\\
copy-1 probe disagreement vs margin & & & & $+0.61$ & [$+0.13$, $+1.14$]\\
\quad raw combiner & & & & $-1.35$ & \\
MC-dropout mean ($k=8$) vs margin & & & & $+0.02$ & [$-0.03$, $+0.07$]\\
second-head average vs margin & & $-1.97$ & & $-0.68$ & [$-1.41$, $+0.05$]\\
\quad raw combiner & & & & $-2.92$ & \\
disagreement beyond the averaged pair & & & $-0.14$ & & \\
\bottomrule
\end{tabular}
\caption{Pooled estimates for the same-network nulls (gain in \eaurc{} $\times10^{-3}$, positive = helps). Pooled = equal-weight mean over the three datasets with the seed-and-item bootstrap (separate item draws per seed, except the shared-draw row). The MC-dropout mean and the disagreement term are equivalent to zero within 0.001 (90\% interval) on every dataset; the copy-1 average is not shown equivalent within half an own-seed gain once items are shared. Under random effects across the three datasets an own-seed partner's pooled gain is $+3.00$ [$-0.10$, $+6.54$]; the per-dataset equivalence tests carry the nulls.}
\label{tab:equiv}
\end{table*}

\begin{table*}[tp]
\centering\small\setlength{\tabcolsep}{3pt}
\begin{tabular}{llcccc}
\toprule
 & & \multicolumn{2}{c}{$k=1$ pass} & \multicolumn{2}{c}{$k=32$ passes}\\
\cmidrule(lr){3-4}\cmidrule(l){5-6}
tagger & dropout & vs margin & vs own seed & vs margin & vs own seed\\
\midrule
Qwen-3B echo & trained rate & $+0.0$ {\scriptsize[$-0.2$, $+0.2$]} & $-3.1$ {\scriptsize[$-5.1$, $-1.2$]} & $+0.0$ {\scriptsize[$+0.0$, $+0.1$]} & $-3.1$ {\scriptsize[$-5.0$, $-1.1$]}\\
 & S1 & $+0.2$ {\scriptsize[$-0.1$, $+0.5$]} & $-2.8$ {\scriptsize[$-4.9$, $-0.6$]} & $+1.2$ {\scriptsize[$+0.1$, $+2.3$]} & $-1.9$ {\scriptsize[$-4.6$, $+0.8$]}\\
 & S2 & $+0.9$ {\scriptsize[$+0.1$, $+1.7$]} & $-2.2$ {\scriptsize[$-4.6$, $+0.3$]} & $+1.4$ {\scriptsize[$+0.0$, $+2.8$]} & $-1.7$ {\scriptsize[$-4.8$, $+1.4$]}\\
 & S3 & $+0.5$ {\scriptsize[$-0.2$, $+1.3$]} & $-2.3$ {\scriptsize[$-4.8$, $+0.1$]} & $+1.6$ {\scriptsize[$+0.3$, $+2.9$]} & $-1.5$ {\scriptsize[$-4.6$, $+1.5$]}\\
\midrule
Gemma echo & trained rate & $+0.0$ {\scriptsize[$+0.0$, $+0.0$]} & $-3.2$ {\scriptsize[$-5.5$, $-0.8$]} & $+0.0$ {\scriptsize[$+0.0$, $+0.1$]} & $-3.1$ {\scriptsize[$-5.5$, $-0.8$]}\\
 & S1 & $+0.3$ {\scriptsize[$-0.1$, $+0.8$]} & $-2.7$ {\scriptsize[$-5.1$, $-0.3$]} & $+0.8$ {\scriptsize[$+0.1$, $+1.6$]} & $-2.2$ {\scriptsize[$-4.8$, $+0.3$]}\\
 & S2 & $+0.5$ {\scriptsize[$-0.3$, $+1.2$]} & $-2.6$ {\scriptsize[$-5.4$, $+0.1$]} & $+1.0$ {\scriptsize[$+0.3$, $+1.8$]} & $-2.0$ {\scriptsize[$-4.6$, $+0.5$]}\\
 & S3 & $+0.2$ {\scriptsize[$-0.3$, $+0.7$]} & $-3.1$ {\scriptsize[$-5.8$, $-0.5$]} & $+1.0$ {\scriptsize[$+0.2$, $+1.7$]} & $-2.1$ {\scriptsize[$-4.8$, $+0.6$]}\\
\midrule
Qwen-3B one copy, no mask & trained rate & $+0.0$ {\scriptsize[$-0.2$, $+0.1$]} & $-3.6$ {\scriptsize[$-5.4$, $-1.8$]} & $+0.0$ {\scriptsize[$+0.0$, $+0.1$]} & $-3.5$ {\scriptsize[$-5.4$, $-1.7$]}\\
 & S1 & $+0.8$ {\scriptsize[$+0.1$, $+1.6$]} & $-2.6$ {\scriptsize[$-4.6$, $-0.5$]} & $+1.2$ {\scriptsize[$+0.4$, $+2.0$]} & $-2.3$ {\scriptsize[$-4.1$, $-0.5$]}\\
 & S2 & $+0.5$ {\scriptsize[$-0.4$, $+1.5$]} & $-3.0$ {\scriptsize[$-5.4$, $-0.5$]} & $+1.2$ {\scriptsize[$+0.5$, $+1.8$]} & $-2.1$ {\scriptsize[$-3.7$, $-0.6$]}\\
 & S3 & $+0.5$ {\scriptsize[$+0.0$, $+1.1$]} & $-2.7$ {\scriptsize[$-4.2$, $-1.2$]} & $+1.4$ {\scriptsize[$+0.7$, $+2.0$]} & $-2.0$ {\scriptsize[$-3.5$, $-0.4$]}\\
\bottomrule
\end{tabular}
\caption{Stronger MC dropout as a same-network partner, pooled over MASSIVE and CoNLL (gain in \eaurc{} $\times10^{-3}$, positive = helps; Knapp--Hartung 95\% CI over 6 seed--dataset units). ``vs margin'': gain from adding the MC mean to the margin. ``vs own seed'': gain of the MC partner minus that of a separately trained seed (negative = MC is worse). S1: dropout 0.10 on all LoRA modules and attention probabilities; S2: 0.20; S3: S1 plus 0.10 on every attention and MLP output.}
\label{tab:mcstrong}
\end{table*}

\paragraph{How often readouts are right.} For the LoRA-tuned echo primaries, readouts are right on 11--14\% of the primary's errors, against 19--24\% for trained partners, and the MC-dropout mean is wrong on 98--100\% of them; it is equivalent to no second opinion within 0.001 \eaurc{} (Table~\ref{tab:equiv}). The MLP second head gains 5.3 points. None of 24 stronger dropout settings beats a trained seed (Table~\ref{tab:mcstrong}).

\paragraph{What stronger dropout does.}
The stronger settings S1--S3 apply dropout rates the LoRA adapters were not trained with.
With the trained dropout, the 32-pass MC mean decodes almost exactly like the deterministic network (word agreement 0.998--1.000). At S2 and S3 a single pass agrees on only 0.897--0.991 of words and is 3--29 F1 points worse.

\paragraph{Not specific to echo taggers.} The no-mask single-copy Qwen-3B tagger reaches 0.9228 and 0.7924 F1 on CoNLL and MASSIVE (echo: 0.9304, 0.7964) with half the input tokens; kept causal, it loses 9--19 F1 points (Table~\ref{tab:singlecopyh2h}).

\begin{table*}[tp]
\centering\small\setlength{\tabcolsep}{4pt}
\begin{tabular}{lcccc}
\toprule
 & \multicolumn{2}{c}{one copy, no causal mask} & \multicolumn{2}{c}{one copy, causal}\\
\cmidrule{2-3}\cmidrule{4-5}
primary & vs margin & vs echo partner & vs margin & vs echo partner\\
\midrule
XLM-R & $+7.2$ {\scriptsize[$+6.2$, $+8.2$]} & $-0.5$ {\scriptsize[$-1.7$, $+0.7$]} & $+3.0$ {\scriptsize[$+1.4$, $+4.6$]} & $-4.7$ {\scriptsize[$-5.7$, $-3.7$]}\\
Gemma echo & $+3.4$ {\scriptsize[$+1.0$, $+5.8$]} & $+0.3$ {\scriptsize[$-0.6$, $+1.2$]} & $+1.4$ {\scriptsize[$+0.3$, $+2.5$]} & $-1.6$ {\scriptsize[$-3.0$, $-0.1$]}\\
DeBERTa-v3 & $+4.3$ {\scriptsize[$+1.8$, $+6.9$]} & $-0.7$ {\scriptsize[$-1.6$, $+0.2$]} & $+2.2$ {\scriptsize[$+0.3$, $+4.0$]} & $-2.8$ {\scriptsize[$-3.4$, $-2.2$]}\\
mDeBERTa-v3 & $+7.2$ {\scriptsize[$+3.4$, $+11.1$]} & $-0.9$ {\scriptsize[$-1.9$, $+0.2$]} & $+3.0$ {\scriptsize[$+0.3$, $+5.8$]} & $-4.9$ {\scriptsize[$-5.4$, $-4.4$]}\\
XLM-R LoRA & $+6.1$ {\scriptsize[$+4.5$, $+7.7$]} & $-0.3$ {\scriptsize[$-1.1$, $+0.5$]} & $+2.2$ {\scriptsize[$+1.1$, $+3.2$]} & $-3.8$ {\scriptsize[$-4.7$, $-3.0$]}\\
Qwen-3B echo & $+3.0$ {\scriptsize[$+1.3$, $+4.6$]} & -- & $+0.7$ {\scriptsize[$+0.2$, $+1.2$]} & --\\
Qwen-0.5B echo & $+5.3$ {\scriptsize[$+4.1$, $+6.6$]} & $+0.1$ {\scriptsize[$-1.3$, $+1.4$]} & $+1.9$ {\scriptsize[$+0.2$, $+3.6$]} & $-3.2$ {\scriptsize[$-3.8$, $-2.7$]}\\
Qwen-7B echo & $+7.4$ {\scriptsize[$+2.1$, $+12.8$]} & $+1.0$ {\scriptsize[$-0.6$, $+2.7$]} & $+2.9$ {\scriptsize[$-0.3$, $+6.1$]} & $-2.8$ {\scriptsize[$-4.6$, $-1.0$]}\\
\bottomrule
\end{tabular}
\caption{The single-copy Qwen-3B taggers as partners (gain in \eaurc{} $\times10^{-3}$, positive = helps; pooled over CoNLL and MASSIVE, Knapp--Hartung 95\% CI). ``vs echo partner'': gain of the single-copy partner minus that of the Qwen-3B echo partner for the same primary (negative = single copy is worse). Without the mask the partner is indistinguishable from the echo partner for all 7 primaries; kept causal it is worse for all 7.}
\label{tab:singlecopyh2h}
\end{table*}

\begin{figure}[tbp]
  \centering
  \includegraphics[width=\columnwidth]{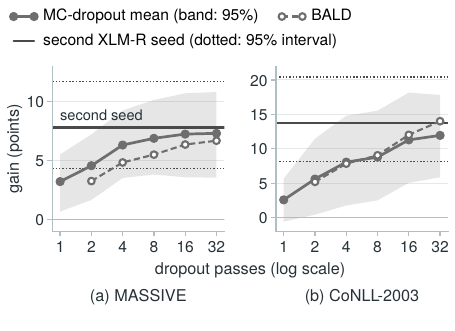}
  \caption{\textbf{For the XLM-R primary, MC dropout needs several passes to come close to a second seed.}
Gain of the MC-dropout mean (and of BALD) as a same-network second opinion, by the number of dropout passes at the trained rate, with the rank-ECDF combiner; solid line: a second XLM-R seed (dotted: its 95\% interval).
The MC mean's difference from the second seed stops being distinguishable from zero at 2 passes on MASSIVE and 4 on CoNLL, but every point estimate stays below the seed; on MASSIVE, $k$ passes cost about $0.9k$ forward passes of the primary.
Post hoc re-analysis of saved features (CoNLL: two seeds).
}
  \label{fig:mcdropout}
\end{figure}

\paragraph{XLM-R's own MC dropout.} Table~\ref{tab:mcrank} and Figure~\ref{fig:mcdropout} re-score XLM-R's MC-dropout features from the saved per-utterance dumps with the rank-ECDF combiner. One pass gives clearly less than a second seed; from 4 passes the gap is not distinguishable from zero, but every point estimate of the MC mean stays below the seed. BALD behaves similarly and matches the seed on CoNLL at 32 passes.

\begin{table*}[tp]
\centering\small\setlength{\tabcolsep}{5pt}
\begin{tabular}{@{}lrrrrr@{}}
\toprule
 & & \multicolumn{2}{c}{MASSIVE (3 seeds)} & \multicolumn{2}{c}{CoNLL (2 seeds)}\\
\cmidrule(lr){3-4}\cmidrule(l){5-6}
second opinion of XLM-R & cost & gain & minus second seed & gain & minus second seed\\
\midrule
second seed & 1.0 & 7.8 &  & 13.7 & \\
MC-dropout mean, 1 pass & 0.9 & 3.2 & $-$4.6 [$-$9.4, $-$0.1] & 2.6 & $-$11.1 [$-$17.2, $-$4.6]\\
MC-dropout mean, 2 passes & 1.9 & 4.6 & $-$3.2 [$-$8.0, 0.9] & 5.6 & $-$8.1 [$-$13.2, $-$2.7]\\
MC-dropout mean, 4 passes & 3.7 & 6.3 & $-$1.5 [$-$6.1, 2.5] & 8.1 & $-$5.6 [$-$11.9, 0.6]\\
MC-dropout mean, 8 passes & 7.5 & 6.9 & $-$0.9 [$-$5.5, 3.1] & 8.8 & $-$4.9 [$-$11.0, 1.0]\\
MC-dropout mean, 32 passes & 29.8 & 7.3 & $-$0.5 [$-$5.7, 3.9] & 11.9 & $-$1.8 [$-$8.5, 5.3]\\
BALD, 4 passes & 3.7 & 4.8 & $-$3.0 [$-$7.4, 0.9] & 7.9 & $-$5.9 [$-$11.9, 0.3]\\
BALD, 32 passes & 29.8 & 6.7 & $-$1.1 [$-$6.3, 3.2] & 14.0 & +0.3 [$-$6.3, 7.2]\\
\bottomrule
\end{tabular}
\caption{\textbf{A second XLM-R seed gains more than XLM-R's own MC-dropout mean at every pass count.} XLM-R's MC dropout against a second XLM-R seed, with the rank-mapped combiner (gain in points; MC at the trained dropout rate). Cost: measured GPU time of the MC passes relative to one primary pass (MASSIVE). Minus seed: MC gain minus second-seed gain, 95\% interval over seeds (resampled) and one shared test-utterance resample (2,000 replicates). One pass gives clearly less than a second seed on both datasets; from 4 passes the difference is no longer distinguishable from zero, but every point estimate of the MC mean stays below the seed. CoNLL seed 0 has no saved weights, so no MC passes.}
\label{tab:mcrank}
\end{table*}

\section{Trained partners in detail}
\label{app:partners}

\begin{table*}[tp]\centering\scriptsize\setlength{\tabcolsep}{3pt}
\begin{tabular}{llrrrrrrr}\toprule
dataset & primary $\leftarrow$ partner & gain & H 95\% CI & Holm $p$ & item-only CI & Knapp--Hartung CI & between-seed SD & within-seed SE\\\midrule
MASSIVE & XLM-R $\leftarrow$ own seed & +3.2 & [+1.8, +4.9] & 0.004 & [+2.4, +4.1] & [+1.1, +5.0] & 0.8 & 1.0\\
MASSIVE & XLM-R $\leftarrow$ Gemma & +7.6 & [+5.1, +10.5] & 0.004 & [+5.6, +9.9] & [+3.7, +11.2] & 1.5 & 1.4\\
MASSIVE & XLM-R $\leftarrow$ Qwen-3B & +8.7 & [+6.1, +11.7] & 0.004 & [+6.4, +11.3] & [+5.6, +11.4] & 1.1 & 1.5\\
MASSIVE & Gemma $\leftarrow$ own seed & +4.4 & [+2.6, +6.3] & 0.005 & [+3.7, +5.3] & [+1.0, +7.8] & 1.4 & 1.0\\
MASSIVE & Gemma $\leftarrow$ XLM-R & +4.7 & [+2.6, +6.9] & 0.004 & [+3.2, +6.2] & [+1.4, +7.9] & 1.3 & 1.1\\
MASSIVE & Gemma $\leftarrow$ Qwen-3B & +5.1 & [+3.3, +7.0] & 0.004 & [+3.9, +6.3] & [+2.5, +7.7] & 1.1 & 1.0\\
MASSIVE & Qwen-3B $\leftarrow$ own seed & +4.7 & [+3.0, +6.4] & 0.004 & [+3.9, +5.6] & [+2.0, +7.6] & 1.1 & 1.0\\
MASSIVE & Qwen-3B $\leftarrow$ XLM-R & +5.3 & [+3.0, +7.5] & 0.004 & [+3.6, +7.0] & [+2.0, +8.4] & 1.3 & 1.2\\
MASSIVE & Qwen-3B $\leftarrow$ Gemma & +4.7 & [+2.5, +6.9] & 0.004 & [+3.4, +6.0] & [+0.8, +8.5] & 1.5 & 1.1\\
CoNLL & XLM-R $\leftarrow$ own seed & +2.7 & [+0.5, +4.9] & 0.083 & [+2.0, +3.5] & [-2.3, +7.5] & 2.0 & 0.9\\
CoNLL & XLM-R $\leftarrow$ Gemma & +8.5 & [+5.9, +11.6] & 0.004 & [+6.4, +10.8] & [+4.1, +12.2] & 1.6 & 1.4\\
CoNLL & XLM-R $\leftarrow$ Qwen-3B & +7.4 & [+4.5, +10.7] & 0.004 & [+5.3, +9.5] & [+1.3, +13.0] & 2.3 & 1.4\\
CoNLL & Gemma $\leftarrow$ own seed & +2.1 & [-0.4, +4.3] & 0.158 & [+1.5, +2.7] & [-3.9, +8.0] & 2.4 & 0.7\\
CoNLL & Gemma $\leftarrow$ XLM-R & +1.6 & [+0.2, +3.4] & 0.083 & [+0.4, +2.9] & [-0.2, +3.0] & 0.8 & 0.8\\
CoNLL & Gemma $\leftarrow$ Qwen-3B & +1.3 & [-0.2, +2.7] & 0.158 & [+0.4, +2.2] & [-1.3, +3.9] & 1.1 & 0.7\\
CoNLL & Qwen-3B $\leftarrow$ own seed & +1.7 & [+0.7, +2.7] & 0.024 & [+1.0, +2.3] & [+1.2, +2.0] & 0.2 & 0.7\\
CoNLL & Qwen-3B $\leftarrow$ XLM-R & +2.5 & [+0.7, +4.8] & 0.042 & [+1.1, +4.2] & [-0.4, +5.2] & 1.1 & 1.0\\
CoNLL & Qwen-3B $\leftarrow$ Gemma & +3.2 & [+1.4, +5.2] & 0.011 & [+2.1, +4.4] & [-0.3, +6.6] & 1.4 & 0.9\\
\bottomrule\end{tabular}
\caption{\textbf{Trained partners help in all 18 pre-specified primary--partner tests, significantly in all 9 on MASSIVE.} Pre-specified secondary family (C1): partner helps, gain $=\text{eAURC}(\text{margin})-\text{eAURC}(\text{margin}+\savg(B))$ in $\times10^{-3}$ \eaurc{}, positive = helps (Holm over 18); rank-ECDF combiner. H = hierarchical seed$\times$item bootstrap with one item draw shared by all seeds; item-only CIs keep seeds fixed; KH = random-effects interval over the 3 seed estimates with Knapp--Hartung $t_2$ (conservative).}\label{tab:secC1}\end{table*}

\begin{table*}[tp]
\centering\scriptsize\setlength{\tabcolsep}{2.4pt}
\begin{tabular}{@{}lrrrr@{}}
\toprule
pooled gain $\times10^{-3}$ & rank-mapped & raw & \hd{ranked by\\$\savg(B)$ alone} & \hd{fixed equal\\weights}\\
\midrule
Gemma $\leftarrow$ own seed & $+3.00$ [$+1.74$, $+4.36$] & $+3.02$ [$+1.72$, $+4.44$] & $+2.86$ [$+1.33$, $+4.46$] & $+2.65$ [$+1.70$, $+3.66$]\\
Gemma $\leftarrow$ XLM-R & $+2.91$ [$+1.95$, $+4.09$] & $+2.86$ [$+1.84$, $+4.14$] & $+2.13$ [$+0.90$, $+3.61$] & $+2.80$ [$+2.04$, $+3.73$]\\
XLM-R $\leftarrow$ own seed & $+3.51$ [$+2.10$, $+5.08$] & $+3.46$ [$+1.97$, $+5.07$] & $+3.26$ [$+1.49$, $+5.11$] & $+2.94$ [$+1.93$, $+4.03$]\\
XLM-R $\leftarrow$ Gemma & $+7.76$ [$+6.22$, $+9.40$] & $+7.84$ [$+6.24$, $+9.62$] & $+7.78$ [$+6.00$, $+9.70$] & $+6.37$ [$+5.22$, $+7.64$]\\
XLM-R $\leftarrow$ Qwen-0.5B & $+5.98$ [$+4.27$, $+7.81$] & $+6.06$ [$+4.20$, $+8.04$] & $+5.59$ [$+3.52$, $+7.79$] & $+5.25$ [$+3.95$, $+6.61$]\\
Gemma $\leftarrow$ copy-1 average & $+0.78$ [$+0.13$, $+1.46$] & $-1.37$ [$-2.54$, $-0.15$] & $-12.00$ [$-14.19$, $-9.91$] & $-0.66$ [$-1.76$, $+0.45$]\\
Gemma $\leftarrow$ MC mean & $+0.02$ [$-0.03$, $+0.07$] & $+0.02$ [$-0.04$, $+0.10$] & $+0.03$ [$-0.07$, $+0.14$] & $+0.02$ [$-0.03$, $+0.08$]\\
Gemma $\leftarrow$ second-head average & $-0.68$ [$-1.41$, $+0.05$] & $-2.92$ [$-4.25$, $-1.68$] & $-8.38$ [$-10.34$, $-6.45$] & $-1.86$ [$-2.93$, $-0.81$]\\
\bottomrule
\end{tabular}
\caption{Trained partners help under every combiner, including none; the linear same-network readouts do not. Pooled over the three datasets, 95\% CIs; gain in \eaurc{}, positive = helps. ``Ranked by $\savg(B)$ alone'' replaces the margin by the averaged margin; for the copy-1 average this alone is much worse than the margin.}
\label{tab:combfree}
\end{table*}

\begin{table*}[tbp]
\centering\small\setlength{\tabcolsep}{5pt}
\begin{tabular}{@{}lrrr@{}}
\toprule
 & pretraining & architecture & recipe\\
\midrule
MASSIVE & $+1.10$ & $+3.42$ & $+0.47$\\
CoNLL & $+2.62$ & $+2.27$ & $+1.43$\\
SNIPS & $+1.04$ & $+1.74$ & $+1.94$\\
pooled & $+1.59$ & $+2.48$ & $+1.28$\\
\quad 95\% CI & [$+0.60$, $+2.61$] & [$+1.23$, $+3.81$] & [$+0.35$, $+2.33$]\\
\quad share of seed gain & 0.45 & 0.71 & 0.37\\
\quad random effects CI & [-0.56, $+4.12$] & [$+0.45$, $+4.98$] & [-0.46, $+2.72$]\\
\midrule
power, rule, $2\times10^{-3}$ & 0.21 & 0.06 & 0.50\\
power, rule, $1\times10^{-3}$ & 0.01 & 0.00 & 0.02\\
smallest effect with 80\% power, rule & 3.00 & 3.88 & 2.43\\
the same, pooled test & 1.37 & 1.74 & 1.11\\
\bottomrule
\end{tabular}
\caption{One-property partners of the XLM-R primary against an own-seed partner (gain of the one-property partner minus that of an own seed, \eaurc{} $\times10^{-3}$; positive = the one-property partner is better). Smallest effect: smallest common gain detected with 80\% power. The architecture partner (Qwen2.5-0.5B with LoRA) also differs in pretraining, tokenizer and recipe.}
\label{tab:power}
\end{table*}

\paragraph{Cheap ensembles.} For echo taggers the own-seed partner is already cheap to train: a second LoRA adapter on the same frozen base. A snapshot ensemble of one Gemma run's checkpoints improves on the margin on MASSIVE (gain $2.8\times10^{-3}$, 95\% CI [1.4, 4.2]) but gains $1.7\times10^{-3}$ less than the own-seed partner (CI [0.0, 3.7] less).

\section{Larger models and GLiNER}
\label{app:gliner}

\begin{table*}[tp]
\centering\scriptsize\setlength{\tabcolsep}{2.4pt}
\begin{tabular}{@{}lllrrrrr@{}}
\toprule
dataset & \hd{pair\\(GFLOPs, \% of 7B)} & metric & \hd{pair\\better by} & \hd{crossed\\95\% CI} & \hd{$t_2$\\95\% CI} & \hd{90\%\\bound} & \hd{2 healthy seeds\\(per seed)}\\
\midrule
MASSIVE & Qwen-3B $\times$2 (186, 85\%) & F1 & $+0.034$ & [$+0.003$, $+0.078$] & [$-0.069$, $+0.136$] & 0.065 & $+0.010$ ($+0.016$, $+0.004$)\\
 & & AUGRC & $+0.0105$ & [$+0.0011$, $+0.0251$] & [$-0.0225$, $+0.0435$] & 0.021 & $+0.0029$ ($+0.0039$, $+0.0018$)\\
 & XLM-R + Qwen-3B (99, 45\%) & F1 & $+0.043$ & [$+0.009$, $+0.091$] & [$-0.069$, $+0.155$] & 0.078 & $+0.017$ ($+0.026$, $+0.009$)\\
 & & AUGRC & $+0.0125$ & [$+0.0025$, $+0.0279$] & [$-0.0225$, $+0.0475$] & 0.023 & $+0.0044$ ($+0.0058$, $+0.0030$)\\
CoNLL & Qwen-3B $\times$2 (486, 85\%) & F1 & $+0.001$ & [$-0.004$, $+0.006$] & [$-0.008$, $+0.010$] & 0.005 & --\\
 & & AUGRC & $+0.0018$ & [$+0.0000$, $+0.0038$] & [$-0.0020$, $+0.0057$] & 0.0035 & --\\
 & XLM-R + Qwen-3B (256, 45\%) & F1 & $+0.000$ & [$-0.006$, $+0.006$] & [$-0.013$, $+0.013$] & 0.005 & --\\
 & & AUGRC & $+0.0009$ & [$-0.0007$, $+0.0028$] & [$-0.0021$, $+0.0040$] & 0.0025 & --\\
\bottomrule
\end{tabular}
\caption{\textbf{Pairs of smaller taggers match a 7B tagger at 45--85\% of its compute; on CoNLL they are equivalent within one F1 point.} One larger model against two pre-specified pairs that use less compute (Holm over 8). Positive = the pair is better: pair minus 7B for F1, 7B minus pair for AUGRC. Pairs average their tempered word posteriors, so they also make new predictions. Crossed CI: seeds resampled, one shared utterance draw. By the pre-specified rule a difference counts only if the bootstrap with Holm and the $t$ interval (2 df) agree; none does. 90\% bound: smallest $d$ with the 90\% crossed interval inside $\pm d$ (with the $t_2$ 90\% interval: 0.007 and 0.009 for CoNLL F1). On CoNLL the pairs are equivalent to the 7B tagger within one F1 point. Last column: MASSIVE without seed 1, which a dev-only rule written afterwards (a run more than 3 F1 points below the mean of its other two seeds) flags as diverged (best dev F1 0.744 against 0.818 and 0.817); among the 27 model $\times$ dataset sets with saved dev logs no other run is flagged, but some XLM-R and all Gemma runs have no saved dev logs. Per-seed values in brackets.}
\label{tab:abe}
\end{table*}

\paragraph{A larger single model.} Two pre-specified pairs using 85\% and 45\% of the FLOPs of a Qwen2.5-7B echo tagger (two Qwen-3B taggers; XLM-R plus Qwen-3B) are equivalent to it on CoNLL within one F1 point; on MASSIVE one 7B run diverged, and without it XLM-R plus Qwen-3B is 1.7 F1 points better on two seeds. In measured time the Qwen-3B pair is slower than the 7B on MASSIVE (Appendix~\ref{app:compute}). This does not test \citet{abe2022}, whose single model is more accurate than the members: our untuned 7B tagger is about as accurate as the 3B one.

\begin{table}[tbp]
\centering\footnotesize\setlength{\tabcolsep}{2pt}
\begin{tabular}{@{}llr@{}}
\toprule
test (CoNLL, rank-mapped) & gain & 95\% CI\\
\midrule
G1 XLM-R + GLiNER vs margin & $+3.2$ & [$+2.4$, $+4.1$]\\
G2 Gemma + GLiNER vs margin & $+1.4$ & [$+0.8$, $+2.1$]\\
G3 Qwen-3B + GLiNER vs margin & $+1.1$ & [$+0.7$, $+1.6$]\\
G4 XLM-R: GLiNER vs own seed & $+0.5$ & n.s.\\
G5 Gemma: GLiNER vs own seed & $-0.7$ & n.s.\\
G6 Qwen-3B: GLiNER vs own seed & $-0.5$ & n.s.\\
\bottomrule
\end{tabular}
\caption{Fine-tuned GLiNER as a partner (tests G1--G6; gain in \eaurc{} $\times10^{-3}$, positive = helps; G4--G6: GLiNER minus own seed; Holm over 6; G1--G3 $p<0.001$, also under the crossed bootstrap). G2 passes the small-sample check under rank-ECDF only narrowly and fails it under the raw combiner ($t_2$ interval [$-0.14$, $+2.9$]).}
\label{tab:gliner}
\end{table}

\begin{figure*}[tp]
  \centering
  \includegraphics[width=\textwidth]{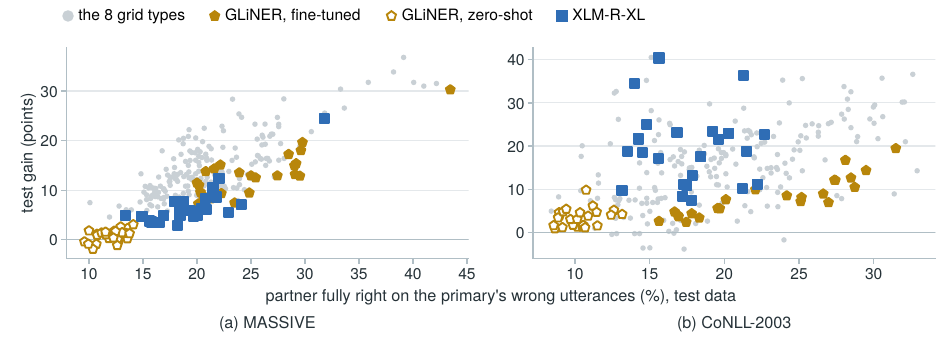}
  \caption{\textbf{A prior from other datasets absorbs partner types that sit off the grid's line (Figure~\ref{fig:priorcheck}b).}
Grey: pairings among the 8 grid types; pentagons: GLiNER fine-tuned (filled) and zero-shot (hollow); squares: XLM-R-XL; one point per primary seed.
Fine-tuned GLiNER, and XLM-R-XL on MASSIVE, gain less than grid partners that are right as often; XLM-R-XL on CoNLL gains more.
With such types among the candidates, adding a prior from the other datasets to the held-out check helps (Figure~\ref{fig:priorcheck}).
}
  \label{fig:newpartners}
\end{figure*}

\paragraph{Why GLiNER is not used as a primary.} After rounding, 89--91\% of fine-tuned GLiNER's CoNLL test sentences share one confidence value and 52--58\% of its errors sit inside that tie, so a partner's ``gain'' for a GLiNER primary would mostly be tie-breaking. Its posteriors also barely move the averaged margin on the primary's errors, which accounts for its shortfall on MASSIVE but only a small part of it on CoNLL.

\section{Multilingual details}
\label{app:mlexp}
\label{app:cjk}

\paragraph{Relative gain and one combiner.} The main text uses rank-ECDF and relative gain (share of the primary's margin \eaurc{}, 0.093, 0.121 and 0.187 for XLM-R, Gemma and Qwen). Regressing the asymmetry on the F1 gap cannot separate model type from accuracy: XLM-R is the more accurate model in 286 of 288 decoder-primary comparisons, and the adjusted decoder coefficient changes sign with the functional form.

\paragraph{Accuracy across locales.} In the 48 locales other than Japanese and Chinese, mean span F1 is 59.6 for XLM-R, 40.5 for Gemma and 26.3 for Qwen-3B; an XLM-R primary gains no more from Gemma than from its own seed (difference $-1.1$ [$-3.5$, $+1.5$] points). Gold segmentation keeps Gemma ahead of XLM-R by 13--23 F1 points in Japanese and Chinese. Over all groups, accuracy on errors and the F1 gap explain the gain equally well ($R^2$ 0.82 each).

\paragraph{Competent taggers.} The $R^2$ comparison of \S\ref{sec:shift} (within each primary-seed-locale group) restricted to groups where the primary and all three partners reach F1 $\geq0.40$ (104 groups, 312 observations, 17 locales): accuracy on errors 0.80, F1 gap 0.57 (linear) or 0.70 (cubic); partner identity alone 0.73. The lead over the cubic F1 gap is 0.09 [0.03, 0.14] (locale bootstrap). Post hoc.

\paragraph{Segmentation of Japanese and Chinese.} MASSIVE writes Japanese and Chinese mostly without spaces. Splitting them into one character per word, the convention of the official baselines, gives our XLM-R-large 6.1, 11.7 and 11.4 F1 on ja-JP, zh-CN and zh-TW, far below the 15.4, 24.8 and 25.2 published for zero-shot XLM-R-base \citep{fitzgerald2023}. We therefore segment with SudachiPy (mode A; \citealp{takaoka2018sudachi}) and jieba\footnote{\url{https://github.com/fxsjy/jieba}} and score exact character spans. Segmentation limits what any model can get right: jieba's words can express 92.0\% and 91.4\% of the gold spans in zh-CN and zh-TW (oracle span F1 92.6 and 92.2; ja-JP 99.2\%). The segmented inputs also keep MASSIVE's original spaces, which fall on gold slot boundaries far more often than chance (42\% of interior spaces in ja-JP against a 9\% base rate; zh-TW 29\% against 14\%; zh-CN 19\% against 15\%), so they carry some boundary information. With gold segmentation instead, Gemma still leads XLM-R (ja-JP 36.2 against 23.4; zh-CN 52.0 against 29.3; zh-TW 46.6 against 30.8), so the accuracy reversal does not depend on jieba's unreachable spans. zh-CN and zh-TW are the same language, Mandarin, in simplified and traditional script \citep{fitzgerald2023}. One primary breaks the pattern: on zh-CN the Qwen-3B primary gains 32 points from XLM-R and 16 from its own seed, although it is slightly more accurate than XLM-R (28.7 against 27.1 F1; Table~\ref{tab:locales}).

\begin{figure*}[tp]
  \centering
  \includegraphics[width=\textwidth]{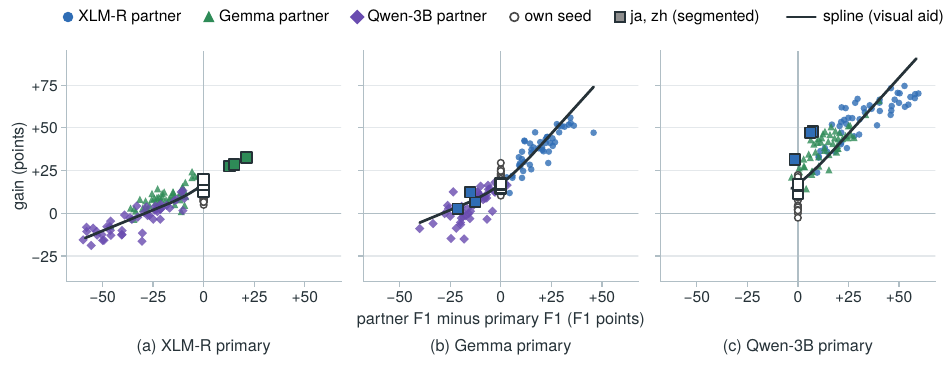}
  \caption{\textbf{Across locales, a partner's gain follows how much more accurate it is than the primary.}
en-US taggers applied zero-shot to 51 MASSIVE locales; one point per locale and partner (mean of 3 seeds), coloured by the partner's model type; hollow: the primary's own seed; squares: Japanese and Chinese after word segmentation.
Curve: one spline over all primaries, a visual aid fitted to the points where both taggers reach span F1 $\geq 10$; all points are drawn.
Because the F1 gaps are large, F1 gap and accuracy on the primary's errors explain the gain about equally well here (Section~\ref{sec:shift}); this is a test-set analysis.
}
  \label{fig:localegain}
\end{figure*}

\paragraph{Gain against relative accuracy.} Figure~\ref{fig:localegain} plots every locale and partner: a partner's gain follows how much more accurate it is than the primary.

\begin{table}[tbp]
\centering\small\setlength{\tabcolsep}{2.5pt}
\begin{tabular}{@{}lccc@{}}
\toprule
predictor of gain & 48 locales & XLM-R & decoders\\
\midrule
F1 gap (spline) & .83 & .66 & .90--.91\\
accuracy on primary's errors & .82 & .45 & .86--.87\\
error correlation $\phi$ & .10 & & \\
both & .87 & & \\
\bottomrule
\end{tabular}
\caption{Across locales, F1 gap and accuracy on the primary's errors explain gain about equally ($R^2$ of relative gain within each primary-seed-locale group; zero-shot, rank-ECDF; 1,296 observations). With a cubic instead of a spline F1 gap the value is .82, as in Section~\ref{sec:shift}. Difference $-0.01$ (95\% CI $-0.04$ to $+0.01$, locale $\times$ utterance bootstrap). In English, where partners differ by at most 9 F1 points, accuracy on errors explains far more than F1 gap (.38/.32/.15 against .09/.08/.02, test data).}
\label{tab:xloverlap}
\end{table}

\begin{table}[tbp]
\centering\footnotesize\setlength{\tabcolsep}{2.2pt}
\begin{tabular}{@{}lrrrrr@{}}
\toprule
subset & obs. & locales & \begin{tabular}[b]{@{}c@{}}acc. on\\errors\end{tabular} & \begin{tabular}[b]{@{}c@{}}cubic\\F1 gap\end{tabular} & \begin{tabular}[b]{@{}c@{}}partner\\identity\end{tabular}\\
\midrule
all F1 $\geq.30$ & 537 & 25 & .83 & .75 & .75\\
all F1 $\geq.40$ & 312 & 17 & .80 & .70 & .73\\
all F1 $\geq.50$ & 150 & 10 & .77 & .60 & .68\\
\midrule
\multicolumn{6}{@{}l}{\emph{48 locales, one primary at a time}}\\
\quad XLM-R primary & 432 & 48 & .45 & .67 & \\
\quad Gemma primary & 432 & 48 & .86 & .92 & \\
\quad Qwen-3B primary & 432 & 48 & .87 & .92 & \\
\bottomrule
\end{tabular}
\caption{Cross-lingual $R^2$ of relative gain within each primary-seed-locale group (zero-shot; accuracy on errors and gain come from the same test utterances, so this is not a held-out test). The competent-subset lead of accuracy on errors over the cubic F1 gap holds at every threshold: +.09 [.05, .12], +.09 [.03, .14] and +.17 [.08, .23] (locale bootstrap). Partner identity alone explains almost as much. Over all 48 locales and one primary at a time, the cubic F1 gap explains more than accuracy on errors for every primary.}
\label{tab:xlingsens}
\end{table}

\begin{table}[tbp]
\centering\footnotesize\setlength{\tabcolsep}{1.6pt}
\begin{tabular}{@{}lrrr@{}}
\toprule
$D$ ($\times10^{-3}$) & Gemma & Qwen-3B & XLM-R\\
\midrule
zero-shot, raw & $+23.7$ & $+87.7$ & $-3.5$\\
zero-shot, rank-ECDF & $+23.3$ & $+88.1$ & $-1.2$\\
few-shot calibration, raw & $+26.7$ & $+86.1$ & $-1.5$\\
zero-shot, AUGRC & $+7.7$ & $+27.0$ & $-0.9$\\
zero-shot, without th, km, my & $+24.1$ & $+87.5$ & $-3.0$\\
zero-shot, ja/zh segmented (51) & $+21.9$ & $+85.3$ & $-2.6$\\
zero-shot, primary F1 $\ge0.5$ & $+13.3$ & $+14.2$ & $-3.9$\\
\quad locales in that subset & 12 & 3 & 39\\
\bottomrule
\end{tabular}
\caption{Other-type minus own-seed partner in the 48 locales other than Japanese and Chinese: $D$ = difference in absolute gain ($\times10^{-3}$ \eaurc{}, positive = the other type helps more), with locale-clustered robust (CR2) intervals. Section~\ref{sec:shift} gives the same contrast relative to the primary's margin \eaurc{} ($+18.7$ points for the Gemma primary and $-1.1$ for XLM-R, zero-shot, rank-ECDF). XLM-R's partner is Gemma; the decoders' partner is XLM-R. Zero-shot raw XLM-R: CI [$-6.2$, $-0.8$]; rank-ECDF and few-shot: not significant.}
\label{tab:xlmain}
\end{table}

\begin{table}[tbp]\centering\small\setlength{\tabcolsep}{3pt}
\begin{tabular}{llrrrr}\toprule
 & & \multicolumn{2}{c}{zero-shot (main)} & \multicolumn{2}{c}{few-shot}\\
primary & locale & raw & rank & raw & rank\\\midrule
XLM-R & ja-JP & +16.3 & +14.9 & +18.5 & +18.2\\
 & zh-CN & +15.1 & +14.6 & +13.6 & +14.3\\
 & zh-TW & +5.4 & +7.3 & +6.1 & +7.7\\
\midrule
Gemma & ja-JP & $-$5.4 & $-$8.0 & +5.6 & +2.2\\
 & zh-CN & $-$12.8 & $-$17.1 & $-$2.8 & $-$4.8\\
 & zh-TW & $-$2.0 & $-$6.6 & +5.6 & $-$0.7\\
\bottomrule\end{tabular}
\caption{Japanese and Chinese (segmented with SudachiPy/jieba): other-type minus own-seed gain under the four calibration conditions ($\times10^{-3}$ \eaurc{}, positive = the other type helps more, mean of 3 seeds). For the XLM-R primary the other type (Gemma) helps more in all 12 locale $\times$ condition combinations. For the Gemma primary the sign depends on the condition: XLM-R helps less than an own seed in every zero-shot case, but not consistently with few-shot calibration.}\label{tab:cjk}\end{table}

\begin{table*}[tp]\centering\scriptsize\setlength{\tabcolsep}{1.6pt}
\begin{tabular}{@{}l rrr rr rr rr @{\hspace{6pt}} l rrr rr rr rr @{\hspace{6pt}} l rrr rr rr rr@{}}\toprule
 & \multicolumn{3}{c}{F1} & \multicolumn{2}{c}{X primary} & \multicolumn{2}{c}{G primary} & \multicolumn{2}{c}{Q primary} &  & \multicolumn{3}{c}{F1} & \multicolumn{2}{c}{X primary} & \multicolumn{2}{c}{G primary} & \multicolumn{2}{c}{Q primary} &  & \multicolumn{3}{c}{F1} & \multicolumn{2}{c}{X primary} & \multicolumn{2}{c}{G primary} & \multicolumn{2}{c}{Q primary}\\
locale & X & G & Q & own & G & own & X & own & X & locale & X & G & Q & own & G & own & X & own & X & locale & X & G & Q & own & G & own & X & own & X\\\midrule
af-ZA & 65 & 56 & 40 & 10 & 18 & 20 & 28 & 14 & 56 & hy-AM & 56 & 20 & 7 & 12 & 3 & 21 & 51 & 8 & 70 & pt-PT & 67 & 56 & 53 & 9 & 12 & 16 & 30 & 17 & 36\\
am-ET & 40 & 16 & 1 & 10 & 6 & 18 & 36 & 0 & 51 & id-ID & 62 & 51 & 39 & 8 & 15 & 19 & 36 & 22 & 55 & ro-RO & 67 & 51 & 37 & 8 & 10 & 22 & 39 & 14 & 67\\
ar-SA & 54 & 39 & 35 & 9 & 12 & 16 & 26 & 14 & 36 & is-IS & 56 & 35 & 19 & 11 & 8 & 17 & 36 & 10 & 61 & ru-RU & 69 & 55 & 49 & 9 & 9 & 18 & 41 & 18 & 51\\
az-AZ & 56 & 26 & 16 & 12 & 3 & 16 & 51 & 10 & 60 & it-IT & 67 & 56 & 54 & 7 & 10 & 15 & 21 & 17 & 38 & sl-SL & 66 & 45 & 30 & 11 & 12 & 18 & 37 & 15 & 63\\
bn-BD & 60 & 38 & 4 & 12 & 6 & 20 & 43 & 16 & 68 & jv-ID & 45 & 41 & 21 & 13 & 22 & 17 & 12 & 10 & 38 & sq-AL & 58 & 37 & 17 & 10 & 10 & 18 & 39 & 4 & 65\\
ca-ES & 59 & 44 & 37 & 8 & 5 & 13 & 28 & 15 & 41 & ka-GE & 58 & 24 & 11 & 14 & $-$4 & 26 & 53 & 2 & 67 & sv-SE & 74 & 57 & 46 & 7 & 9 & 23 & 47 & 17 & 65\\
cy-GB & 47 & 23 & 16 & 13 & 4 & 10 & 29 & 3 & 46 & km-KH & 44 & 33 & 13 & 9 & 1 & 16 & 25 & 5 & 55 & sw-KE & 42 & 36 & 14 & 11 & 24 & 17 & 21 & 2 & 45\\
da-DK & 68 & 56 & 46 & 11 & 8 & 17 & 43 & 22 & 62 & kn-IN & 56 & 23 & 3 & 12 & 1 & 21 & 52 & 4 & 62 & ta-IN & 56 & 27 & 1 & 12 & 4 & 24 & 47 & 2 & 64\\
de-DE & 70 & 64 & 59 & 12 & 21 & 18 & 22 & 19 & 36 & ko-KR & 63 & 40 & 30 & 13 & 14 & 19 & 45 & 13 & 59 & te-IN & 60 & 28 & 2 & 11 & 8 & 21 & 51 & 12 & 67\\
el-GR & 64 & 46 & 18 & 10 & 8 & 23 & 38 & $-$3 & 63 & lv-LV & 65 & 39 & 14 & 10 & 8 & 17 & 39 & 4 & 70 & th-TH & 51 & 49 & 42 & 15 & 15 & 17 & 17 & 14 & 24\\
es-ES & 66 & 56 & 56 & 7 & 5 & 16 & 29 & 18 & 33 & ml-IN & 59 & 25 & 1 & 11 & 3 & 30 & 56 & 9 & 70 & tl-PH & 55 & 53 & 34 & 12 & 21 & 16 & 21 & 10 & 47\\
fa-IR & 68 & 40 & 15 & 9 & 3 & 22 & 52 & 12 & 75 & mn-MN & 55 & 9 & 5 & 13 & $-$8 & 13 & 47 & 2 & 65 & tr-TR & 65 & 42 & 32 & 10 & 9 & 20 & 45 & 16 & 61\\
fi-FI & 72 & 43 & 26 & 10 & 7 & 19 & 51 & 11 & 73 & ms-MY & 60 & 49 & 33 & 10 & 16 & 20 & 32 & 20 & 56 & ur-PK & 54 & 28 & 8 & 10 & 12 & 25 & 44 & 12 & 60\\
fr-FR & 66 & 57 & 56 & 7 & 13 & 15 & 27 & 20 & 33 & my-MM & 50 & 23 & 4 & 14 & 5 & 14 & 42 & 4 & 58 & vi-VN & 51 & 36 & 30 & 7 & 10 & 19 & 35 & 21 & 53\\
he-IL & 45 & 21 & 24 & 5 & 9 & 19 & 41 & 12 & 40 & nb-NO & 69 & 57 & 46 & 10 & 10 & 20 & 42 & 20 & 62 & ja-JP$^*$ & 23 & 36 & 16 & 13 & 28 & 15 & 7 & 12 & 48\\
hi-IN & 65 & 46 & 5 & 8 & 12 & 21 & 42 & 18 & 70 & nl-NL & 70 & 61 & 53 & 10 & 10 & 21 & 33 & 17 & 52 & zh-CN$^*$ & 27 & 48 & 29 & 17 & 33 & 17 & 3 & 16 & 32\\
hu-HU & 64 & 41 & 23 & 13 & 9 & 17 & 41 & 8 & 66 & pl-PL & 64 & 49 & 40 & 7 & 13 & 18 & 37 & 20 & 52 & zh-TW$^*$ & 28 & 44 & 22 & 20 & 29 & 17 & 12 & 17 & 47\\
\bottomrule\end{tabular}
\caption{Per-locale results in the main setting (en-US taggers applied zero-shot; rank-ECDF combiner fit on en-US). F1: test span F1$\times100$ (X = XLM-R, G = Gemma, Q = Qwen-3B). Gain columns: share of the primary's margin \eaurc{} removed (\%, positive = helps) by an own-seed partner and by the other model type (Gemma for XLM-R, XLM-R for the decoders); mean of 3 seeds. $^*$Segmented with SudachiPy or jieba. Thai, Khmer and Burmese use MASSIVE's whitespace chunks as words (Appendix~\ref{app:data}).}
\label{tab:locales}\end{table*}

\paragraph{The calibration law.} We had pre-specified that the benefit of a second model grows with the primary's calibration error across locales. The law failed: once the error rate is controlled, the slope of the gain on calibration error is $+0.083$ (97.5\% CI [$-0.007$, $+0.175$]). Copy-1 and MC slopes are flat; the second-head slope is negative.
\paragraph{Published numbers.} On the 46 locales other than Japanese and Chinese that have a published XLM-R-base zero-shot score \citep{fitzgerald2023}, our XLM-R-large averages 59.4 F1 against 52.3, correlates at $r=0.80$ and is higher in 42 of 46.

\section{Serving}
\label{app:systems}

\begin{table*}[tp]\centering\footnotesize\setlength{\tabcolsep}{2.4pt}
\begin{tabular}{llcccccc}\toprule
 & & \multicolumn{2}{c}{wrong outputs avoided per 1,000} & \multicolumn{2}{c}{batch 1 (ms)} & batch 32 & memory\\\cmidrule(lr){3-4}\cmidrule(lr){5-6}
data & system & 80\% coverage & 90\% coverage & p50 & p95 & q/s & GB\\\midrule
MASSIVE & XLM-R & +0.0 [0.0, 0.0] & +0.0 [0.0, 0.0] & 11.3 & 11.6 & 1,801 & 1.1\\
 & XLM-R $\times$2 & +5.5 [1.2, 9.4] & +7.6 [3.7, 11.5] & 12.4 & 13.3 & 1,471 & 2.1\\
 & XLM-R + Gemma & +7.4 [1.7, 12.9] & +8.6 [2.9, 14.5] & 51.4 & 54.1 & 494 & 9.7\\
 & Gemma & $-$5.2 [$-$16.1, 4.7] & $-$11.4 [$-$21.9, $-$1.3] & 49.1 & 51.1 & 501 & 8.7\\
 & Gemma $\times$2 & +3.1 [$-$4.7, 11.3] & $-$0.4 [$-$8.2, 7.6] & 54.5 & 57.7 & 409 & 17.3\\
 & re-ranked by 2nd seed, $\savg$ & +5.6 [1.9, 9.0] & +2.4 [$-$1.3, 5.9] & 12.4 & 13.3 & 1,471 & 2.1\\
 & re-ranked by 2nd seed, $\slab$ & +5.6 [1.6, 9.4] & +4.9 [2.4, 8.9] & \multicolumn{4}{c}{as above}\\
 & XLM-R $\to$ Gemma 20\%, tiered & +0.0 [0.0, 1.9] & +7.5 [3.1, 12.3] & 11.3 & 62.0 & 430$^\dagger$ & 9.7\\
 & XLM-R $\to$ Gemma 20\%, calibrated & +2.0 [$-$0.8, 5.4] & +7.5 [3.1, 12.3] & \multicolumn{4}{c}{as above}\\
\midrule
CoNLL & XLM-R & +0.0 [0.0, 0.0] & +0.0 [0.0, 0.0] & 11.5 & 11.8 & 1,669 & 1.1\\
 & XLM-R $\times$2 & +4.0 [0.3, 7.2] & +1.7 [$-$2.8, 6.1] & 12.4 & 13.4 & 1,404 & 2.2\\
 & XLM-R + Gemma & +12.8 [8.9, 17.7] & +12.5 [8.1, 16.9] & 53.4 & 56.8 & 359 & 9.8\\
 & Gemma & +12.2 [7.0, 18.0] & +13.5 [6.9, 20.0] & 51.1 & 53.7 & 372 & 8.7\\
 & Gemma $\times$2 & +16.3 [11.0, 22.7] & +17.0 [9.9, 24.2] & 56.3 & 59.6 & 198 & 17.4\\
 & re-ranked by 2nd seed, $\savg$ & +2.8 [0.2, 5.8] & +2.7 [$-$0.8, 6.1] & 12.4 & 13.4 & 1,404 & 2.2\\
 & re-ranked by 2nd seed, $\slab$ & +3.1 [0.3, 6.1] & +2.7 [$-$0.9, 6.2] & \multicolumn{4}{c}{as above}\\
 & XLM-R $\to$ Gemma 20\%, tiered & +0.7 [$-$0.2, 3.9] & +9.6 [6.2, 12.9] & 11.7 & 65.2 & 410$^\dagger$ & 9.7\\
 & XLM-R $\to$ Gemma 20\%, calibrated & +0.7 [$-$0.2, 3.9] & +9.6 [6.2, 12.9] & \multicolumn{4}{c}{as above}\\
\bottomrule\end{tabular}
\caption{\textbf{A second XLM-R seed avoids 4.0--5.5 wrong outputs per 1,000 at 80\% coverage and keeps 82--84\% of XLM-R's throughput.} Serving systems (one H100, eager PyTorch, bf16; single checkpoints, each model in its own process; peak allocated memory). Errors avoided: wrong outputs among the accepted ones, per 1,000 queries, against XLM-R alone (mean of 3 seeds, 95\% CIs). ``Re-ranked'': XLM-R's labels are kept and its second seed only re-scores them, with the averaged margin $\savg$ or the label-aware score $\slab$. The cascade routes the 20\% least confident queries to XLM-R + Gemma; ``tiered'' ranks every routed query below every unrouted one, ``calibrated'' ranks both on fit-slice error probabilities. $^\dagger$Measured with a batch-synchronous server in which the unrouted queries of a batch wait for its routed pass, so this throughput is understated.}
\label{tab:serving}\end{table*}

\paragraph{Fixed labels.} At 80\% coverage labels change only on utterances that are rejected anyway, so re-ranking alone matches the label-changing pair (5.6 against 5.5 on MASSIVE). At 90\% most of a second seed's 7.6 comes from better labels: the averaged posterior raises exact match by 8.4 per 1,000.

\paragraph{The routed cascade.} The cascade sends queries whose XLM-R margin falls below a fit-slice threshold to XLM-R + Gemma. Two artefacts affect a simple evaluation of it. First, routed queries were scored $10+$ their averaged-margin score, so every routed query ranked below every unrouted one; for coverage $c\geq1-r$ the accepted set was then XLM-R's own by construction. Ranking both groups on fit-slice error probabilities instead (one logistic model per group) changes little: routing 20\% avoids 2.0 [$-0.8$, 5.4] errors per 1,000 on MASSIVE at 80\% coverage (tiered: 0.0) and is unchanged elsewhere (Table~\ref{tab:serving}).

\begin{figure}[tbp]
  \centering
  \includegraphics[width=\columnwidth]{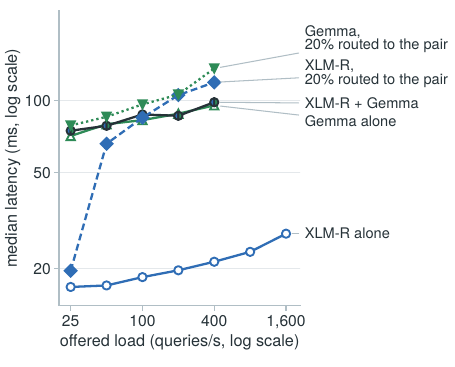}
  \caption{\textbf{XLM-R alone sustains 1,600 queries per second; systems with the Gemma decoder sustain up to 400.}
Median latency under open-loop load on MASSIVE (dynamic batching up to 32 requests or 5\,ms; one repeat; saturated points omitted).
``XLM-R, 20\% routed to the pair'': XLM-R tags every query and sends its 20\% least confident to XLM-R + Gemma; ``Gemma, 20\% routed'' is the same with Gemma first.
The routed curves come from a batch-synchronous server and overstate their latency.
}
  \label{fig:load}
\end{figure}

\paragraph{Cascade timing.} Second, our server handled each batch synchronously, so unrouted queries waited for the batch's routed pass; batch-32 throughput (430 q/s) and the latency under load (Figure~\ref{fig:load}) therefore understate what a cascade with per-request completion would achieve. At batch size 1 the cascade keeps XLM-R's median latency (11--12\,ms), with a 95th percentile of 62--65\,ms.

\paragraph{Compiled and multi-adapter serving.} torch.compile speeds models up 1.7--3.2 times but changes 0.2--0.5\% of predictions, so we report eager numbers. A second LoRA adapter on one base saves memory (9.1 against 17.3\,GB), but our naive version is the slowest system; dedicated multi-adapter servers \citep{sheng2024slora,chen2024punica} were not tested.

\section{Evaluation lessons in detail}
\label{app:lessons}
At this resolution, evaluation choices matter: four reversed one of our conclusions before we fixed them (Table~\ref{tab:lessons}); others also question evaluation practice for selective prediction \citep{ding2020,jaeger2023,gu2023,traub2024}.
\begin{table}[tbp]
\centering\footnotesize\setlength{\tabcolsep}{3pt}
\begin{tabular}{@{}>{\raggedright\arraybackslash}p{0.69\columnwidth}>{\raggedright\arraybackslash}p{0.28\columnwidth}@{}}
\toprule
Pitfall we hit & What we do now\\
\midrule
\textbf{Combiner.} A noise feature cost a raw logistic combiner 28\% of a seed's gain (MASSIVE) to about 3 times it (CoNLL). & Rank-map the features.\\
\textbf{Shared resamples.} Separate test resamples per seed made the copy-1 gain significant ($p=0.02$; shared: $p\approx0.07$). & Share one resample.\\
\textbf{Low power.} A rule with 6--50\% power found no partner property that matters; pooled over datasets, all three help a little. & Compute power first.\\
\textbf{Same-set predictor.} A quantity measured on test data looked like an explanation; only held-out and prospective tests showed it predicts. & Fix the rule; test on new data.\\
\bottomrule
\end{tabular}
\caption{Four evaluation choices that each reversed one of our conclusions (Appendix~\ref{app:lessons}).}
\label{tab:lessons}
\end{table}

\paragraph{Combiner.} A logistic model on standardised raw features penalises any added feature, even pure noise (Table~\ref{tab:noise}). The cost follows the skew of the raw margin (Spearman $+0.79$ across primaries) and the shift between fit slice and test ($+0.39$). \eaurc{} is dominated by the order of the most confident utterances; the raw margin squeezes them into a near-tie that any extra standardised feature reshuffles, while the rank map spreads them out. Post-hoc fixes to badly behaved confidence estimators are related \citep{cattelan2024}.

\begin{table}[tbp]
\centering\small\setlength{\tabcolsep}{3pt}
\begin{tabular}{lrrr}
\toprule
cost of one noise feature & MASSIVE & CoNLL & SNIPS\\
\midrule
raw combiner ($\times10^{-3}$) & 0.98 & 4.68 & 1.84\\
\quad share of own-seed gain & 28\% & 286\% & 98\%\\
rank-ECDF ($\times10^{-3}$) & 0.07 & 0.03 & 0.04\\
\quad share of own-seed gain & 2.0\% & 1.6\% & 2.3\%\\
\bottomrule
\end{tabular}
\caption{Median increase in \eaurc{} when one $N(0,1)$ feature is added to the margin (20 draws, 24 primaries per dataset). An independent re-run with other random draws gives 4.18 on CoNLL (about 2--3 times an own-seed gain).}
\label{tab:noise}
\end{table}

\paragraph{Other lessons.} \emph{Outside the data:} whether the MC mean sat above, on or below the curve depended only on the curve's form, because it lies outside the range of the trained partners. Do not place points outside the fitted range. \emph{Fit slice and cross-fitting:} CoNLL test is temporally shifted from train, so a fit slice drawn from train is shifted too; this made a learned router look worse than the margin. Cross-fitting within test must compute the metric within each held-out fold; ranking pooled out-of-fold scores from differently fitted models produced a spurious ``XLM-R partner is worse''. \emph{Metric:} the calibration-law slope was not significant under \eaurc{} and significant under AUGRC. Fix the metric in the protocol and report the other. \emph{Bootstrap:} with 10,000 replicates the Monte-Carlo error of a Holm-adjusted $p$ near 0.05 is about 0.01, and between-seed standard deviations are often as large as within-seed bootstrap errors, so item-only intervals understate uncertainty.

\section{Further signals tested}
\label{app:neg}
\paragraph{Copy-1 signal across locales.} Across the 51 locales the copy-1 signal is more often harmful (significantly, in 37 of 153 locale--seed cells) than helpful (6 of 153).

\section{Error analysis}
\label{app:error}
We looked at the Gemma echo primary on MASSIVE-en (seed 0). Of 689 wrong utterances, 53 fall in the more confident half by margin. The XLM-R pair moves 6 of these into its 30\% most uncertain utterances and leaves 38 in its confident half. Missed spans dominate the confident errors (26 of 53). A partner catches exactly the errors it does not share: XLM-R is right on 4 of the 6 caught errors and on none of the 38 missed ones. Several shared errors look like annotation choices, for example \textit{amazon} labelled as a person in ``check any mail from amazon''.
\begin{table}[!htb]\centering\scriptsize\setlength{\tabcolsep}{3pt}
\begin{tabular}{lrrrrr}\toprule
set & $n$ & missed & spurious & wrong type & boundary\\\midrule
all errors & 689 & 27\% & 24\% & 8\% & 42\%\\
confident errors & 53 & 49\% & 28\% & 9\% & 13\%\\
caught by partner & 6 & 67\% & 17\% & 0\% & 17\%\\
still missed & 38 & 53\% & 26\% & 8\% & 13\%\\
\bottomrule\end{tabular}
\caption{Error types of the Gemma echo primary (MASSIVE-en test, seed 0). Confident errors: errors in the more confident half by margin. Caught: confident errors that the XLM-R ensemble signal places among the 30\% most uncertain utterances; still missed: placed in its more confident half. XLM-R's own output is fully correct on 67\% of caught and 0\% of missed confident errors.}\label{tab:errtypes}\end{table}

\section{Extended related work}
\label{app:related}
\paragraph{Complementary errors and diversity.}
The complementary rate of \citet{brill1998} counts where a second tagger is right and the first is wrong; voting and stacking of taggers go back to part-of-speech tagging \citep{brill1998,vanhalteren2001} and CoNLL-2003 NER \citep{florian2003,wu2003}.
\citet{kuncheva2003} compare pairwise diversity measures, among them the double-fault rate \citep{giacinto2001}, the Q statistic and error correlation, and find their link to ensemble accuracy weak for real classifier ensembles.
Theory relates diversity to ensemble error \citep{krogh1995,wood2023}, negative correlation learning trains members to differ \citep{liu1999}, \citet{brown2005} survey how diversity is created, and \citet{theisen2023} ask when ensembles help.
Ensemble selection picks members by their gain on held-out data \citep{caruana2004,zhou2002}.
\emph{We} turn complementarity into a partner-selection rule, test it prospectively, and find held-out gain equivalent (Table~\ref{tab:policies}).

\paragraph{Ensembles and uncertainty.}
Independent ensembles give better uncertainty than single models \citep{lakshminarayanan2017,ovadia2019,fort2019}; members trained more differently make less correlated errors \citep{gontijolopes2022}, varying hyperparameters helps \citep{wenzel2020}, and disagreement tracks error \citep{jiang2022sgd,baek2022}; \citet{abe2022} argue that one more accurate model can match much of this.
For LLMs, disagreement with other models complements self-consistency \citep{hamidieh2026}.
Cheaper alternatives share most weights \citep{wen2020,havasi2021,laurent2023,wang2023lora,balabanov2024}, approximate a posterior over adapter weights \citep{yang2024laplace}, distil an ensemble into one model \citep{hinton2015,malinin2020,tran2020hydra} or avoid a second model \citep{liu2020sngp}.
\emph{These} works ask whether a second model helps; we ask which one (Table~\ref{tab:policies}).

\paragraph{Confidence for taggers and selective prediction.}
Our richer single-model baseline is the signal set of \citet{vazhentsev2022} (see also \citealp{shelmanov2021}).
Calibration has been studied for neural networks \citep{guo2017}, pre-trained transformers \citep{desai2020}, structured outputs \citep{kuleshov2015,jagannatha2020} and across languages \citep{jiang2022,ahuja2022}.
Other work studies ensemble uncertainty for structured prediction \citep{malinin2021}, uncertainty estimation for NER \citep{hashimoto2024}, conformal prediction sets for NER \citep{singer2026}, selective prediction for NER under domain shift \citep{zheng2026} and selective prediction in NLP \citep{kamath2020,varshney2022sel,vazhentsev2023}; learned or second-model confidence scores are a further line \citep{jiang2018trust,corbiere2019}, and several papers question evaluation practice \citep{ding2020,jaeger2023,galil2023,gu2023,traub2024}.
\emph{These} works improve one tagger's confidence; we choose and use a second model on top of it, and our partners help beyond the richer baseline.

\paragraph{Decoders and serving.}
Repeating the input lets a causal decoder's second copy attend to the whole input, for embeddings \citep{springer2025echo} and for tagging \citep{kukic2026,ewais2026}; removing the causal mask does so directly \citep{li2023lsllama,behnamghader2024}.
UniversalNER \citep{zhou2024}, GoLLIE \citep{sainz2024} and GLiNER \citep{zaratiana2024} could also serve as partners.
Deferral and cascades \citep{wang2022committees,varshney2022casc,jitkrittum2023,gupta2024,chen2023frugal} decide when to call a larger model; multi-adapter servers \citep{sheng2024slora,chen2024punica}, LLM serving \citep{kwon2023vllm} and prediction-serving systems \citep{crankshaw2017,kang2017noscope} lower the cost of several models.

\section{Reproducibility and the Responsible NLP checklist}
\label{app:repro}
\begin{itemize}[leftmargin=*,itemsep=1pt]
\item \textbf{Code and outputs.} Code for training, logit dumping, analysis and re-computation is saved next to each result, with the protocol and its hash; a numbers file maps each number in the paper to its result file. We will release the code, the hashed protocols, the per-item outputs and the trained taggers.
\item \textbf{Artifacts and licences.} Datasets and their licences: Table~\ref{tab:splits} and the Ethics Statement; models, sizes and training recipes: Table~\ref{tab:models}; pinned revisions: Gemma-4-E2B revision d29ff6b4, Qwen2.5-7B-Instruct snapshot a09a3545. CoNLL text cannot be redistributed.
\item \textbf{Hyperparameters and seeds.} Three training seeds per trained tagger; bootstrap random seeds are fixed in code. Nothing was tuned on test; the only selections were learning rates for XLM-R with LoRA and XLM-R-XL on seed-0 dev F1 (Appendix~\ref{app:models}).
\item \textbf{Statistics, compute and data checks.} Every interval states what it resamples (\S\ref{sec:setup}, Appendix~\ref{app:signals}); Appendix~\ref{app:prereg} lists the Holm-corrected confirmatory families and every protocol with its outcome, including the rules that failed. GPU hours and FLOPs: Appendix~\ref{app:compute}; splits are rebuilt by scripts, and duplicates are reported in Appendix~\ref{app:data}.
\end{itemize}

\end{document}